\documentclass[aps,prx,noshowpacs,twocolumn,nofootinbib]{revtex4-2}
\usepackage[utf8]{inputenc}
\usepackage{ifthen}
\usepackage{color}
\usepackage{graphics,graphicx,epsfig}
\usepackage{epsf,epstopdf,wrapfig}
\usepackage{amssymb,amsfonts,amsmath,babel,mathtools,bm}
\usepackage[export]{adjustbox}
\include{epsf}
\usepackage{ifthen}		
\usepackage{subfigure}

\usepackage{mathtools}
\usepackage{amsthm}

\usepackage{multirow}

\definecolor{ocre}{RGB}{10,100,185}
\usepackage[colorlinks = true,
            linkcolor = ocre,
            urlcolor  = ocre,
            citecolor = ocre,
            anchorcolor = ocre]{hyperref}
\usepackage[normalem]{ulem}

\def\eqref#1{equation~\ref{#1}}

\def\1{\bm{1}}

\def\vh{{\bm{h}}}

\def\vr{{\bm{r}}}

\def\vx{{\bm{x}}}
\def\vy{{\bm{y}}}
\def\vz{{\bm{z}}}

\def\mD{{\bm{D}}}

\def\mI{{\bm{I}}}

\def\mW{{\bm{W}}}

\DeclareMathAlphabet{\mathsfit}{\encodingdefault}{\sfdefault}{m}{sl}
\SetMathAlphabet{\mathsfit}{bold}{\encodingdefault}{\sfdefault}{bx}{n}

\usepackage[capitalize,noabbrev]{cleveref}

\theoremstyle{plain}

\theoremstyle{definition}

\theoremstyle{remark}

\usepackage{soul}

\newcommand{\EQ}{\begin{equation}}
\newcommand{\EE}{\end{equation}}
\newcommand{\EQA}{\begin{eqnarray}}
\newcommand{\EEA}{\end{eqnarray}}
\renewcommand{\d}{{\text{d}}}

\newcommand{\newlinecell}[2][c]{%
  \begin{tabular}[#1]{@{}c@{}}#2\end{tabular}}
  
\newcommand{\corres}{Correspondence should be addressed to:
Gian Marco Visani: {gvisan01@cs.washington.edu}, and Armita Nourmohammad: {armita@uw.com}.}

\begin{document}

\title{Holographic-(V)AE: an end-to-end SO(3)-Equivariant \\ (Variational) Autoencoder in Fourier Space}
\author{Gian Marco Visani}
\thanks{\corres}
\affiliation{Paul G. Allen School of Computer Science and Engineering, University of Washington, 85 E Stevens Way NE, Seattle, WA 98195, USA}
\author{Michael N. Pun}
\affiliation{Department of Physics, University of Washington, 3910 15th Avenue Northeast, Seattle, WA 98195, USA}
\author{Arman Angaji}
\affiliation{Institute for Biological Physics, University of Cologne, Z\"ulpicher Str. 77, 50937 Cologne, Germany}
\author{Armita Nourmohammad}
\thanks{\corres}
 \affiliation{Department of Physics, University of Washington, 3910 15th Avenue Northeast, Seattle, WA 98195, USA}
\affiliation{Paul G. Allen School of Computer Science and Engineering, University of Washington, 85 E Stevens Way NE, Seattle, WA 98195, USA}
\affiliation{Department of Applied Mathematics, University of Washington, 4182 W Stevens Way NE, Seattle, WA 98105, USA}
 \affiliation{Fred Hutchinson Cancer Center, 1241 Eastlake Ave E, Seattle, WA 98102, USA}

\begin{abstract}
Group-equivariant neural networks have emerged as a data-efficient approach to solve classification and regression tasks, while respecting the relevant symmetries of the data. However, little work has been done to extend this paradigm to the unsupervised and generative domains. Here, we present \textit{Holographic}-(Variational) Auto Encoder (H-(V)AE), a fully end-to-end SO(3)-equivariant (variational) autoencoder in Fourier space, suitable for unsupervised learning and generation of data  distributed around a specified origin in 3D. 
H-(V)AE is trained to reconstruct the spherical Fourier encoding of data, learning in the process a low-dimensional representation of the data (i.e., a latent space) with a maximally informative rotationally invariant embedding alongside an equivariant frame describing the orientation of the data. We extensively test the performance of H-(V)AE on diverse datasets. We show that the learned latent space efficiently encodes the categorical features of spherical images. Moreover, H-(V)AE's latent space can be used to extract compact embeddings for protein structure microenvironments, and when paired with a  Random Forest Regressor, it enables  state-of-the-art predictions of protein-ligand binding affinity.
\end{abstract}
\maketitle

\section{Introduction}
In supervised learning, e.g. for classification tasks, the success of state-of-the-art algorithms is often attributed to respecting known inductive biases of the function they are trying to approximate. One such bias is the invariance of the function to  certain transformations of the input. For example, image classification should be  translationally invariant, in that the output should not depend on the position of the object in the image. To achieve such invariance, conventional techniques use data augmentation to train an algorithm on many transformed forms of the data. However, this solution is only approximate and increases training time significantly, up to prohibitive scales for high-dimensional and continuous transformations ($\sim$500 augmentations are required to learn 3D rotation-invariant patterns~\cite{geiger_e3nn_2022}). Alternatively, one could use invariant features of the data (e.g. pairwise distance between different features) as input to train any machine learning algorithm~\cite{capecchi_one_2020}. However, the choice of these invariants is arbitrary and the resulting network could lack in expressiveness.

Recent advances have brought concepts from group theory to develop symmetry-aware neural network architectures that  are equivariant under actions of different symmetry groups~\cite{weiler_3d_2018, cohen_spherical_2018,thomas_tensor_2018, kondor_clebsch-gordan_2018, esteves_spin-weighted_2020, fuchs_se3-transformers_2020, brandstetter_geometric_2022, musaelian_learning_2022, satorras_en_2022, liao_equiformer_2022}. Equivariance {with respect to a symmetry group} is the property that, if the input is transformed {via a group action}, then the  output is transformed according to a linear operation determined by the symmetry group {itself; it is easy to see that invariance is a special case of equivariance, where the linear operation is simply the identity}.  These group equivariant networks can systematically treat and interpret various transformation in data, and learn models that are agnostic to the specified transformations. For example, models equivariant to euclidean transformations have recently advanced the state-of-the-art on {many} supervised tasks for classification and regression~\cite{weiler_3d_2018, cohen_spherical_2018,thomas_tensor_2018, kondor_clebsch-gordan_2018, esteves_spin-weighted_2020, fuchs_se3-transformers_2020, brandstetter_geometric_2022, musaelian_learning_2022, satorras_en_2022, liao_equiformer_2022}. These models are more flexible and data-efficient compared to their purely invariant counterparts~\cite{geiger_e3nn_2022, batzner_e3-equivariant_2022}.

Extending such group invariant and equivariant paradigms to unsupervised learning (i.e., for modeling the data distribution) could provide compact representations of data that are agnostic to a specified symmetry transformation. In machine learning, auto-encoders (AE's) and their probabilistic version,  variational auto-encoders (VAE's), are among the artificial neural networks that are commonly used for unsupervised learning, in that they provide an efficient representation of unlabeled data~\cite{kramer_autoassociative_1992, kingma_introduction_2019}. However, little work has been done to implement group equivariance in auto-encoder architectures~\cite{winter_unsupervised_2022}.

Here, we focus on developing  neural network architectures for unsupervised learning that are equivariant to rotations around a specified origin in 3D, denoted by the group SO(3). To define rotationally equivariant transformations, it is convenient to project data to spherical Fourier space~\cite{kondor_clebsch-gordan_2018}. Accordingly, we encode the data in spherical Fourier space by constructing {\em  holograms} of the data that are conveniently structured for equivariant operations. These data holograms are inputs to our SO(3)-equivariant (variational) autoencoder in spherical Fourier space, with a fully equivariant encoder-decoder architecture trained to reconstruct the Fourier coefficients of the input; we term this approach \textit{Holographic}-(V)AE (or H-(V)AE). Our network learns an SO(3)-equivariant latent space composed of a maximally informative set of invariants and an equivariant frame describing the orientation of the data.

{We extensively test the performance and properties of H-(V)AE on two domains. First, we focus on {spherical images}, demonstrating high accuracy in unsupervised classification and clustering tasks. Second, we focus on {structural biology}, and demonstrate that H-(V)AE can be effectively used to construct compact, informative, and symmetry-aware representations of protein structures, which can be used for downstream tasks. Specifically, we leverage H-(V)AE trained on a large corpus of protein structure micro-environments to construct local representations of protein-ligand binding pockets that are both rotationally and translationally equivariant (i.e., SE(3) equivariant). When combined with a simple Random Forest Regressor, we achieve state-of-the-art accuracy on the task of  predicting the binding affinity between a protein and a ligand in complex.
Our code and pre-trained models are available at \href{https://github.com/gvisani/Holographic-VAE}{https://github.com/gvisani/Holographic-VAE}.}

\section{Model}
\subsection{Representation and transformation of 3D data in spherical bases}
We are interested in modeling 3D data (i.e., functions in $\mathbb{R}^{3}$), for which the global orientation of the data should not impact the inferred model~\cite{einstein_grundlage_1916}. We consider functions distributed around a specified origin, which we  express by the resulting spherical coordinates $(r,\theta,\phi)$ around the origin; $\theta$ and $\phi$ are the azimuthal and the polar angles and $r$ defines the distance to the reference point in the spherical coordinate system. In this case, the set of rotations about the origin define the 3D rotation group SO(3), and we will consider models that are rotationally equivariant under SO(3).

To define rotationally equivariant transformations, it is convenient to project data to spherical Fourier space. We use spherical harmonics to encode the angular information of the data. Spherical harmonics are a class of functions that form a complete and orthonormal basis for functions $f(\theta,\phi)$ defined on a unit sphere ($r=1$). In their complex form, spherical harmonics are defined as,
\begin{equation}
	Y_{\ell m}(\theta,\phi) = \sqrt{\frac{2n+1}{4\pi}\frac{(n-m)!}{(n+m)!}}e^{im\phi}P^m_\ell(\cos\theta)
	\label{eq:YLM}
\end{equation}
where $\ell$ is a non-negative integer ($0\leq \ell$) and $m$ is an integer within the interval $-\ell\leq m \leq \ell$. $P^m_\ell(\cos \theta)$ is the Legendre polynomial of degree $\ell$ and order $m$, which, together with the complex exponential $e^{im\phi}$, define sinusoidal functions over the angles $\theta$ and $\phi$. In quantum mechanics, spherical harmonics are used to  represent  the orbital angular momenta, e.g. for an electron in a hydrogen atom. In this context, the degree $\ell$ relates to the eigenvalue of the square of the angular momentum, and the order $m$ is the eigenvalue of the angular momentum  about the azimuthal axis.

To encode a general function $\rho(r,\theta,\phi)$ with both radial and angular components, we use the Zernike Fourier transform,
\begin{equation}
    \hat{Z}^{n}_{\ell m} = \int\,\rho(r, \theta, \phi) \, Y_{\ell m}(\theta,\phi) R^{n}_{\ell}(r)  \,\d V
    \label{eq:ZFT}
\end{equation}
where $Y_{\ell m}(\theta,\phi) $ is the spherical harmonics of degree $\ell$ and order $m$, and $R^{n}_{\ell}(r)$ is the radial Zernike polynomial in 3D (Eq.~\ref{eqn:zernike}) with radial frequency $n \geq 0$ and degree $\ell$. $R^{n}_{\ell}(r)$ is non-zero only for even values of $n-\ell\geq 0$. Zernike polynomials  form a complete orthonormal basis in 3D, and therefore, can be used to expand and retrieve 3D shapes, if large enough $\ell$ and $n$ values are used; approximations that restrict the series to finite $n$ and $\ell$ are often sufficient for shape retrieval, and hence, desirable algorithmically. 
Thus, in practice, we cap the resolution of the ZFT to a maximum degree $L$ and a maximum radial frequency $N$.

A class of functions that we  consider in this work are 3D point clouds, e.g., the atomic composition of a protein in space. We represent point clouds $\rho(\vr) \equiv \rho(r,\theta,\phi)$ by the sum of Dirac-$\delta$ functions centered at each point:
\begin{equation}
    \rho(\vr) = \sum_{i\in \text{points}} \delta(\rho(\vr_{i}) - \rho(\vr))
    \label{eq:dirac_delta}
\end{equation}
where $\delta(x)=1$ for $x=0$ and it is zero, otherwise. The resulting ZFT of a point cloud follows a closed form, and notably,  it does not require a discretization of 3D space for numerical computation,
\begin{equation}
    \hat{Z}_{\ell m}^{n} = \sum_{i\in \text{points}} R_{n}^{\ell}(r_i) Y_{\ell m}(\theta_i, \varphi_i)
    \label{eqn:spherical_and_radial_with_dirac_ft}
\end{equation}
We can  reconstruct the data using the inverse ZFT and define approximations by truncating the angular and radial frequencies at $L$ and $N$ (see Section~\ref{sec:expanded_background}).

Conveniently, the angular representation of the data by spherical harmonics in a ZFT transform form an equivariant basis under rotation in 3D, implying that if the input  (i.e., atomic coordinates of a protein) is rotated, then the  output is transformed according to a linear operation determined by the rotation. The linear operator that describes how spherical harmonics transform under rotations are called the Wigner D-matrices. Notably, Wigner-D matrices are the irreducible representations (irreps) of SO(3). Therefore, the SO(3) group acts on spherical Fourier space via a direct sum of irreps. Specifically, the ZFT encodes a data point into a \textit{tensor} composed of a direct sum of \textit{features}, each associated with a degree $\ell$  indicating the irrep that it transforms with under the action of SO(3). We refer to these tensors as SO(3)\textit{-steerable tensors} and to the vector spaces they occupy as SO(3)\textit{-steerable vector spaces}, or simply \textit{steerable} for short since we only deal with the SO(3) group in this work. 

We note that a tensor may contain multiple features of the same degree $\ell$, which we generically refer to as distinct \textit{channels} $c$. For example, for 3D atomic point clouds, these features include the identity and chemical properties of the constituent atoms. Throughout the paper, we refer to generic steerable tensors as $\vh$ and index them by $\ell$ (degree of $Y_{\ell m}$), $m$ (order of $Y_{\ell m}$) and $c$ (channel type). We adopt the ``hat" notation for individual entries (e.g. $\hat h _{\ell m}$) to remind ourselves of the analogy with Fourier coefficients; see Figure~\ref{fig:model}A for a graphical illustration of a tensor.

 One key feature of neural networks is applying nonlinear activations,  which enable a network to approximately model complex and non-linear phenomena. Commonly used nonlinearities include reLU,  tanh, and softmax functions. However, these conventional non-linearities can break rotational equivariance in the Fourier space. To construct expressive rotationally equivariant neural networks we can use the Clebsch-Gordan (CG) tensor  product $\otimes_{cg}$,  which is the natural nonlinear (more specifically, bilinear in the case of using two sets of Fourier coefficients) operation in the space of spherical harmonics~\cite{tung_group_1985}.

 The CG tensor product combines two features of degrees $\ell_1$ and $\ell_2$ to produce another feature of degree $|\ell_2-\ell_1|\leq\ell_3\leq |\ell_1+\ell_2|$. Let $\vh_{\ell} \in \mathbb{R}^{2\ell + 1}$ be a generic degree $\ell$ tensor, with individual components $\hat h_{\ell m}$ for $-\ell\leq m \leq \ell $. The CG tensor product is given by,
\begin{equation}
\begin{aligned}
    \hat  h_{\ell_3 m_3} &= (\vh_{\ell_1} \otimes_{cg} \vh_{\ell_2})_{\ell_3 m_3} \\
    &= \sum_{m_1 = -\ell_1}^{\ell_1} \sum_{m_2 = -\ell_2}^{\ell_2} C_{(\ell_1 m_1)(\ell_2 m_2)}^{(\ell_3 m_3)} \hat h_{\ell_1 m_1} \hat h_{\ell_2 m_2}
    \label{eqn:cg_tensor_product}
\end{aligned}
\end{equation}
where $C_{(\ell_1 m_1)(\ell_2 m_2)}^{(\ell_3 m_3)}$ are the Clebsch-Gordan coefficients, and can be precomputed for all degrees of spherical tensors~\cite{tung_group_1985}. Similar to spherical harmonics, Clebsch-Gordan tensor products also appear in quantum mechanics, and they are used to express couplings between angular momenta. In following with recent work on group-equivariant machine learning~\cite{kondor_clebsch-gordan_2018}, we will use Clebsch-Gordan products to express nonlinearities in 3D rotationally equivariant neural networks for protein structures.

\section{Holographic-(V)AE (H-(V)AE)}
\label{sec:hvae}

\begin{figure*}[t!]
    \centering
    \includegraphics[width=0.80\textwidth]{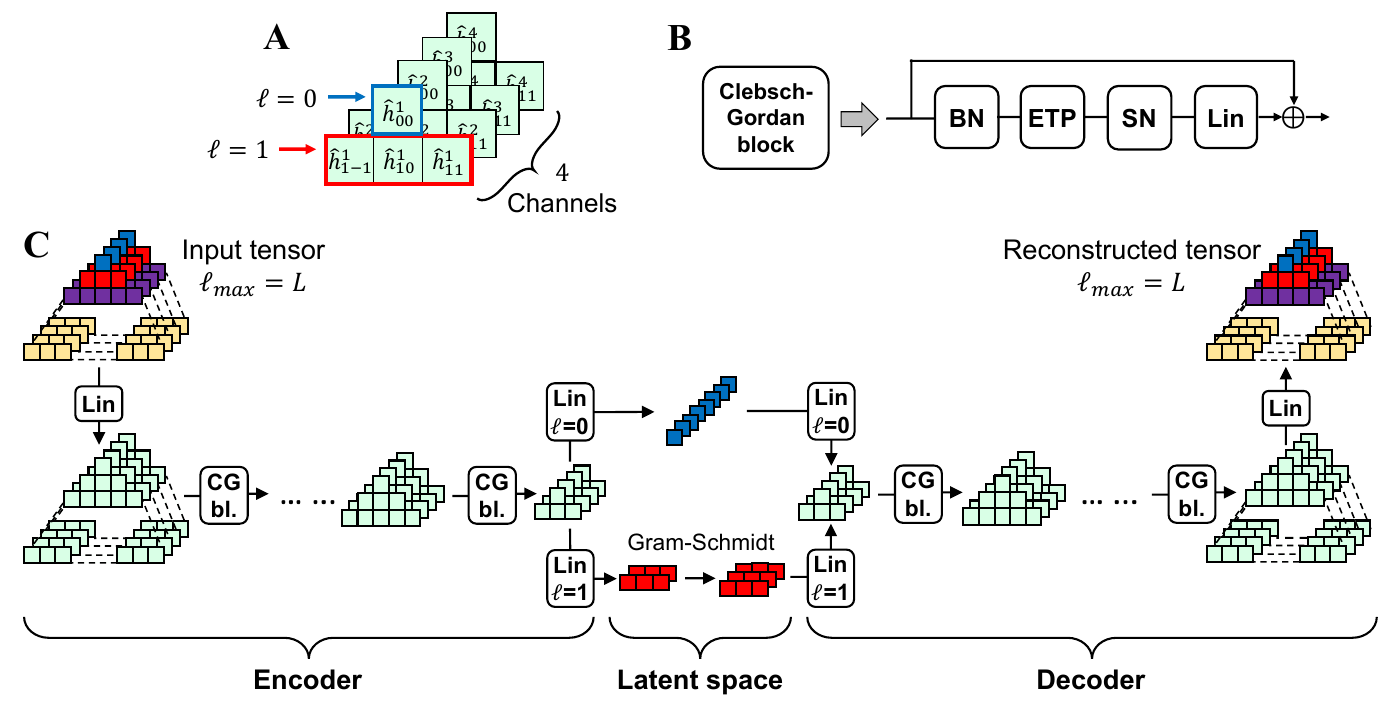}
    \caption{{\bf \small Schematic of the Network architecture.} \textbf{A:} Schematic of a steerable tensor with $\ell_\text{max} = 1$ and 4 channels per feature degree. We choose a pyramidal representation that naturally follows the expansion in size of features of higher degree. \textbf{B:} Schematic of a Clebsch-Gordan Block (CG bl.), with batch norm (BN), efficient tensor product (ETP), and signal norm (SN), and Linear (Lin) operations. \textbf{C:} Schematic of the H-AE architecture. We color-code features of different degrees in the input and in the latent space for clarity. The H-VAE schematic differs only in the latent space, where two sets of invariants are learned (means and standard deviations of an isotropic Gaussian distribution).}
    \label{fig:model}
\end{figure*}

\noindent {\bf Network architecture and training.}  H-(V)AE consists of an encoder that, through learned linear projections and pre-set nonlinear operations, project the data  onto a compressed rotationally equivariant latent space. A trained decoder that is similarly constructed then takes this latent projection and reconstructs the input data. The combination of leaned linear and pre-set nonlinear operations form equivariant Clebsch-Gordan blocks (CG bl.) both for the encoder and the decoder; see Figure~\ref{fig:model} and below for details on the structure of  a Clebsch-Gordan block. \\

Using the Clebsch-Gordan blocks, we  construct a fully rotationally equivariant architecture for unsupervised learning. Specifically, the encoder takes as input a steerable tensor with maximum degree $\ell_\text{max} = L$ and, via a stack of Clebsch-Gordan blocks, iteratively and equivariantly transfers information from higher degrees to lower ones, down to the final encoder layer with  $\ell_\text{max} = 1$, resulting in the invariant ($\ell=0$) and the frame-defining equivariant ($\ell=1$) embeddings.  The frame is constructed by learning two vectors from the $\ell=1$ embedding in the final layer of the encoder and using Gram-Schmidt to find the corresponding orthonormal basis~\cite{schmidt_zur_1907}. The third orthonomal basis vector is then calculated as the cross product of the first two.

The decoder learns to reconstruct the input from the invariant ($\ell=0$) embedding of the encoder's final layer $\mathbf{z}$ and the frame, iteratively increasing the maximum degree $\ell_\text{max}$ of the intermediate representations by leveraging the CG Tensor Product within the Clebsch-Gordan blocks (Figure~\ref{fig:model}C). We refer the reader to Section~\ref{sec:architecture_details} for further details on the design choices of the network.

To add stochasticity and make the model variational (i.e., constructing H-VAE as opposed to H-AE), we parameterize the \textit{invariant} part of the latent space by an isotropic Gaussian, i.e., we learn two sets of size $z$ invariants, corresponding to means and standard deviations. \\

We train H-(V)AE to minimize the reconstruction loss $\mathcal{L}_{rec}$ between the original and the reconstructed tensors, and, for H-VAE only, to minimize the Kullback-Leibler divergence  $D_{KL}$ of the posterior invariant latent space distribution $q(\vz|\vx)$ from the selected prior $p(\vz)$~\cite{kingma_auto-encoding_2013}:
\begin{equation}
    \mathcal{L}(\vx, \vx') = \alpha \mathcal{L}_{\text{rec}}(\vx, \vx') + \beta D_{KL}(q(\vz|\vx) || p(\vz))
    \label{eqn:training_objective}
\end{equation}
We use mean square error (MSE) for $\mathcal{L}_{\text{rec}}$, which as we show in Section~\ref{sec:pairwise_inv_rec_loss}, respects the necessary property of SO(3) pairwise invariance, ensuring that the model remains rotationally equivariant. We cast an isotropic normal prior to the invariant latent space: $p(\vz) = \mathcal{N}(\bf 0, \mI)$. Hyperparameters $\alpha$ and $\beta$ control the trade-off between reconstruction accuracy and latent space regularization~\cite{higgins_beta-vae_2022}; see Section~\ref{sec:training_objective_hyperparams} for details on tuning of these rates during training.\\

As a result of this training, H-(V)AE learns a \textit{disentangled} latent space consisting of a maximally informative invariant ($\ell = 0$) component $\mathbf{z}$ of arbitrary size, as well as three orthonormal vectors ($\ell = 1$), which represent the global 3D orientation of the object and reflect the {\em coordinate frame} of the input tensor. Crucially, the disentangled nature of the latent space is respected at all stages of training, and is guaranteed by the model's rotational equivariance; see Figure~S1 for a visual example. We discuss the equivariant properties of H-(V)AE in more detail in Figure~\ref{fig:canonical_frames_1,7}, and empirically verify the equivariance of our model up to numerical errors in Table~\ref{table:equivariance_error}.\\

\begin{figure*}[t!]
    \centering
    \includegraphics[width=0.83\textwidth]{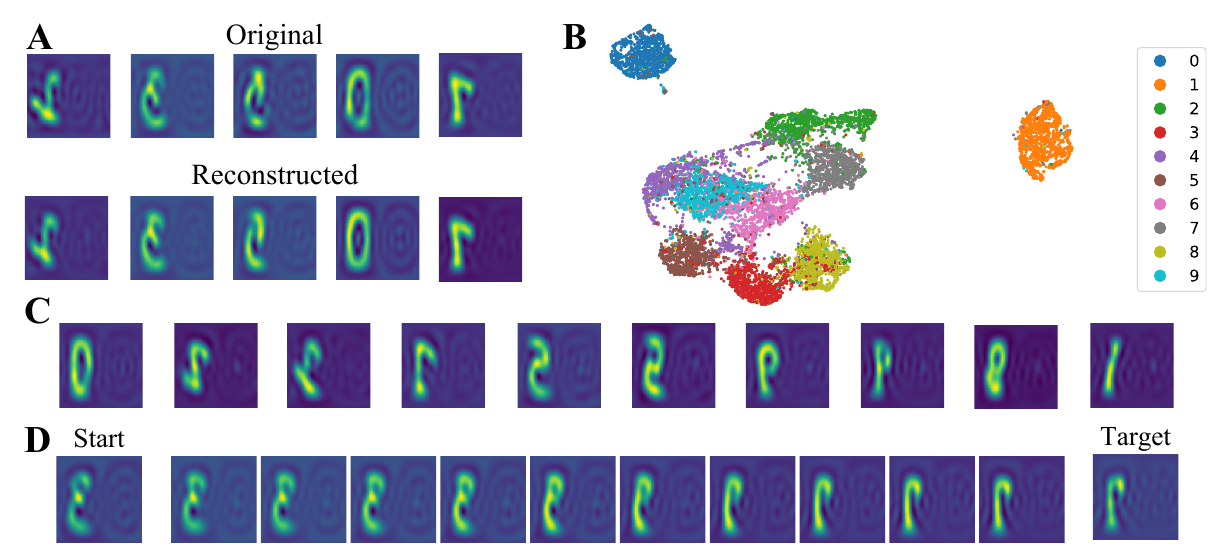}
    \caption{{\bf \small H-VAE on MNIST-on-the-sphere.} Evaluation on rotated digits for an H-VAE trained on non-rotated digits with $z = 16$. \textbf{(A)} Original and reconstructed images in the canonical frame after inverse transform from Fourier space. The images are projected onto a plane. Distortions at the edges and flipping are side-effects of the projection. \textbf{(B)} visualization of the latent space via 2D UMAP~\cite{mcinnes_umap_2020}. Data points are colored by digit identity. \textbf{(C)} Cherry-picked images generated by feeding the decoder invariant embeddings sampled from the prior distribution and the canonical frame. \textbf{(D)} Example image trajectory by linearly interpolating through the learned invariant latent space.
    Interpolated invariant embeddings are fed to the decoder alongside the canonical frame. MNIST-on-the-sphere dataset is created by projecting data from the planar MNIST on a discrete unit sphere, using the Driscoll-Healey (DH) method with a bandwidth (bw) of 30~\cite{cohen_spherical_2018}.
    }
    \label{fig:mnist_viz}
\end{figure*}

\noindent {\bf Architecture of a Clebsch-Gordan block.} 
Each Clebsch-Gordan block (CG bl.) consists of a trained linear layer ({\em Linearity} (Lin)), an efficient tensor product (ETP) to inject nonlinearity in the network, and normalization steps by ({\em Batch Norm} (BN)) and ({\em Signal Norm} (SN)) to respectively speed-up convergence and stabilize training.\\

\noindent {\em Linearity (Lin).} A linear layer acts on steerable tensors by learning degree-specific linear operations. Linear layers are trained in that we learn weight matrices specific to each degree $\ell$, and use them to map across degree-$\ell$ feature spaces by learning linear combinations of degree-$\ell$ features in the input tensor. Specifically, let us consider a vector $\bf h_\ell$, containing features of the same degree $\ell$. We train the network to learn the weight matrix $\bf W_\ell$ to map $\bf h_\ell$ to $\bf h'_\ell$ between the network's layers, i.e., $\bf h'_\ell=\bf W_\ell \bf h_\ell$  (see Section~\ref{sec:linearity} for details).\\

\noindent {\em Nonlinearity with efficient tensor product (ETP).}  
One key feature of neural networks is applying nonlinear activations, which enable a network to approximately model complex and non-linear phenomena. As noted above, we use the Clebsch-Gordan tensor product to inject rotationally equivariant nonlinearities in the network. Specifically, within a Clebsch-Gordan block the output of a linear layer is acted upon by bi-linear CG tensor product, as was originally prescribed by in ref.~\cite{kondor_clebsch-gordan_2018} for SO(3)-equivariant convolutional neural networks. This bilinear operation enables information flow between features of different degrees, which is necessary for constructing expressive models, and  for transferring higher-$\ell$ information to $\ell = 0$ in H-(V)AE's invariant encoder, and back in the decoder.
 
To significantly reduce the computational and memory costs of the tensor products, we  perform {\em efficient tensor products} (ETPs) by leveraging some of the modifications proposed in ref.~\cite{cobb_efficient_2021}. Specifically, we compute tensor products channel-wise, i.e., only between features belonging to the same channel, and we limit the connections between features of different degrees. We found these modifications to be necessary to efficiently work with data encoded in large number of channels $C$ and with large maximum degree $L$; see Section~\ref{sec:ETP_details} for details, and Table~\ref{table:tp_ablation} for an ablation study showing the improvement in parameter efficiency provided by the ETP.

\noindent {\em Batch and signal Norm.} We normalize intermediate tensor representations degree-wise and channel-wise by the batch-averaged norms of the features, as initially proposed in ref.~\cite{kondor_clebsch-gordan_2018}; see Figure~\ref{fig:model}B, Section~\ref{sec:batch_norm}, and Fig.~S2 for details. We found using batch norm alone often caused activations to explode in the decoder during evaluation. Thus, we introduce Signal Norm, whereby we divide each steerable tensor by the square root of its \textit{total} norm, defined as the sum of the norms of each of the tensor's features, and apply a degree-specific affine transformation for added flexibility; see Section~\ref{sec:signal_norm} for mathematical details. Signal Norm can be seen as a form of the classic Layer Normalization that respects SO(3) equivariance~\cite{ba_layer_2016}.

\section{Results}
\subsection{Rotated MNIST on the sphere}
We extensively test the performance of  H-(V)AE on the MNIST-on-the-sphere dataset~\cite{deng_mnist_2012}. Following ref.~\cite{cohen_spherical_2018}, we project the MNIST dataset, which includes images of handwritten numbers,  onto the lower hemisphere of a discrete unit sphere. We consider two variants of training/test set splits, NR/R and R/R, differing in whether the training/test images have been randomly rotated (R) or not (NR). For each dataset, we map the images to steerable tensors via the Zernike Fourier Transform (ZFT) in eq.~\ref{eq:ZFT} and train models with different different sizes of latent spaces and model types (AE vs. VAE). In all cases the model architecture follows from Fig.~\ref{fig:model}C; see Section~\ref{sec:mnist_details} for details.  

We use the metric {\em Cosine loss} to measure a model's reconstruction ability. Cosine loss is a normalized dot product generalized to operate on pairs of steerable tensors (akin to cosine similarity), and modified to be interpreted as a loss. Importantly, unlike MSE, Cosine loss is dimensionless, and therefore, comparable across different datasets and encodings of data in tensors of different sizes (network hyperparameters); see Section~\ref{sec:cosine_loss_appendix} for details.  

All trained models achieve very low reconstruction Cosine loss (Table~\ref{table:all_scores}) with no significant difference between training modes, indicating that the models successfully leverage SO(3)-equivariance to generalize to unseen orientations. Predictably, AE models have lower reconstruction loss than VAE models (since they do not need to find a trade-off between reconstruction error and KL divergence, Eq.~\ref{eqn:training_objective}), and so do models with a larger latent space. Nonetheless, H-VAE achieves reliable reconstructions, as shown in Figure~\ref{fig:mnist_viz}A and Table~\ref{table:all_scores}.

\newcommand{\frameimgsize}{0.18}
\begin{figure*}[t!]
    \centering
    \resizebox{0.83\textwidth}{!}{
    \begin{tabular}{c | c | c | c | c}
    & AE - NR/R & AE - R/R & VAE - NR/R & VAE - R/R
    \\
    \hline
    \newlinecell[c]{Trained with\\all digits}
    &
    \includegraphics[valign=m,width=\frameimgsize\textwidth]{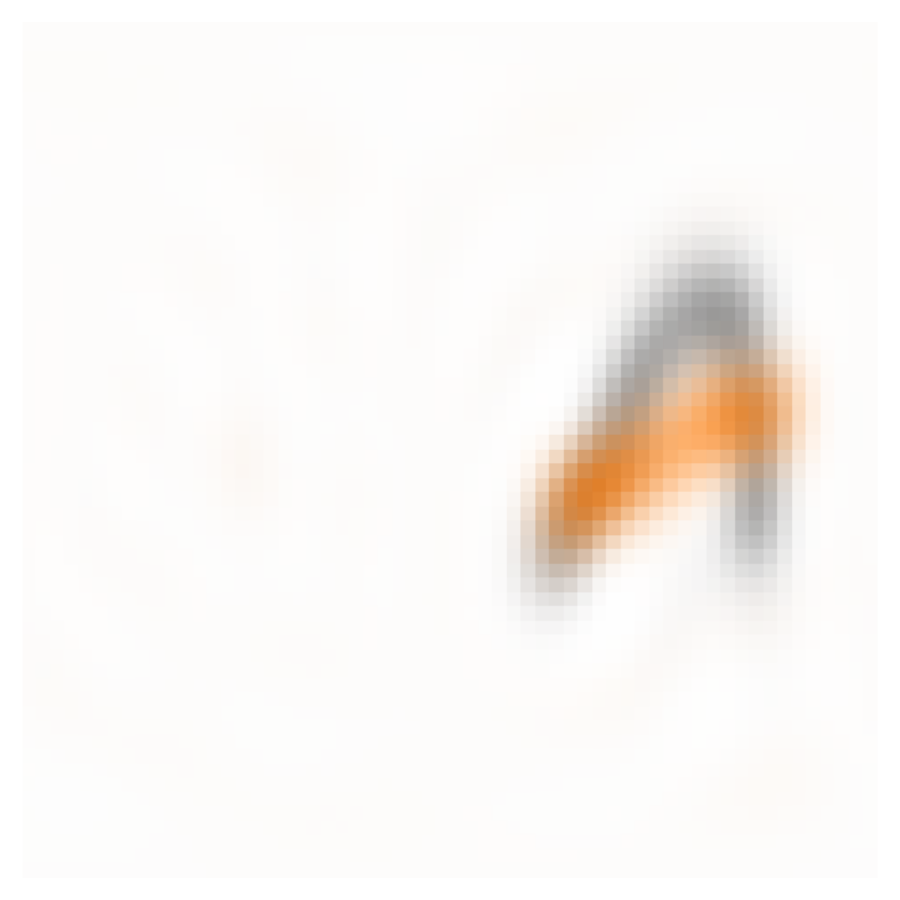}
    &
    \includegraphics[valign=m,width=\frameimgsize\textwidth]{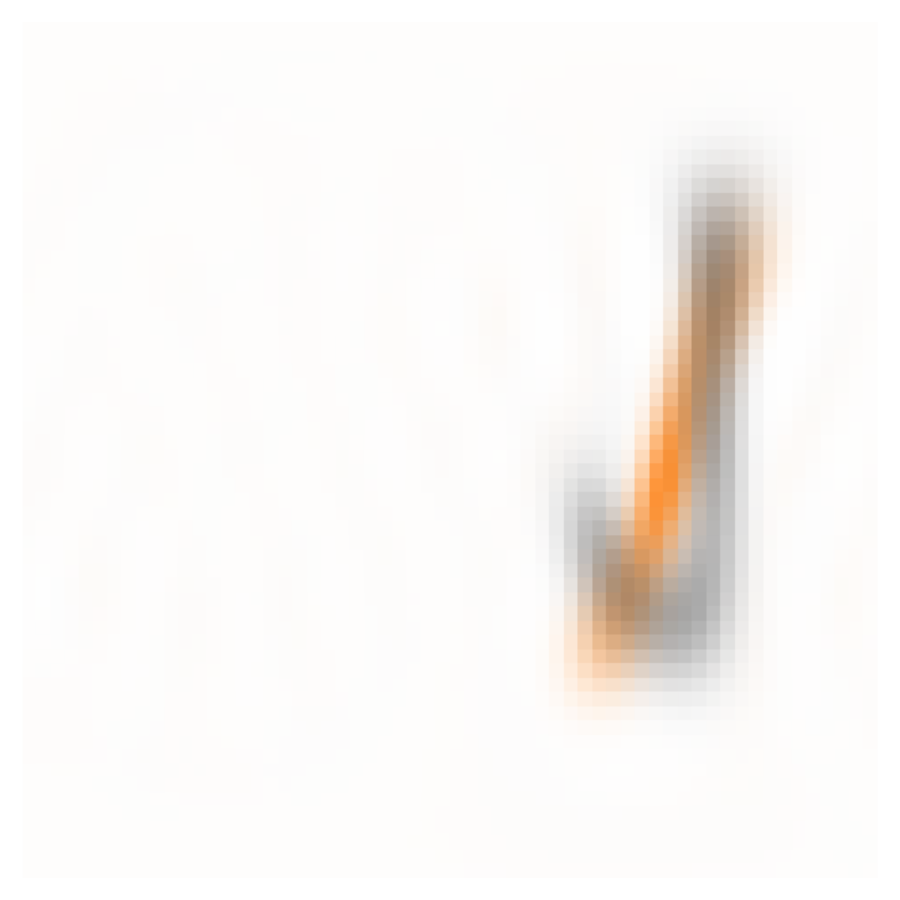}
    &
    \includegraphics[valign=m,width=\frameimgsize\textwidth]{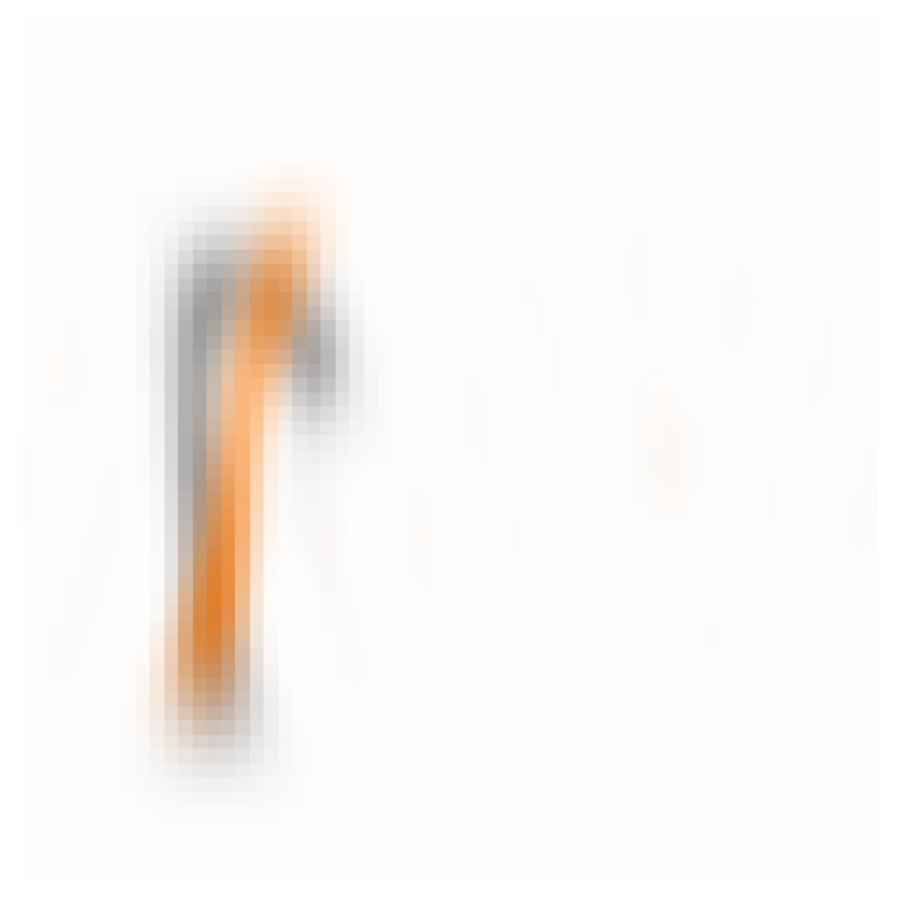}
    &
    \includegraphics[valign=m,width=\frameimgsize\textwidth]{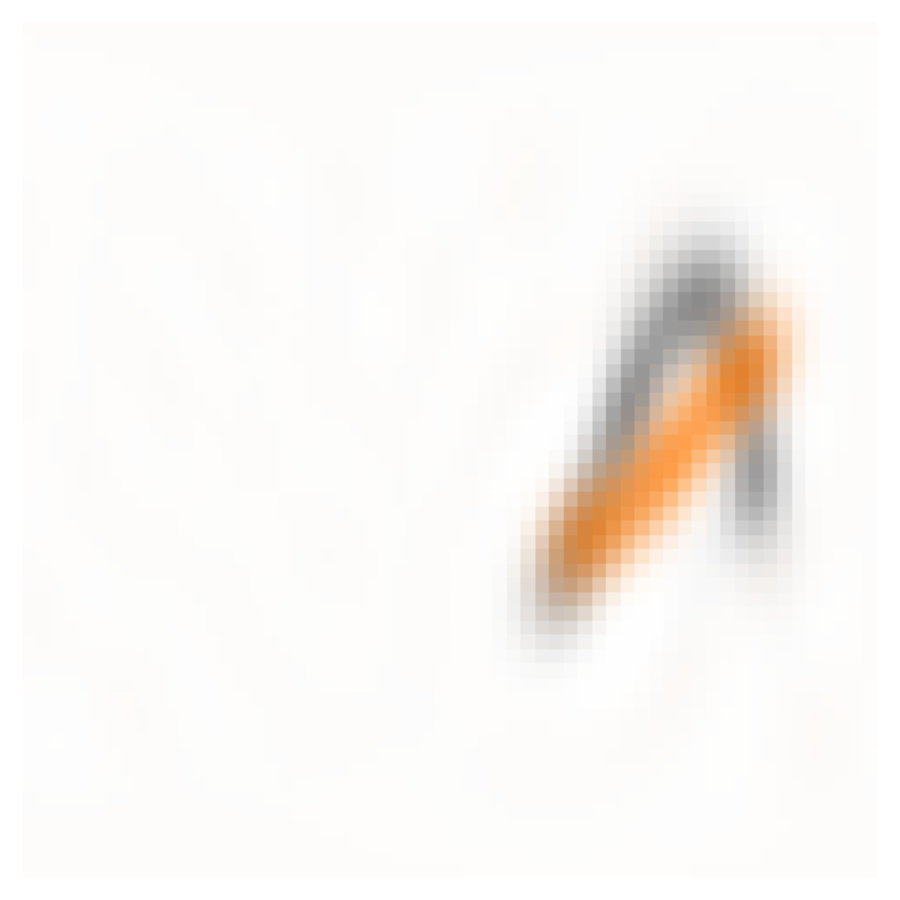}
    \\
    \hline
    \newlinecell[c]{Trained with\\1s and 7s only}
    &
    \includegraphics[valign=m,width=\frameimgsize\textwidth]{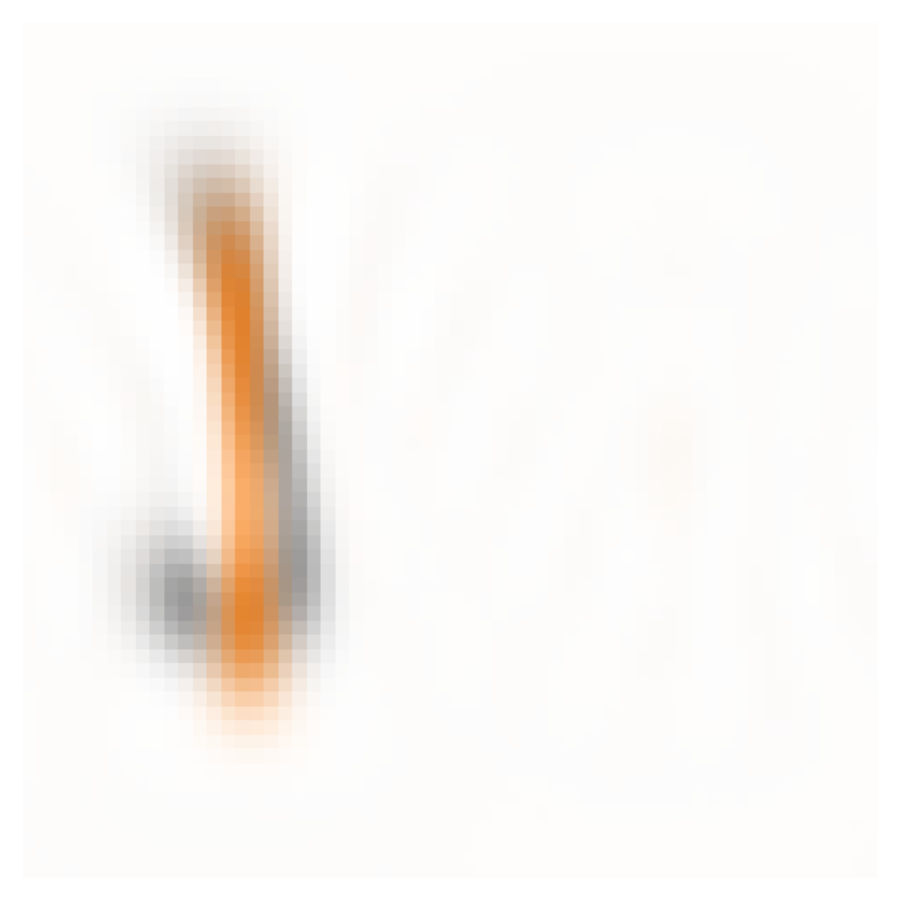}
    &
    \includegraphics[valign=m,width=\frameimgsize\textwidth]{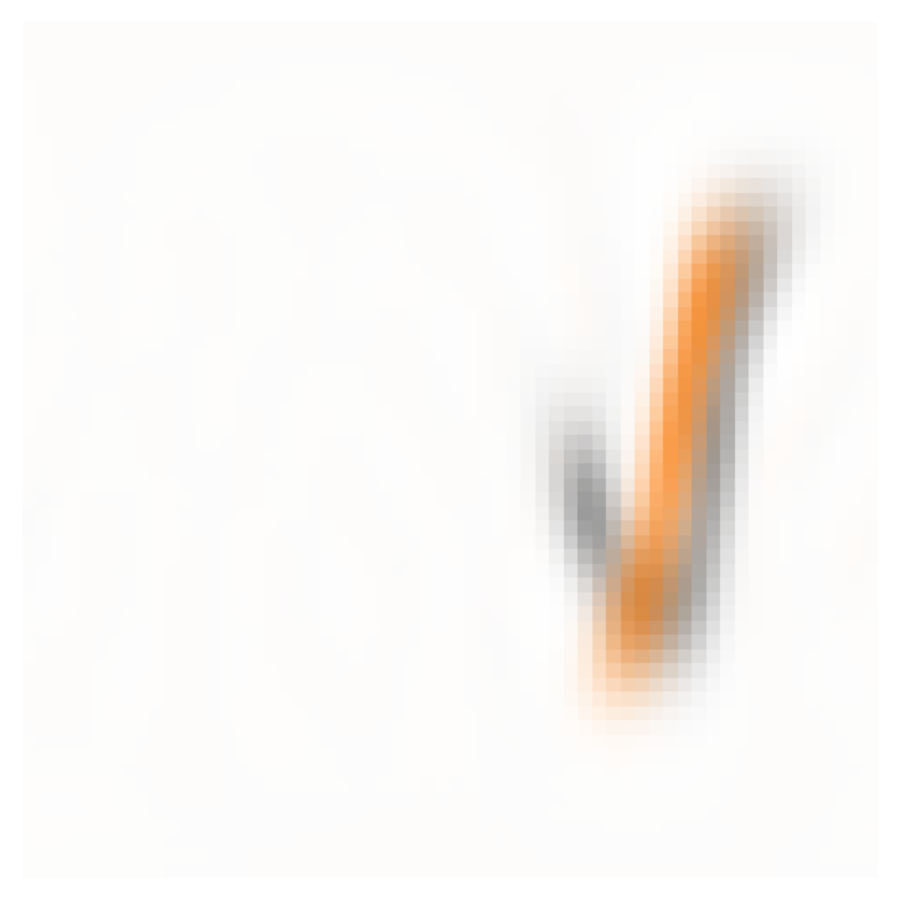}
    &
    \includegraphics[valign=m,width=\frameimgsize\textwidth]{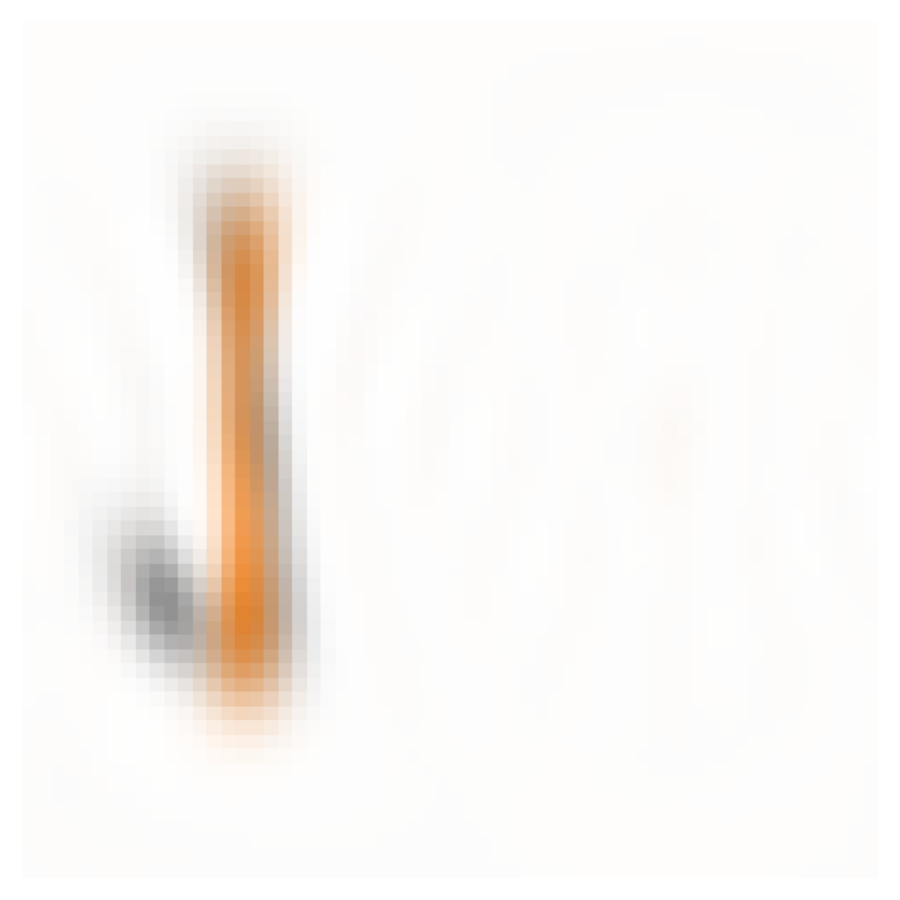}
    &
    \includegraphics[valign=m,width=\frameimgsize\textwidth]{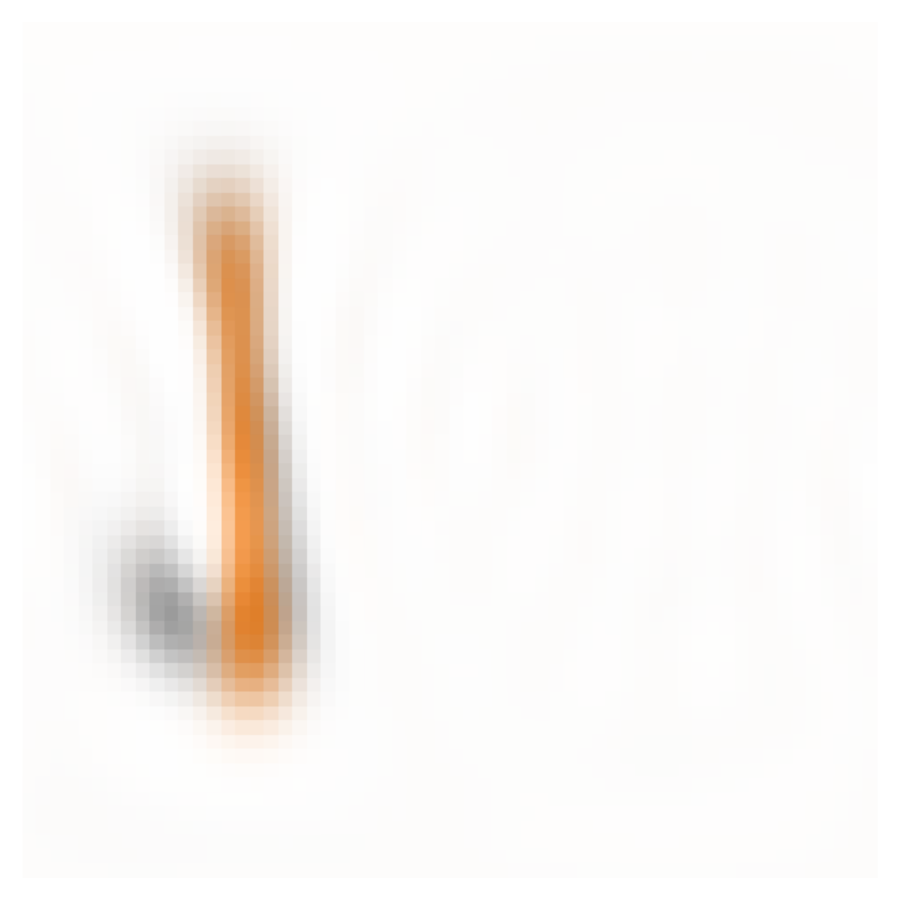}
    \end{tabular}}
    \caption{\textbf{H-(V)AE implicitly learns to maximally overlap training images on MNIST-on-the-sphere.}
    For each of the 4 models with $z = 16$, we train a version using only images containing 1s and 7s. For each of the resulting 8 models, we visualize the sum of training images of digits 1 and 7, when rotated to the canonical frame. We compute the sums of images with the same digit, and overlay them with different colors for ease of visualization. We test the hypothesis as whether H-(V)AE learns frames that align the training images such that they maximally overlap; we do so in two ways.\\
    \textbf{First:} if the hypothesis were true, all canonical images of the same digit should maximally or near-maximally overlap - since they have very similar shape - and thus, their overlays would look like a ``smooth" version of that digit. Indeed, we find this statement to be true for all models irrespective of their training strategy.\\
    \textbf{Second:} we consider the alignment of images of different digits. We take 1s and 7s as examples given their similarity in shape. If the hypothesis were true, models trained with only 1s and 7s should align canonical 1s along the long side of canonical 7s; indeed we find this to be consistently the case. The same alignment between 1s and 7s, however, does not necessarily hold for models trained with all digits. This is because maximizing overlap across a set of diverse shapes does not necessarily maximize the overlap within any independent pair of such shapes. Indeed, we find that canonical 1s and canonical 7s do not overlap optimally with each other for models trained with all digits.\\
    We note that these tests do not provide a formal proof, but rather empirical evidence of the characteristics of frames learned by H-(V)AE on the MNIST-on-the-sphere task.}
    \label{fig:canonical_frames_1,7}
\end{figure*}

Interpretability of the latent space is also an important feature of a model.  All 8 models produce an invariant latent space that naturally clusters by digit identity. We show this qualitatively for one of the models in Fig.~\ref{fig:mnist_viz}B, and quantitatively by clustering the data via K-Means with 10 centroids and computing standard clustering metrics of Purity~\cite{aldenderfer_cluster_1984} and V-measure~\cite{rosenberg_v-measure_2007} in Table~\ref{table:all_scores}. Crucially, the built-in SO(3)-equivariance enables models trained on the non-rotated images to seamlessly generalize to images that have been randomly rotated, as seen by the equivalent performance between models trained and evaluated on the NR/R and R/R splits (Table~\ref{table:all_scores}). 

\begin{table*}
\begin{center}
\resizebox{\textwidth}{!}{
\begin{tabular}{l | l l c c | c | c c | c | c c c c c}
    \bf Dataset & \bf Type & \bf Method & \bf z & \bf bw & \bf \newlinecell{\bf Cosine} & \bf Purity & \bf V-meas. & \bf \newlinecell{Class.\\Acc.} & \bf P@N & \bf R@N & \bf F1@N & \bf mAP & \bf NDCG \\
    \hline
    \hline
    \multirow{11}{*}{MNIST}
    & \multirow{1}{*}{Supervised}
    & \cite{cobb_efficient_2021} NR/R      & - & 30 & - & - & - & 0.993 & - & - & - & - & - \\
    \cline{2-14}
    & \multirow{9}{*}{Unsupervised}
    & \cite{lohit_rotation-invariant_2020} NR/R      & 120 & 30 &      - & 0.40 & 0.35  & \textbf{0.894} & - & - & - & - & - \\
    & & H-AE NR/R (Ours)     & 120 & 30 & 0.017 & 0.62 & 0.48 & 0.877 & - & - & - & - & - \\
    & & H-AE R/R (Ours)      & 120 & 30 & 0.018 & 0.51 & 0.41 & 0.881 & - & - & - & - & - \\
    & & H-AE NR/R (Ours)     &  16 & 30 & 0.025 & 0.62 & 0.51 & 0.820 & - & - & - & - & - \\
    & & H-AE R/R (Ours)      &  16 & 30 & 0.024 & 0.65 & 0.52 & 0.833 & - & - & - & - & - \\
    & & H-VAE NR/R (Ours)    & 120 & 30 & 0.037 & 0.70 & \textbf{0.59} & 0.883 & - & - & - & - & - \\
    & & H-VAE R/R (Ours)     & 120 & 30 & 0.037 & 0.65 & 0.53 & 0.884 & - & - & - & - & - \\
    & & H-VAE NR/R (Ours)    &  16 & 30 & 0.057 & 0.67 & 0.54 & 0.812 & - & - & - & - & - \\
    & & H-VAE R/R (Ours)     &  16 & 30 & 0.055 & \textbf{0.72} & 0.57 & 0.830 & - & - & - & - & - \\
    \hline
    \hline
    \multirow{5}{*}{Shrec17}
    & \multirow{2}{*}{Supervised}
    & \cite{esteves_spin-weighted_2020}     & - & 128 & - & - & - & - & 0.717 & 0.737 &   -   & 0.685 &   -   \\
    & & \cite{cobb_efficient_2021}      & - & 128 & - & - & - & - & 0.719 & 0.710 & 0.708 & 0.679 & 0.758 \\
    \cline{2-14}
    & \multirow{3}{*}{Unsupervised}
    & \cite{lohit_rotation-invariant_2020}        & 120 & 30 &     - & 0.41 & 0.34 & 0.578 & 0.351 & 0.361 & 0.335 & 0.215 & 0.345 \\
    & & H-AE (Ours)       &  40 & 90 & 0.130 & 0.50 & 0.41 & \textbf{0.654} & \textbf{0.548} & \textbf{0.569} & \textbf{0.545} & \textbf{0.500} & \textbf{0.597} \\
    & & H-VAE (Ours)      &  40 & 90 & 0.151 & \textbf{0.52} & \textbf{0.42} & 0.631 & 0.512 & 0.537 & 0.512 & 0.463 & 0.568 \\
\end{tabular}}
\end{center}
\caption{{\bf Performance metrics on MNIST-on-the-sphere and Shrec17}. Reconstruction Cosine loss, clustering metrics (Purity and V-measure), classification accuracy in the latent space using a linear classifier, and retrieval metrics (Shrec17 only) are shown. 
For MNIST, we follow  ref.~\cite{cohen_spherical_2018}  to create the   MNIST-on-the-sphere dataset by   projecting data from the planar MNIST on a discrete unit sphere, using the Driscoll-Healey  method with a bandwidth (bw) of 30. We then map the images to steerable tensors via the Zernike Fourier Transform (ZFT) with $L = 10$, and a constant radial function $R^{n}_{\ell} = 1$, resulting in tensors with 121 coefficients. We train 8 models with different sizes of latent spaces $z$  (16 vs. 120) and model types (AE vs. VAE). 
For Shrec17, we follow ref.~\cite{cohen_spherical_2018} and project surface information from each model onto an enclosing Driscoll-Healey spherical grid with a bandwidth of 90 via a ray-casting scheme, generating spherical images with 6 channels. We then apply the ZFT with $L = 14$ and a constant radial function $R^{n}_{\ell} = 1$ to each channel individually, resulting in a tensor with 1350 coefficients.
We only report scores presented in the corresponding papers, and only the best-performing supervised method from the literature; see Fig.~S5 for visualization of the latent embeddings for the Shrec17 dataset.
}
\label{table:all_scores}
\end{table*}

All trained models achieve much better clustering metrics than Rot-Inv AE~\cite{lohit_rotation-invariant_2020}, with the VAE models consistently outperforming the AE models. We also train a linear classifier (LC) to predict digit identity from invariant latent space descriptors, achieving comparable accuracy to Rot-Inv AE with the same latent space size. We do not observe any difference between VAE and AE models in terms of classification accuracy. Using a K-nearest neighbor (KNN) classifier instead of LC further improves performance (Table~\ref{table:mnist_scores_knn}). 

As H-VAE is a generative model, we generate random spherical images by sampling invariant latent embeddings from the prior distribution, and observing diversity in digit type and style (Fig.~\ref{fig:mnist_viz}C and Fig.~S3). We further assess the quality of the invariant latent space by generating images via linear interpolation of the invariant embeddings associated with two test images. The interpolated images present spatially consistent transitions (Fig.~\ref{fig:mnist_viz}D and Fig.~S4), which is a sign of a semantically well-structured latent space.

To understand the meaning of the learned frames, we ask ourselves what the output of the decoder looks like if the frame is held constant; for simplicity, we set it equal to the $3 \times 3$ identity matrix. We find that the reconstructed elements tend to be aligned with each other and hypothesize that the model is implicitly learning to maximize the overlap between training elements, providing empirical evidence in Figure~\ref{fig:canonical_frames_1,7}. We call this frame the \textbf{canonical} frame. We note that it is possible to rotate original elements to the canonical frame thanks to the equivalence between the frame we learn and the rotation matrix within our implementation; in fact in our experiments, when visualizing reconstructed or sampled MNIST images, we first rotate them to the canonical frame for ease of visualization.

\subsection{Shrec17}
\begin{figure*}[t!]
    \includegraphics[width=\textwidth]{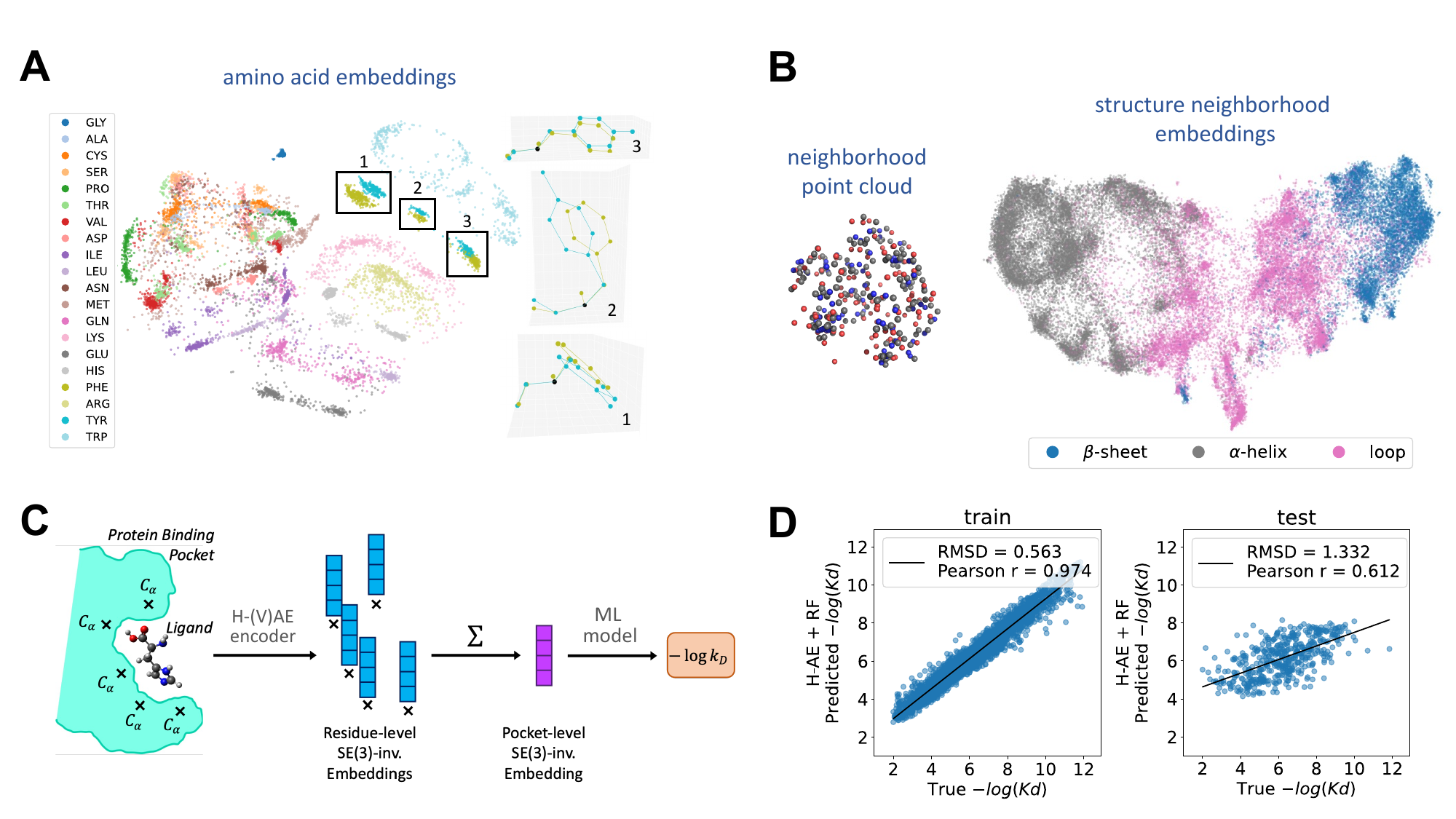}\\
    \caption{{\bf Structural embeddings to predict protein-ligand binding affinities with H-(V)AE.} 
    \textbf{(A)}     H-VAE was trained to reconstruct the Fourier representation of 3D atomic point clouds representing amino acids (colors). The invariant latent space clusters by amino acid conformations. 
    The highlighted clusters for PHE and TYR contain residue pairs with similar conformations; TYR and PHE differ by one oxygen at the end of their benzene rings. We compare conformations by plotting each residue in the standard backbone frame (right); $x$ and $y$ axes are set by the orthonormalized C$\alpha$-N and C$\alpha$-C vectors, and $z$ axis is their cross product. For this plot, 1,000 amino acids were used as training data, with network parameters: $\beta = 0.025$ and $z = 2$.     
    {\bf (B)} Left: An example protein neighborhood (point cloud of atoms) of 10 \AA\, around a central residue, used to train the H-AE models, is shown. Right: 2D UMAP visualization of the 128-dimensional invariant latent space learned by H-AE trained on the protein structure neighborhoods with $L=6$ can separate neighborhoods by the secondary structure of their focal amino acid (colors). Each point represents a neighborhood; see Section~\ref{sec:protein_neighborhoods_details} for details on the network architecture and training procedure.
  {\bf (C)}  We use H-AE to extract the residue-level SO(3)-invariant embeddings in the binding pocket of a protein-ligand structure complex (data from PDBbind~\cite{su_comparative_2019}). We then sum over these embeddings to form an SE(3)-invariant pocket embedding that is used as an input to a standard Machine Learning model to predict the  binding affinity between the protein and the ligand.
 {\bf (D)} The predictions on the protein-ligand binding affinities from (C) is shown against the true values  for the training (left) and the test (right) sets. We use the data split provided by ATOM3D~\cite{townshend_atom3d_2021}, which devises training and test sets respectively containing 3507, and 490 protein-ligand complexes, with maximum 30\% sequence similarly between training and test proteins; see Table~\ref{table:lba_results} for a comparison against state-of-the-art methods.}
    \label{fig:lba_diagram}
\end{figure*}

The Shrec17 dataset consists of 51k colorless 3D models belonging to 55 object classes, with a 70/10/20 train/valid/test split~\cite{noauthor_shrec_nodate}. We use the variant of the dataset in which each object is randomly rotated. Converting 3D shapes into spherical images preserves topological surface information, while significantly simplifying the representation. We follow ref.~\cite{cohen_spherical_2018} 
and project surface information for each image onto an enclosing spherical grid via a ray-casting scheme and apply ZFT (eq.~\ref{eq:ZFT}) on these transformed images.

We train an AE and a VAE model on ZFT transformed data (Section~\ref{sec:shrec17_details}); Fig.~S5 shows the resulting latent embeddings for both H-AE and H-VAE on this dataset.  Similarly to the MNIST dataset, we compute cosine loss, clustering metrics and classification accuracy via a linear classifier. We also compute the standard Shrec17 object retrieval metrics via the latent space linear classifier's predictions (see~\cite{noauthor_shrec_nodate} for a description of the retrieval metrics). H-AE achieves the best classification and retrieval results for autoencoder-based models, and is competitive with supervised models despite the lower grid bandwidth and the small latent space (Table~\ref{table:all_scores}). Using KNN classification instead of a linear classifier further improves performance (Table~\ref{table:shrec17_evaluation_knn}). H-VAE achieves slightly worse classification results but better clustering metrics compared to H-AE. While reconstruction loss is low, there is still significant margin of improvement.

\subsection{Structural  embeddings of amino acid neighborhoods to predict function}
Here, we provide a strong use case for H-(V)AE in structural biology. 
Specifically, we learn expressive embeddings for amino acids neighborhoods within protein structures that can be used to learn protein function.\\

\noindent {\bf \small Embeddings of amino acid conformations.}  As a first, propedeutic step, we train H-(V)AE to reconstruct the 3D structure of individual amino acids, represented as atomic point clouds, extracted from protein structures in the Protein Data Bank (PDB)~\cite{berman_protein_2000}. Residues of the same type have different conformations and naturally have noisy coordinates, making this problem a natural benchmark for our rotationally equivariant method. 

We represent an amino acid by atom-type-specific clouds (C, O, N and S; we exclude H) centered at the residue's C$\alpha$ and compute the ZFT (eq.~\ref{eq:ZFT}) with $L = 4$ and $N = 20$ within a radius of $10$\AA\, from the residue's C$\alpha$, and concatenate features of the same degree $\ell$, resulting in a tensor with 940 coefficients. We train several H-AE and H-VAE with different architectures, but all with the latent space sizes $z=2$; see Sec.~\ref{sec:toy_aminoacids_details} for details.\\

We consistently find that the latent space clusters by amino acid conformations (Figure~\ref{fig:lba_diagram}A), with sharper cluster separations as more training data is added (Figure~S6 and~S7). We find that test reconstruction loss decreases with more training data but the reconstruction is accurate even with little training data (from 0.153 Cosine loss with 400 training residues to 0.034 with 20,000);  Table~\ref{table:toy_aas_data_ablation_z_eq_2}. A similar trend is observed for KNN-based classification accuracy of residues (from 0.842 with 400 training residues to 0.972 with 20,000); (Table~\ref{table:toy_aas_data_ablation_z_eq_2}). Notably, an untrained model, while achieving random reconstruction loss, still produces an informative invariant latent space (0.629 residue type accuracy), suggesting that the forced SO(3)-invariance grants a ``warm start" to the encoder. We do not find significant improvements in latent space classification by training with a variational objective, and present ablation results in Table~\ref{table:toy_aas_kld_ablation_z_eq_2}.\\

\noindent {\bf  \small Embeddings of amino acid structure neighborhoods.}  The structural neighborhood surrounding an amino acid provides a context for its function (e.g., whether it takes part in interaction with other proteins at a protein-protein interface or not). Indeed, our previous work has shown that supervised learning algorithms can accurately classify focal amino acids based on the composition of their surrounding neighborhoods~\cite{pun_learning_2022}. Here, we  train H-(V)AE to reconstruct residue-specific protein \textit{neighborhoods} - which we define as the point clouds of atoms within 10\AA\, of a residue's C$\alpha$ - across the protein universe (Fig.~\ref{fig:lba_diagram}A).  We extract these protein neighborhoods from all proteins in ProteinNet~\cite{alquraishi_proteinnet_2019}. We then construct each neighborhood's Fourier representation by computing the ZFT (Eq.~\ref{eq:ZFT}) over the point clouds associated with each atom type within the neighrbohood (C, N, O and S) and concatenating atom-specific features of the same degree $\ell$.

We train several H-AE models with varying architectures with different maximum spherical degree $L$ and latent space sizes $z$ (see details in Sec.~\ref{sec:protein_neighborhoods_details}); note that we do not experiment with variational models for this task. H-AE shows strong reconstruction ability, but its accuracy worsens with smaller latent space sizes and higher maximum spherical degree $L$ (Table~\ref{table:mse_and_cosine}). Notably, the learned latent space is structured according to the secondary structure of the neighborhood's focal amino acid (Fig.~\ref{fig:lba_diagram}B).\\

\begin{table}[t]
\begin{center}
\resizebox{0.48\textwidth}{!}{
\begin{tabular}{l | c c c c}
     & \multicolumn{3}{c}{\newlinecell{\textbf{Ligand Binding Affinity}\\\textbf{30\% Similarity}}} \\
    {Model} & {RMSD $\downarrow$} & {Pearson's r $\uparrow$} & {Spearman's r $\uparrow$} \\
    \hline
    DeepDTA &                  $1.565$           & $0.573$           & $0.574$           \\ 
    3DCNN &                    $1.414 \pm 0.021$ & $0.550$           & $0.553$           \\
    GNN &                      $1.570 \pm 0.025$ & $0.545$           & $0.533$           \\
    MaSIF &                    $1.484 \pm 0.018$ & $0.467 \pm 0.020$ & $0.455 \pm 0.014$ \\
    EGNN &                     $1.492 \pm 0.012$ & $0.489 \pm 0.017$ & $0.472 \pm 0.008$ \\
    GBPNet &                   $1.405 \pm 0.009$ & $0.561$           & $0.557$           \\
    EGNN + PLM &               $1.403 \pm 0.013$ & $0.565 \pm 0.016$ & $0.544 \pm 0.005$ \\
    ProtMD  &                  \underline{$1.367 \pm 0.014$} & \underline{$0.601 \pm 0.036$} & \underline{$0.587 \pm 0.042$}\\
    \hline
    H-AE + L.R. & $1.397 \pm 0.019$ & $0.560 \pm 0.017$ & $0.568 \pm 0.018$ \\
    H-AE + R.F. &     $\mathbf{1.332 \pm 0.012}$ & $\mathbf{0.612 \pm 0.009}$ & $\mathbf{0.619 \pm 0.009}$
\end{tabular}
}
\end{center}
\caption{{\bf Results on the Ligand Binding Affinity task.} Prediction accuracies using H-AE embeddings with  Linear Regression (L.R.), and  Random Forest (R.F.) Regression are benchmarked against other methods. We choose the H-AE model with best RMSD on validation split, which is the model with $L = 6$ and $z = 128$ for both Linear Regression and Random Forest. For each set of predictions, we use an ensemble of 10 Regressors as we noted a small but consistent improvement in performance. Best scores are in \textbf{bold} and second-best scores are \underline{underlined}. H-AE+R.F. delivers state-of-the-art predictions. Methods are ordered by date of release; see Table~\ref{table:lba_results_appendix} for more details.}
\label{table:lba_results}
\end{table}

\noindent {\bf \small Predicting protein-Ligand Binding Affinity (LBA).}  The binding interaction between proteins and ligands should be primarily determined  by the composition of the protein's binding pocket in complex with the ligand. Therefore, we hypothesized that the inferred protein structure  embeddings  (Fig.~\ref{fig:lba_diagram}B) should contain information about protein-ligand binding interactions, if the neighborhood is defined along the binding pocket and contains atoms from the ligand. To test this hypothesis, we  follow the pipeline in Fig.~\ref{fig:lba_diagram}C. Specifically,  given a protein-ligand structure complex, we identify residues in the binding pocket (i.e., residues with C-$\alpha$ within 10\AA\, of the ligand) and extract their structure neighborhoods, which include atoms from both the protein and the ligand. We then pass the residue-centered neighborhoods of the binding pocket through a trained H-AE's encoder to extract their rotationally-invariant embeddings. We highlight that, since the neighborhood centers are well-defined at the residues' C$\alpha$'s, the embeddings are not only rotationally invariant about their center, but also translationally invariant with the respect to translations of the whole system, i.e., they are effectively SE(3)-invariant. 

Since protein-ligand binding affinity is an extensive quantity in the number of interacting residues, we construct a pocket embedding by summing over residue-level embeddings; the resulting pocket embedding is SE(3)-invariant, reflecting the natural symmetry of the LBA task. We use these pocket embeddings as feature vectors to train simple Machine Learning models to predict protein-ligand binding affinities.

To test the performance of our method, we use the LBA dataset in ATOM3D~\cite{townshend_atom3d_2021} that provides the PDB structure of the protein-ligand complex together with   either the measured dissociation constant $K_d$ or the inhibition constant $K_i$; see Sec.~\ref{sec:lba_details} for further details. To map between the learned pocket embeddings to the log dissociation (or inhibition) constants we train both a simple Linear and a Random Forest Regressor on a training sets provided by ATOM3D.  Fig.~\ref{fig:lba_diagram}D shows the performance of the model with the Random Forest Regressor and Table~\ref{table:lba_results} provide a detailed benchmark of our methods against prior approaches.  The Linear model achieves competitive results, whereas the Random Forest Regressor achieves state-of-the-art. 

These results demonstrate the utility of unsupervised learning for residue-level protein structure representations in predicting complex protein functions. Most of the competing structure-based methods for LBA (Table~\ref{table:lba_results}) learn complex graph-based functions on top of simple atomic representation, whereas our method uses simpler machine learning models over rich residue-level representations. Notably, ProtMD (the method competitive to ours) performs a pre-training scheme using expensive molecular dynamics simulations that informs the model about conformational flexibility, information that our method does not have access to. Given the computational cost of training complex atom-level graph-based models, our residue-based approach can offer a more viable alternative for modeling large protein interfaces.

\section{Discussion}
In this work, we have developed the first end-to-end SO(3)-equivariant unsupervised algorithm, termed Holographic (Variational) Auto-Encoder (H-(V)AE), suitable for data distributed in three dimensions around a given central point. The model learns an invariant embedding describing the data in a ``canonical" orientation alongside an equivariant frame describing the data's original orientation relative to the canonical one.

Prior works have attempted to learn representations that are invariant to certain transformations. For example, in refs.~\cite{shu_deforming_2018, koneripalli_rate-invariant_2020} general ``shape" embeddings are learned by characterizing a separate ``deformation" embedding. However, these networks are not explicitly equivariant to the transformations of interest. 

Others proposed to learn an exactly invariant embedding alongside an approximate (but not equivariant) group action to align the input and the reconstructed data. For example, {\em Mehr et al}~\cite{mehr_manifold_2018}  learns in quotient space by pooling together the latent encodings of copies of the data that have been transformed with sampled group actions, and back-propagate the minimum reconstruction loss between the decoded element and the transformed copies of the data. This approach is best suited for discrete and finite groups, for which it does not require approximations, and it is computationally expensive as it is akin to data augmentation. {\em Lohit et al}~\cite{lohit_rotation-invariant_2020} construct an SO(3)-invariant autoencoder for spherical signals by learning an invariant latent space and minimizing a loss which first finds the rotation that best aligns the true and reconstructed signals. Although this approach is effective for non-discrete data, it still manually imposes rotational invariance, and can only reconstruct signals up to a global rotation.  In contrast, H-(V)AE  is fully equivariant and only requires simple MSE for reconstruction of data in its original orientation.

A small body of work went beyond invariance to develop equivariant autoencoders. Several methods construct data and group-specific architectures to auto-encode data equivariantly, learning an equivariant representation in the process~\cite{hinton_transforming_2011, kosiorek_stacked_2019}. Others use supervision to extract class-invariant and class-equivariant representations~\cite{feige_invariant-equivariant_2022}. A recent theoretical work proposes to train an encoder that encodes elements into an invariant embedding and an equivariant group action, then using a standard decoder that uses the invariants to reconstruct the elements in a canonical form, and finally applying the learned group action to recover the data's original form~\cite{winter_unsupervised_2022}. Our method in SO(3) is closely related to this work, with the crucial differences that our network is end-to-end rotationally equivariant in that we use an equivariant decoder, and that we learn to reconstruct the Fourier encoding of the data. A more detailed comparison of the two approaches and the benefits of our fully equivariant approach can be found in the Appendix (Section~\ref{sec:mlp_decoder_appendix})  and in Table~\ref{table:mlp_decoder_comparison}.

There is also a diverse body of literature on using Fourier transforms and and CG tensor products to construct representations of atomic systems that are invariant/equivariant to euclidean symmetries~\cite{drautz_atomic_2019, musil_physics-inspired_2021, uhrin_through_2021}, but  without reducing the dimensionality of the representations in a data-driven way.

H-(V)AE's learned embeddings are highly expressive. For example, we used the learned invariants to achieve state-of-the-art unsupervised clustering and classification results on various spherical image datasets. By making our model variational in its invariant latent space, we enhanced the quality of clustering and made the model generative. Our model is defined fully in spherical Fourier space, and thus, can reach a desired expressiveness without a need for excessive computational resources.

H-(V)AE also produces  rich residue-level representations of local neighborhoods in protein structures, which we use as embeddings for downstream structure-based tasks such as Ligand Binding Affinity prediction. Indeed, H-(V)AE representations paired with a simple Random Forest Regressor achieve state-of-the-art results on learning the binding affinity between proteins and small molecule ligands. 

More broadly, we expect that H-(V)AE can be used to extract rich, symmetry-aware features from local neighborhoods in spherical images and complex 3D objects, to be used in more complex downstream tasks that benefit from the symmetry constraints. For example, we expect our method can be leveraged for modeling diffusion MRI data, for which rotation-equivariant methods have recently proven to be highly beneficial~\cite{muller_rotation-equivariant_2021}. In structural biology, we expect our method to be useful for coarse-graining full-atom representations of protein structures - or other biomolecules - to facilitate structure-based predictions of function. For example, a large protein graph can be coarse-grained by substituting its full-atom representation with rich embeddings of local structural neighborhoods learned from an unsupervised model. With an added supervised step, these coarse-grained embeddings can be leveraged to predict complex protein functions, as we show for predicting ligand binding affinity. This approach is akin to using protein embeddings for sequence data, learned by language models, to inform (few-shot) predictions for protein function~\cite{swanson_predicting_2022}.

\section{Acknowledgements}
This work has been supported by the National Institutes of Health MIRA award (R35 GM142795), the CAREER award from the National Science Foundation (grant No: 2045054), the Royalty Research Fund from the University of Washington  (no. A153352), the Microsoft Azure award from the eScience institute at the University of Washington, and the Allen School Computer Science \& Engineering Research Fellowship from the Paul G. Allen School of Computer Science \& Engineering at the University of Washington. This work is also supported, in part, through the Departments of Physics and Computer Science and Engineering, and the College of Arts and Sciences at the University of Washington.


\newpage

\onecolumngrid
\appendix
%

\setcounter{table}{0}
\setcounter{figure}{0}

\renewcommand{\thetable}{S\arabic{table}}

\counterwithin{equation}{section}
\setcounter{equation}{0}
\setcounter{figure}{0}
\renewcommand{\thefigure}{S\arabic{figure}}

\section{Appendix}

\subsection{Expanded background on SO(3)-equivariance}
\label{sec:expanded_background}

\subsubsection{Invariance and Equivariance} Let $f : X \rightarrow Y$ be a function between two vector spaces and $\mathfrak{G}$ a group, where $\mathfrak{G}$ acts on $X$ and via representation $\mD_{X}$ and on $Y$ via representation $\mD_{Y}$. Then, $f$ is said to be  $\mathfrak{G}$-equivariant iff $f(\mD_{X}(\mathfrak{g})\vx) = \mD_{Y}(\mathfrak{g})f(\vx) \,, \forall \vx \in X \land \forall \mathfrak{g} \in \mathfrak{G}$. 
We note that invariance is a special case of equivariance where $\mD_{Y}(\mathfrak{g}) = \mI \,, \forall \mathfrak{g} \in \mathfrak{G}$.

\subsubsection{Group representations and the irreps of SO(3)} Groups can concretely act on distinct vector spaces via distinct group representations. Formally, a group representation defines a set of invertible matrices $\mD_{X}(\mathfrak{g})$ parameterized by group elements $\mathfrak{g} \in \mathfrak{G}$, which act on vector space $X$. As an example, two vector spaces that transform differently under the 3D rotation group SO(3)- and thus have different group representations - are scalars, which do not change under the action of SO(3), and 3D vectors, which rotate according to the familiar 3D rotation matrices. \\
A special kind of representation for any group are the irreducible representations (irreps) which are provably the ``smallest" nontrivial (i.e., they have no nontrivial group-invariant subspaces) representations. The irreps of a group are special because it can be proven that any finite-dimensional unitary group representation can be decomposed into a direct sum of irreps~\cite{tung_group_1985}. This applies to SO(3) as well, whose irreps are the Wigner-D matrices, which are $(2\ell + 1 \times 2\ell+1)$-dimensional matrices, each acting on a $(2\ell+1)$-dimensional vector space:
\begin{equation}
    \mD_{\ell}(\mathfrak{g}) \quad \text{for} \,\,\, \ell = 0, 1, 2, ...
\end{equation}
Therefore, every element of the SO(3) group acting on any vector space can be represented as a direct sum of Wigner-D matrices.

\subsubsection{Steerable features} A G-steerable vector is a vector $\vx \in {X}$ that under some transformation group $\mathfrak{G}$, transforms via matrix-vector multiplication $\mD_{X}(\mathfrak{g})\vx$; here, $\mD_{X}(\mathfrak{g})$ is the group representation of $\mathfrak{g} \in \mathfrak{G}$. For example, a vector in 3D Euclidean space is SO(3)-steerable since it rotates via matrix-vector multiplication using a rotation matrix.

However, we can generalize 3D rotations to arbitrary vector spaces by employing the irreps of SO(3). We start by defining a degree-$\ell$ feature as a vector that is SO(3)-steerable by the $\ell^{th}$ Wigner-D matrix $\mD_{\ell}$. Given the properties of irreps, we can represent any SO(3)-steerable vector as the direct sum of two or more independent degree-$\ell$ features, e.g. $\vx = \vx_{\ell_1} \oplus \vx_{\ell_2} \oplus ... \oplus \vx_{\ell_{n}}$. The resulting vector, which we refer to as a \textit{tensor} to indicate that it is composed of multiple individually-steerable vectors, is SO(3)-steerable via the direct sum of Wigner-D matrices of corresponding degrees. This tensor is a block-diagonal matrix with the Wigner-D matrices along the diagonal: $\mD(\mathfrak{g}) = \mD_{\ell_1}(\mathfrak{g}) \oplus \mD_{\ell_2}(\mathfrak{g}) \oplus ... \oplus \mD_{\ell_{n}}(\mathfrak{g})$.

\subsubsection{Spherical harmonics and the Spherical Fourier Transform}
Spherical harmonics are a class of functions that form a complete and orthonormal basis for functions $f(\theta,\phi)$ defined on a unit sphere; $\theta$ and $\phi$ are the azimuthal and the polar angles in the spherical coordinate system. In their complex form, spherical harmonics are defined as,
\begin{equation}
	Y_{\ell m}(\theta,\phi) = \sqrt{\frac{2n+1}{4\pi}\frac{(n-m)!}{(n+m)!}}e^{im\phi}P^m_\ell(\cos\theta)
	\label{eq:YLM}
\end{equation}
where $\ell$ is a non-negative integer ($0\leq \ell$) and $m$ is an integer within the interval $-\ell\leq m \leq \ell$. $P^m_\ell(\cos \theta)$ is the Legendre polynomial of degree $\ell$ and order $m$, which together with the complex exponential $e^{im\phi}$ define sinusoidal functions over the angles  $\theta$ and $\phi$ in the spherical coordinate system. Spherical harmonics are used to describe angular momentum in quantum mechanics.\\
\\
Notably, spherical harmonics also form a basis for the irreps of SO(3), i.e.,  the Wigner-D matrices. Specifically, the SO(3) group acts on the $\ell^{th}$ spherical harmonic via the $\ell^{th}$ Wigner-D matrix:
\begin{equation}
	Y_{\ell m}(\theta,\phi) \xrightarrow{\mathfrak{g} \in SO(3)} \sum_{m' = -\ell}^\ell D_{\ell}^{m'm}(\mathfrak{g}) Y_{\ell m'}(\theta,\phi)
	\label{eq:YLMWignerD}
\end{equation}
Therefore, any data encoded in a spherical harmonics basis is acted upon by the SO(3) group via a direct sum of the Wigner-D matrices corresponding to the basis functions being used. Using our nomenclature, any such data encoding constitutes a steerable tensor. We can thus map any function $f(\theta,\phi)$ defined on a sphere into a steerable tensor using the Spherical Fourier Transform (SFT):
\begin{equation}
\hat f_{\ell m} =\int_0^{2\pi}  \int_0^\pi f(\theta,\phi) Y_{\ell m} (\theta,\phi) \sin \theta \,\d\theta\,\d\phi
\label{eq:SFT_appendix}
\end{equation}
The signal can be reconstructed in the real space using the corresponding inverse Fourier transform. For computational purposes, we truncate Fourier expansions at a maximum angular frequency $L$, which results in an approximate reconstruction of the signal $ \tilde{f}(\theta, \varphi)$ up to the angular resolution allowed by $L$,
\begin{equation}
    \tilde{f}(\theta, \varphi) = \sum_{\ell = 0}^{L} \sum_{m = -\ell}^{\ell} \hat{f}_{\ell m} Y_{\ell m}(\theta, \phi)
    \label{eqn:spherical_ift}
\end{equation}
Here, $\hat{f}_{\ell m}$ are the functions' Spherical Fourier coefficients. %

\subsubsection{Zernike polynomials and the Zernike Fourier transform}
To encode a function $\rho(r,\theta,\phi)$ with both radial and angular components, we use Zernike Fourier transform,
\begin{equation}
    \hat{Z}^{n}_{\ell m} = \int\,\rho(r, \theta, \phi) \, Y_{\ell m}(\theta,\phi) R^{n}_{\ell}(r)  \,\d V
    \label{eq:ZFT_appendix}
\end{equation}
where $R^{n}_{\ell}(r)$ is the radial Zernike polynomial in 3D defined as,
\begin{equation}
    R^{n}_{\ell}({r}) =  (-1) ^{\frac{n-\ell}{2}} \sqrt{2n +3} \,
    { \frac{n+\ell+3}{2} -1 \choose \frac{n-\ell}{2}} |{r}|^{\ell}\, {}_{2}F_{1}\left ( - \frac{n-1}{2},\frac{n+\ell +3}{2}; \ell + \frac{3}{2}; |{r}|^2\right)
    \label{eqn:zernike}
\end{equation}
Here, ${}_{2}F_1(\cdot)$ is an ordinary hypergeometric function, and $n$ is a non-negative integer representing a radial frequency, controlling the radial resolution of the coefficients.  $R^{n}_{\ell}(r)$ is non-zero only for even values of $n-\ell\geq 0$.  Zernike polynomials form a complete orthonormal basis in 3D, and therefore, can be used within a Fourier transform to expand and retrieve any 3D shape, if large enough $\ell$ and $n$ coefficient are used. We refer to the Fourier transform of Eq.~\ref{eq:ZFT_appendix} as the Zernike Fourier Trasform (ZFT).\\
\\
To represent point clouds, a common choice for the function $\rho(\vr) \equiv \rho(r,\theta,\phi)$ is the sum of Dirac-$\delta$ functions centered at each point:
\begin{equation}
    \rho(\vr) = \sum_{i\in \text{points}} \delta(\rho(\vr_{i}) - \rho(\vr))
    \label{eq:dirac_delta}
\end{equation}
This choice is powerful because the forward transform has a closed-form solution that does not require a discretization of 3D space for numerical computation. Specifically,  the ZFT of a point cloud follows:
\begin{equation}
    \hat{Z}_{\ell m}^{n} = \sum_{i\in \text{points}} R_{n}^{\ell}(r_i) Y_{\ell m}(\theta_i, \varphi_i)
    \label{eqn:spherical_and_radial_with_dirac_ft}
\end{equation}
Similar to SFT, we can  reconstruct the data using inverse ZFT and define approximations by truncating the angular and radial frequencies at $L$ and $N$, respectively,
\begin{equation}
    \tilde{\rho}(r, \theta, \varphi) = \sum_{\ell = 0}^{L} \sum_{m = -\ell}^{\ell} \sum_{n}^{N} \hat{Z}_{\ell m}^{n} R_{\ell}^{n}(r) Y_{\ell m} (\theta, \varphi)
    \label{eqn:spherical_and_radial_ift}
\end{equation}
\\
The use of other radial bases is possible within our framework, as long as they are complete. Orthonormality is also desirable as it ensures that each basis encodes different information, resulting in a more efficient encoding of the coefficients. We use Zernike polynomials following ref.~\cite{boyd_comparing_2011}, which demonstrates that encoding with Zernike polynomials result in a faster convergence compared to the radial basis functions localized at different radii, as well as most other orthogonal harmonic bases, with the exception of Logan-Shepp. ``Faster convergence" indicates that fewer frequencies are required to encode the same information. ref.~\cite{uhrin_through_2021} also uses Zernike to construct invariant descriptors of atomic environments. However, this choice is not unique and other special functions such as Bessel functions can be used to encode the data~\cite{musaelian_learning_2022}.

\subsection{Holographic-(V)AE (H-(V)AE) details}

\subsubsection{Linearity}
\label{sec:linearity}
Let us consider a feature $\vh_{\ell}$ containing $C$ features of the same degree $\ell$. $\vh_{\ell}$ can be represented as a $C \times (2\ell + 1)$ matrix where each row constitutes an individual feature. Then, we learn weight matrix $\mW_{\ell} \in \mathbb{R}^{C \times K}$ that linearly maps $h_{\ell}$ to $\overline{h}_{\ell} \in \mathbb{R}^{K \times (2\ell + 1)}$:
\begin{equation}
    \overline{\vh}_{\ell} = \mW_{\ell}^{T} \vh_{\ell}
    \label{eqn:linearity}
\end{equation}

\subsubsection{Efficient Tensor Product (ETP)}
\label{sec:ETP_details}

\noindent \textbf{Channel-wise tensor product nonlinearity.} We effectively compute $C$ tensor products, each between features belonging to the same channel $c$, and concatenate  all output features of the same degree. In other words, features belonging to different channels are not mixed in the nonlinearity; the across-channel mixing is instead done in the linear layer. This procedure reduces the computational time and the output size of the nonlinear layers with respect to the number of channels $C$, from $\mathcal O(C^2)$ for a ``fully-connected" tensor product down to $\mathcal O(C)$. The number of learnable parameters in  a linear layer are proportional to  the size of the output space in the preceding nonlinear layer. Therefore, reducing the size of the nonlinear output substantially reduces the complexity of the model and the number of model parameters. This procedure also forces the input tensor to have the same number of channels for all degrees. We refer the reader to ref.~\cite{cobb_efficient_2021} for further details and for a nice visualization of this procedure.\\
\\
\textbf{Minimum Spanning Tree (MST) subset for degree mixing.}
To compute features of the same degree $\ell_3$ using the CG Tensor Product, pairs of features of varying degrees may be used, up to the rules of the CG Tensor Product. Specifically, pairs of features with any degree pair ($\ell_1$, $\ell_2$) may be used to produce a feature of degree $\ell_3$ as long as $|\ell_1 - \ell_2| \leq \ell_3 \leq \ell_1 + \ell_2$. Features of the same degree are then concatenated to produce the final equivariant (steerable) output tensor.

Since each produced feature (often referred to as a ``fragment" in the literature \cite{kondor_clebsch-gordan_2018, cobb_efficient_2021}) is independently equivariant, computing only a subset of them still results in an equivariant output, albeit with lower representational power. Reducing the number of computed fragments is desirable since their computation cannot be easily parallelized. In other words, to reduce complexity we should identify a small subset of fragments that can still offer sufficient representational power. In this work we adopt the ``MST subset" solution proposed in~\cite{cobb_efficient_2021}, which adopts the following strategy: when computing features of the same degree $\ell_3$, exclude the degree pair ($\ell_0$, $\ell_2$) if the ($\ell_0$, $\ell_1$) and the ($\ell_1$, $\ell_2$) pairs have already been computed. The underlying assumption behind this solution is that the last two pairs already contain some information about the first pair, thus making its computation redundant.\\
The resulting subset of pairs can be efficiently computed via the Minimum Spanning Tree of the graph describing the possible pairs used to generate features of a single degree $\ell_3$, given the maximum desired degree $\ell_{max}$. As multiple such trees exist, we choose the one minimizing the computational complexity by weighting each edge (i.e. each pair) in the graph accordingly (edge ($\ell_1$, $\ell_2$) gets weight $(2\ell_1+1)(2\ell_2+1)$). The subset is also augmented to contain all the pairs with same degree to inject more nonlinearity. This procedure reduces the complexity in number of pairs with respect to $\ell_{\text{max}}$ from $\mathcal O(\ell_{\text{max}}^2)$ - when all possible pairs are used - down to $\mathcal O(\ell_{\text{max}})$. We refer the reader to ref.~\cite{cobb_efficient_2021} for more details.

\subsubsection{Batch Norm}
\label{sec:batch_norm}

Let us consider a batch of steerable tensors $h$ which we index by batch $b$, degree $\ell$, order $m$ and channel $c$. During training, we compute a batch-averaged norm for each degree $\ell$ and each channel $c$ as,
\begin{equation}
    N_{\ell}^{c} = \frac{1}{B} \sum_{b=1}^{B} \frac{1}{2\ell+1} \sum_{m=-\ell}^{\ell} (\hat{h}_{\ell m}^{c b})^2
\end{equation}
Similar to standard batch normalization, we also keep a running estimate of the training norms $N_{\ell}^{c, tr(i)}$ using momentum $\xi$, set to $0.1$ in all our experiments:
\begin{equation}
    N_{\ell}^{c, tr(i)} = \xi N_{\ell}^{c} + (1 - \xi) N_{\ell}^{c, tr(i-1)}
\end{equation}
We then update the features of the steerable tensor using the real batch-averaged norms during training, and the running batch-averaged norms during testing, together with a learned affine transformation:
\begin{align}
    \overline{\hat h_{\ell m}^{c b}} &= \frac{\hat{h}_{\ell m}^{c b}}{\sqrt{N_{\ell}^{c}}} \, w_{\ell}^{c} \qquad \qquad \text{training} \\
    \overline{\hat h_{\ell m}^{c b}} &= \frac{\hat{h}_{\ell m}^{c b}}{\sqrt{N_{\ell}^{c, tr(i)}}} \, w_{\ell}^{c} \qquad \, \text{evaluation}
\end{align}

\subsubsection{Signal Norm}
\label{sec:signal_norm}

Similar to for Batch Norm above, let us consider a batch of steerable tensors $h$ which we index by batch $b$, degree $\ell$, order $m$ and channel $c$. Formally, the total norm for an individual tensor $h$ is computed as:
\begin{equation}
N_{tot} = \sum_{\ell} \frac{\sum_{c} \sum_{m = -\ell}^{\ell} (\hat h_{\ell m}^{c})^2}{2\ell+1}
\label{eqn:signal_norm}
\end{equation}
Then, the tensor features are updated as $\overline{\hat h_{\ell m}^{c}} = {\hat h_{\ell m}^{c}} w_{\ell}/{\sqrt{N_{tot}}}$, where $w_{\ell}$ is a degree-specific affine transformation for added flexibility

\subsubsection{Network architecture}
\label{sec:architecture_details}
Certain features of our problem formulation pose constraints upon the specific design choice of the network. For example, within the decoder, the maximum degree $\ell_{\text{max},b}$ that can be outputted by each block $b$ is constrained by the sparsity of the CG tensor product. Specifically, $\ell_{\text{max},b} \leq 2^{b}$ where $b$ ranges from $1$ (first block) to $B$ (last block). Since we need to reconstruct features up to degree $L$ in the decoder, we arrive at a lower bound for the number of blocks in  the decoder set by $\ell_{\text{max},B} \geq L$, or $B \geq \log_2 L$. In our experiments, we set $\ell_{\text{max},b} = \min\{2^{b}, L\}$ and do not let $\ell_{\text{max},b}$ exceed the input's maximum degree $L$. Relaxing this condition might increase the expressive power of the network but at a significant increase in runtime and memory. We leave the analysis of this trade-off to future work. For the encoder and the decoder to have similar expressive power, we construct them to be symmetric with respect to the latent space (Fig.~\ref{fig:model}C). Optionally, we apply a linearity at the beginning of the encoder and at the end of the decoder. However this is required for input data that does not have the same number of channels per degree since the ETP operates channel-wise.

\subsubsection{Tuning the hyperparameters of the training objective}
\label{sec:training_objective_hyperparams}
We find it practical to scale the reconstruction loss by a dataset-specific scalar $\alpha$ since the MSE loss varies in average magnitude across datasets. When training H-VAE, we find it beneficial to keep $\beta = 0$ for a few epochs ($E_\text{rec}$) so that the model can learn to perform meaningful reconstructions, and then linearly increasing it to the desired value for $E_\text{warmup}$ epochs to add structure to the latent space, an approach first used by ref.~\cite{bowman_generating_2016}.

\subsubsection{Data normalization}
\label{sec:data_normalization}
As per standard machine learning practice~\cite{shanker_effect_1996}, we normalize the data. We do this by dividing each tensor by the average square-root total norm of the training tensors, analogously to the Signal Norm. This strategy puts the target values on a similar scale as the normalized activations learned by the network, which we speculate to favor gradient flow.

\subsubsection{Pairwise invariant reconstruction loss}
\label{sec:pairwise_inv_rec_loss}
To reconstruct a signal within an equivariant model it is desirable to have a \textit{pairwise invariant} reconstruction loss, i.e., a loss $\mathcal{L}_{rec}$ such that $\mathcal{L}_{rec}(\vx, \vy) = \mathcal{L}_{rec}(\mD(\mathfrak{g})\vx, \mD(\mathfrak{g})\vy)$ where $\mD$ is the representation of the group element $\mathfrak{g}$ acting on the space that $x$ and $y$ inhabit (e.g. a rotation matrix if $\vx$ and $\vy$ are vectors in Euclidean 3D space, or a degree-$\ell$ wigner-D matrix if $\vx$ and $\vy$ are degree-$\ell$ vectors). This  property is necessary for the model to remain equivariant, i.e., given that the network is agnostic to the transformation of the input under group operation $\vx \to \mD(\mathfrak{g})\vx$ by producing a similarly transformed output $\vy \to \mD(\mathfrak{g})\vy$, we want the reconstruction loss to be agnostic to the same kind of transformation as well\\
\\
The MSE loss is pairwise invariant for any degree-$\ell$ feature on which SO(3) acts via the $\ell$'s Wigner-D matrix. Consider two degree-$\ell$ features $\vx_{\ell}$ and $\vy_{\ell}$ acted upon by a Wigner-D matrix $D_{\ell}(\mathfrak{g})$ parameterized by rotation $\mathfrak{g}$ (we drop the $\mathfrak{g}$ and $\ell$ indexing for clarity):
\begin{align}
 \nonumber    \text{MSE}(\mD\vx, \mD\vy) &= (\mD\vx - \mD\vy)^{T} (\mD\vx - \mD\vy) \\
  \nonumber                       &= (\mD(\vx - \vy))^{T} (\mD(\vx - \vy)) \\
  \nonumber                       &= (\vx - \vy)^{T}\mD^{T} \mD(\vx - \vy) \\
  \nonumber                       &= (\vx - \vy)^{T} (\vx - \vy) \qquad \qquad \text{since Wigner-D matrices are unitary} \\ 
  \nonumber                       &= \text{MSE}(\vx, \vy)\\
\end{align}
Since the MSE loss is pairwise invariant for every pair of degree-$\ell$ features, it is thus pairwise invariant for pairs of steerable tensors composed via direct products of steerable features.

\subsubsection{Dependence of the reconstruction loss on feature degree}
We observe that features of larger degrees $\ell$ are harder to reconstruct accurately. Specifically, Fig.~S8 shows that the test reconstruction loss (MSE) increases with degree $\ell$ for both the MNIST-on-the-sphere  and the Toy amino acids problems.

\subsubsection{Cosine loss}
\label{sec:cosine_loss_appendix}
We use the metric {\em Cosine loss} to measure a model's reconstruction ability. Cosine loss is a normalized dot product generalized to operate on pairs of steerable tensors (akin to cosine similarity), and modified to be interpreted as a loss (see Sec.~\ref{sec:cosine_loss_appendix} for details),
\begin{equation}
\begin{split}
    \text{Cosine}(\vx, \vy) = 1 - \frac{\vx \odot \vy}{\sqrt{(\vx \odot \vx)(\vy \odot \vy)}}, \\
    \text{with}\quad \vx \odot \vy = \sum_{\ell'} (\vx_{\ell'} \otimes_{cg} \vy_{\ell'})_{\ell = 0}
\end{split}
    \label{eqn:cosine_loss}
\end{equation}
Importantly, unlike MSE, which depends on the characteristics of the data (e.g., the size of the data tensors),  Cosine loss is dimensionless and therefore, interpretable and comparable across datasets.  A measure with these characteristics is practically useful for evaluating a network because it provides an estimate for how much better the reconstructions can get if the network’s hyperparameters were to be further optimized. For example, looking at the Cosine loss in Table~\ref{table:mse_and_cosine}, we see that our model trained on Shrec17 (best Cosine = 0.130) is not as well optimized as our model trained on MNIST (best Cosine = 0.017). Using MSE, the trend is reversed (1.8$\times 10^{-3}$ vs. 6.7$\times 10^{-3}$), since the scale of MSE depends on the size of the irreps of the data (Fig.~S9). Nonetheless, as a measure of a model's reconstruction ability, Cosine loss correlates almost perfectly with MSE for a given dataset and network, especially in the mid-to-low reconstruction quality regime (SpearmanR~$= 0.99$, Fig.~S9 and Table~\ref{table:mse_and_cosine}). Crucially, however, as Cosine loss ignores the relative norms of data features, it is unable to reconstruct norms and thus, not suitable as a training objective. \\

\noindent \textbf{Proof: Invariance of  Cosine loss.}
The generalized dot product $\odot$ from Eq.~\ref{eqn:cosine_loss} is pairwise invariant in the same way that the dot product between two 3D vectors depends only on their relative orientations but not the global orientation of the whole two-vector system. Therefore, the whole Cosine loss expression is pairwise invariant, since all of its components are pairwise invariant.\\

\subsection{Using a non-equivariant decoder}
\label{sec:mlp_decoder_appendix}
{\em Winter et al}~\cite{winter_unsupervised_2022} propose to construct group-equivariant autoencoders by using an equivariant encoder that learns an invariant embedding and a group element, and an unconstrained decoder which uses the invariants alone to reconstruct each datapoint in the "canonical" form, before applying the learned group action in the output space. By contrast, for SO(3) we propose to use an equivariant decoder, whereby the learned group element is fed as input to the decoder. Such ``unconstrained decoder" procedure can in principle be merged with our equivariant encoder and Fourier-space approach in two ways. For each, we argue in favor of using our equivariant decoder.\\
\\
\textbf{1) Reconstructing the Fourier coefficients of the data.} To apply the learned group element on the decoder’s output, the Wigner-D matrices for the data’s irreps need to be computed from the group element. Then, the Wigner-D matrices can be used to “rotate” the tensor. This has to be done on-the-fly, and it can be done quickly using functions provided in the e3nn package~\cite{geiger_e3nn_2022} and by smartly vectorizing operations. We implemented this procedure by using a simple Multi-Layer Perceptron with SiLU non-linearities as a decoder. By using e3nn to compute Wigner-D matrices in batches, and by clever construction of tensor multiplications such that runtime scales linearly with $\ell_{max}$ and is constant with regards to number of channels and batch size, we achieve models that run with comparable speed to those using our equivariant decoder, and have comparable performance on MNIST (Table~\ref{table:mlp_decoder_comparison}). Given the empirical similarities we observe, though on a limited use case, we favor the simplicity and elegance of our equivariant decoder. “Simplicity” because we construct the decoder to be symmetric to the encoder, thus endowing it automatically with similar representational power and without the need to tune an architecture made with different base components. Furthermore, we highlight that our method generates intermediate equivariant representations in the decoder, rather than intermediate invariant representations. These intermediate equivariant representations may be of interest to study in and of themselves.\\
\\
\textbf{2) Reconstructing the data in real space.} In this case, we do not have to compute Wigner-D matrices on-the-fly, since the learned frame can be used directly in the output space as a rotation matrix. However, since the encoder only sees a truncated Fourier representation of the data, which is by construction lossy, while the loss is computed over fine-grained real-space, this model might be too difficult to train. We suspect this would make the model akin to a denoising autoencoder~\cite{vincent_extracting_2008} and it might be interesting to analyze, but that would be beyond the scope of this paper. To avoid the denoising effect, we could learn to reconstruct data in real space after an Inverse Fourier Transform (IFT). However, computing the IFT on-the-fly is very expensive and requires a discretization of the input space, to the point of being prohibitive for point clouds. This is not a bottleneck for the Forward Fourier Transform if the cloud is parameterized by Dirac-Delta distributions, i.e., for point clouds, as the integral can be computed exactly (Eq.~\ref{eqn:spherical_and_radial_with_dirac_ft}).

\begin{table}[h]
\begin{center}
\resizebox{0.8\textwidth}{!}{
\begin{tabular}{l c | c | c c | c c | c c}
    \bf Method & \bf z & \bf Speed & \bf MSE & \bf Cosine & \bf Purity & \bf V-meas. & \bf \newlinecell{LC Class.\\Acc.} & \bf \newlinecell{KNN Class.\\Acc.} \\
    \hline
    \hline
    H-AE NR/R                       & 16 & \textbf{1.0}& $\mathbf{0.9 \times 10^{-3}}$ & \textbf{0.025} & \textbf{0.62} & \textbf{0.51} & \textbf{0.820} & \textbf{0.862} \\
    H-AE unconst. decoder NR/R     & 16 & 1.3 & $1.3 \times 10^{-3}$ & 0.037 & 0.61 & 0.48 & 0.802 & 0.856 \\
    \hline
    H-VAE NR/R                      & 16 & \textbf{1.0} & $\mathbf{1.4 \times 10^{-3}}$ & \textbf{0.057} & \textbf{0.67} & \textbf{0.54} & \textbf{0.812} & 0.848 \\
    H-VAE unconst. decoder NR/R    & 16 & 1.3 & $2.1 \times 10^{-3}$ & \textbf{0.057} & 0.62 & 0.51 & 0.781 & \textbf{0.853} \\
\end{tabular}
}
\end{center}
\caption{\textbf{Performance comparison between our H-(V)AE and a H-(V)AE constructed with ref.~\cite{winter_unsupervised_2022}'s non-equivariant decoder formulation, on the MNIST-on-the-sphere dataset.} The non-equivariant decoders are constructed as simple MLPs with SiLU non-linearities, with the following hidden layer sizes: [32,64,128,160,256]. We keep the number of parameters approximately the same to make model comparison fair, but we do not tune the architecture of the invariant decoders. All other training details are kept the same (Sec.~\ref{sec:exp_details}).}
\label{table:mlp_decoder_comparison}
\end{table}

\subsection{Implementation details}

Without loss of generality, we use real spherical harmonics for implementation of H-(V)AE. 
We leverage \verb|e3nn|~\cite{geiger_e3nn_2022}, using their computation of the real spherical harmonics and their Clebsch-Gordan coefficients.\\
\\
In our code, we offer the option to use the Full Tensor Product instead of the ETP. Specifically, at each block we allow the users to specify whether to compute the Tensor Product channel-wise or fully-connected across channels, and whether to compute using efficient or fully connected degree mixing.

\subsection{Experimental details}
\label{sec:exp_details}

\subsubsection{Architecture specification}
We describe model architectures as follows. We specify the number of blocks $B$, which is the same for the encoder and the decoder. We specify two lists, (i) DegreesList which contains the maximum degree $\ell_{{max},b}$ of the output of each block $b$, and (ii) ChannelsList, containing the channel sizes $C_{b}$, of each block $b$. These lists are in the order as they appear in the encoder, and are reversed for the decoder. When it applies, we specify the number of output channels of the initial linear projection $C_\text{init}$. As noted in the main text, we use a fixed formula to determine $\ell_{{max},b}$, but we specify it for clarity.

\subsubsection{MNIST on the sphere}
\label{sec:mnist_details}
\textbf{Model architectures.}
For models with invariant latent space size $z = 16$, we use 6 blocks, $\text{DegreesList} = [10, 10, 8, 4, 2, 1]$ and $\text{ChannelsList} = [16, 16, 16, 16, 16, 16]$, with a total of 227k parameters.\\
For models with invariant latent space size $z = 120$, we use 6 blocks, $\text{DegreesList} = [10, 10, 8, 4, 2, 1]$ and $\text{ChannelsList} = [16, 16, 16, 32, 64, 120]$, with a total of 453k parameters.\\
\\
\textbf{Training details.}
We keep the learning schedule as similar as possible for all models. We use $\alpha = 50$. We train all models for $80$ epochs using the Adam optimizer~\cite{kingma_adam_2017} with default parameters, a batch size of $100$, and an initial learning rate of $0.001$, which we decrease exponentially by one order of magnitude over $25$ epochs; see Fig.~S2 for the behavior of the training loss. For VAE models, we use $\beta = 0.2$, $E_\text{rec} = 25$ and $E_\text{warmup} = 35$. We utilize the model with the lowest loss on validation data, only after the end of the warmup epochs for VAE models. Training took $\sim 4.5$ hours on a single NVIDIA A40 GPU for each model.

\subsubsection{Shrec17}
\label{sec:shrec17_details}
\textbf{Model architectures.} Both AE and VAE models have $z = 40$, 7 blocks, $\text{DegreesList} = [14, 14, 14, 8, 4, 2, 1]$, $\text{ChannelsList} = [12, 12, 12, 20, 24, 32, 40]$, $C_\text{init} = 12$, with a total of 518k parameters.\\
\\
\textbf{Training details.} We keep the learning schedule as similar as possible for all models. We use $\alpha = 1000$. We train all models for $120$ epochs using the Adam optimizer with default parameters, a batch size of $100$, and an initial learning rate of $0.0025$, which we decrease exponentially by two orders of magnitude over the entire $120$ epochs. For VAE models, we use $\beta = 0.2$, $E_\text{rec} = 25$ and $E_\text{warmup} = 10$. We utilize the model with the lowest loss on validation data, only after the end of the warmup epochs for VAE models. Training took $\sim 11$ hours on a single NVIDIA A40 GPU for each model.

\subsubsection{Embedding of amino acid conformations}
\label{sec:toy_aminoacids_details}

\noindent \textbf{Pre-processing of protein structures.} We sample residues from the set of training structures pre-processed as described in Sec.~\ref{sec:protein_neighborhoods_details}.\\

\noindent \textbf{Fourier projection.} We set the maximum radial frequency to $N = 20$ as  it corresponds to a radial resolution matching the minimum inter-atomic distances after rescaling the atomic neighborhoods of radius  $10.0\AA$ to fit within a sphere of radius 1.0, necessary for Zernike transform.\\
The channel composition of the data tensors can be described  in a notation - analogous to that used by e3nn but without parity specifications - which specifies the number of channels $C$ for each feature of degree $\ell$ in single units $C\text{x}\ell$: 44x0 + 40x1 + 40x2 + 36x3 + 36x4.\\
\\
\textbf{Model architectures.} All models have $z = 2$, 6 blocks, $\text{DegreesList} = [4, 4, 4, 4, 2, 1]$, $\text{ChannelsList} = [60, 40, 24, 16, 16, 8]$, $C_\text{init} = 48$, with a total of 495k parameters. We note that the initial projection is necessary since the number of channels differs across feature degrees in the data tensors.\\
\\
\textbf{Training details.} We keep the learning schedule as similar as possible for all models. We use $\alpha = 400$. We train all models for $80$ epochs using the Adam optimizer with default parameters and an initial learning rate of $0.005$, which we decrease exponentially by by one order of magnitude over $25$ epochs. For VAE models, we use $E_\text{rec} = 25$ and $E_\text{warmup} = 10$. We utilize the model with the lowest loss on validation data, only after the end of the warmup epochs for VAE models.\\
We vary the batch size according to the size of the training and the validation datasets. We use the following (dataset\_size-batch\_size) pairs: (400-4), (1,000-10), (2,000-20), (5,000-50), (20,000-20). Training took $\sim 45$ minutes on a single NVIDIA A40 GPU for each model.\\
\\
\textbf{Evaluation.} We perform our data ablations by considering training and validation datasets of the following sizes: 400, 1,000, 2,000, 5,000 and 20,000. We keep relative proportions of residue types even in all datasets. We perform the data ablation with H-AE as well as H-VAE models with $\beta = 0.025$ and $0.1$.\\
We further perform a $\beta$ ablation using the full (20,000) dataset, over the following choices of $\beta$: $[0 \text{(AE)}, 0.025, 0.05, 0.1, 0.25, 0.5]$.\\
\\
For robust results, we train 3 versions of each model and compute averages of quantitative metrics of reconstruction loss and classification accuracy.\\
\\
For a fair comparison across models with varying amounts of training and validation data, we perform a 5-fold cross-validation-like procedure over the 10k test residues, where the classifier is trained over 4 folds of the test data and evaluated on the fifth one. If validation data is needed for model selection (e.g. for LC), we use 10\% of the training data.

\subsubsection{Embedding of protein structure neighborhoods}
\label{sec:protein_neighborhoods_details}
\noindent \textbf{Pre-processing of protein structures}.
We model protein neighborhoods extracted from tertiary protein structures from the Protein Data Bank (PDB)~\cite{berman_protein_2000}. 
We use ProteinNet’s splittings for training and validation sets to avoid redundancy, e.g. due to similarities in homologous protein domains~\cite{alquraishi_proteinnet_2019}. Since PDB ids were only provided for the training and validation sets, we used ProteinNet’s training set as both our training and validation set and ProteinNet’s validation set as our testing set. Specifically, we make a $[80\%, \, 20\%]$ split in the ProteinNet's training data to define our training and validation sets. This splitting resulted in 10,957 training structures, 2,730 validation structures, and 212 testing structures. \\ %
\\
\textbf{Projection details.} 
We set the maximum radial frequency to $N = 20$ as it corresponds to a radial resolution matching the minimum inter-atomic distances after rescaling the atomic neighborhoods of radius  $10.0$ \AA\, to fit within a sphere of radius 1.0, necessary for Zernike transform. We vary $L$ and construct models tailored to each one (Table~\ref{table:NBs_model_architectures}). \\
\\
\textbf{Model architectures.} See Table~\ref{table:NBs_model_architectures}. We note that the initial projection ($C_{\text{init}}$) is necessary since the number of channels differs across feature degrees in the data tensors, and the ETP necessitates equal number of channels for all degrees (Sec.~\ref{sec:ETP_details}).\\
\\
\textbf{Training details.} We keep the learning schedule the same for all models. We use $\alpha = 1000$. We train all models for $8$ epochs using the Adam optimizer with default parameters, a batch size of 512, and a constant learning rate of $0.001$. We utilize the model with the lowest loss on validation data.\\

\begin{table}[h!]
\begin{center}
\begin{tabular}{c c c c c c c c}
    \bf $z$ & \bf $L$ & \bf Tensor size & $\mathbf{C_{init}}$ & \bf ChannelsList & \bf DegreesList & \bf \# Params & \bf Training Speed (hours) \\
    \hline
     64 & 3 &  616 & 44 &    [50,50,50,64,64,64] & [3,3,3,3,2,1] & 631k &  3.0 \\
     64 & 4 &  940 & 44 &    [50,50,50,64,64,64] & [4,4,4,4,2,1] & 950k &  3.8 \\
     64 & 6 & 1708 & 44 &    [50,50,50,64,64,64] & [6,6,6,4,2,1] & 1.6M &  5.9 \\
     64 & 8 & 2604 & 44 &    [50,50,50,64,64,64] & [8,8,8,4,2,1] & 2.4M &  9.3 \\
    \hline
    128 & 3 &  616 & 44 &  [60,60,60,90,128,128] & [3,3,3,3,2,1] & 1.2M &  3.8 \\
    128 & 4 &  940 & 44 &  [60,60,60,90,128,128] & [4,4,4,4,2,1] & 1.7M &  4.4 \\
    128 & 6 & 1708 & 44 &  [60,60,60,90,128,128] & [6,6,6,4,2,1] & 2.6M &  6.0 \\
    128 & 8 & 2604 & 44 &  [60,60,60,90,128,128] & [8,8,8,4,2,1] & 3.8M & 10.1 \\
    \hline
    256 & 3 &  616 & 44 & [90,90,90,120,200,256] & [3,3,3,3,2,1] & 2.8M &  4.9 \\
    256 & 4 &  940 & 44 & [90,90,90,120,200,256] & [4,4,4,4,2,1] & 3.8M &  5.6 \\
    256 & 6 & 1708 & 44 & [90,90,90,120,200,256] & [6,6,6,4,2,1] & 4.4M &  6.0 \\
    256 & 8 & 2604 & 44 & [90,90,90,120,200,256] & [8,8,8,4,2,1] & 7.9M & 14.5
\end{tabular}
\end{center}
\caption{{\bf Model architecture hyperparameters and training time for each H-AE trained on protein neighborhoods.}}
\label{table:NBs_model_architectures}
\end{table}

\subsubsection{Latent space classification}
\label{sec:latent_space_classification}
\noindent \textbf{Linear classifier.}
We implement the linear classifier as a one-layer fully connected neural network with input size equal to the invariant embedding of size $z$, and output size equal to the number of desired classes. We use cross entropy loss with logits as training objective, which we minimize for 250 epochs using the Adam optimizer with batch size 100, and initial learning rate of 0.01. We reduce the learning rate by one order of magnitude every time the loss on validation data stops improving for 10 epochs (if validation data is not provided, the training data is used). At evaluation time, we select the class with the highest probability value. We use PyTorch for our implementation.

\noindent\textbf{KNN Classifier}
We use the \texttt{sklearn}~\cite{pedregosa_scikit-learn_2011} implementation with default parameters. At evaluation time, we select the class with the highest probability value.

\subsubsection{Clustering metrics}
\label{sec:clustering_metrics}
\noindent\textbf{Purity~\cite{aldenderfer_cluster_1984}.} We first assign a class to each cluster based on the most prevalent class in it. Purity is computed as the sum of correctly classified items divided by the total number of items. Purity measures classification accuracy, and ranges between $0$ (worst) and $1$ (best).\\
\\
\textbf{V-measure~\cite{rosenberg_v-measure_2007}.} This common clustering metric strikes a balance between favoring homogeneous (high homogeneity score) and complete (high completeness score) clusters. Clusters are defined as homogeneous when all elements in the same cluster belong to the same class (akin to a precision). Clusters are   defined as  complete when all elements belonging to the same class are put in the same cluster (akin to a recall). The V-measure is  computed as the harmonic mean of homogeneity and completeness in a given clustering.

\subsubsection{On the complementary nature of classification accuracy and clustering metrics}
The clustering metrics ``purity" and ``V-measure” and the supervised metric  ``classification accuracy” characterize different qualities of the latent space, and, while partly correlated, they are complementary to each other.\\
\\
Both classes of metrics are computed by comparing the ground truth labels to the predicted labels, and they mainly differ by how the predicted labels are assigned; the clustering metrics use  an unsupervised clustering algorithm, while the classification metric uses a supervised classification algorithm to do so. As a result, these metrics focus on different features of the latent space. For example, the clustering metrics are largest when the test data naturally forms clusters with all data points of the same label.  While this case can result in a high supervised classification accuracy, clustering is not a necessary condition for high classification accuracy. Indeed, the supervised signal could make the predicted labels depend more heavily on a subset of the latent space features, instead of relying on all of them equally, which is what the clustering algorithm naturally does. Therefore, it is reasonable to conclude that having higher clustering metrics and a lower classification accuracy is a sign that class-related information is more evenly distributed across the latent space dimensions. Overall, the complementary aspect of these metrics makes it necessary to use all of them when comparing the performance of different models in each task.

\subsection{Protein-Ligand Binding Affinity Prediction}
\label{sec:lba_details}
\subsubsection{Data Preprocessing Details}
We leverage the H-AE models trained on Protein Neighborhoods (Sec.~\ref{sec:protein_neighborhoods_details}) to predict the binding affinity between a protein and a ligand, given their  structure complex-- an important task in structural biology. For this task, we use the data from the ``refined-set" of PDBBind~\cite{su_comparative_2019}, containing $\sim 5,000$ structures. We use the dataset splits provided by ATOM3D~\cite{townshend_atom3d_2021} to benchmark our predictions. ATOM3D provides two splits based on the maximum sequence similarity between proteins in the training set and the validation / test sets. We use the most challenging split for our benchmarking in which the sequences of the two sets have at most 30\% similarity.

For each complex, we first identify residues in the binding pocket, which we define as residues for which the C$\alpha$ is within 10\AA\, of any of the ligand's atoms. We then extract the 10\AA\, neighborhood for each of the pocket residues that can contain atoms from both the protein and the ligand. We compute the ZFT (Eq.~\ref{eq:ZFT}) for each neighborhood with maximum degree $L$ matching the degree used by the H-AE of interest. We then compute residue-level SO(3)-invariant embeddings by running the pre-trained H-AE encoder on the neighborhoods surrounding each of the residues in the pocket. We sum over the residue-level embeddings to compute a single pocket-level embedding, which is SE(3)-invariant by construction. Each pocket embedding is used as a feature vector within a standard Machine Learning Regression model to predict protein-ligand binding affinity, provided by PDBBind either in the form of a \textit{dissociation constant} $K_d$, or an \textit{inhibition constant} $K_i$. Due to data limitations, and consistent with other studies on LBA, we do not distinguish between $K_d$ and $K_i$ and regress over  the negative log of either of the constants that is provided by PDBBind; the transformed quantity is closely related to the binding free energy of protein-ligand interactions. 

The protein-ligand binding free energy is an extensive quantity, meaning that its magnitude depends on the number of residues in the binding pocket. In other words, a larger protein-ligand complex can establish a stronger binding. The pocket embedding, which we define as the sum of the residue-level embeddings, is a simple quantity that is extensive and SE(3)-invariant, and therefore, suitable for regressing over protein-ligand binding free energy. Nonetheless, this map is not unique and other extensive transformations to pool together residue-level embeddings into a pocket-level embedding can be used for this purpose.

It should be noted that in the protein-ligand structural neighborhoods we only include the ligand atoms that are found in proteins and were used in the training of the H-AE models (i.e., C, N, O, and S). About 31\% of the ligands contain other kinds of atoms, and since our model does not ``see" these atoms, we hypothesize that our predictions are worse in these cases. In fact, for H-AE+R.F. (Table~\ref{table:lba_results}), the Spearman's correlation over the ligands containing only protein atoms is higher than for the ligands containing other kinds of atoms by about 0.3 points. Therefore, we expect training H-(V)AE to recognize more atom types - or at  least making it aware that some other unspecified non-protein atom is present - would yield even better results; as protein structures are often found in complex with other non-protein entities, e.g. ligands and ions, there is training data available for constructing such models.

\subsubsection{Machine Learning models used for prediction}
\noindent \textbf{Linear Regression.} We implement a linear regression model as a one-layer fully connected neural network with one output layer. We use MSE loss as training objective, which we minimize for 250 epochs using the Adam optimizer with batch size 32, and initial learning rate of 0.01. We reduce the learning rate by one order of magnitude every time the loss on validation data stops improving for 10 epochs. We use PyTorch for our implementation.\\

\noindent \textbf{Random Forest Regression.} We use the \texttt{sklearn}~\cite{pedregosa_scikit-learn_2011} implementation. We tune hyperparameters via grid search, choosing the combination minimizing RMSD on validation data. We consider the following grid of hyperparameter values:

\begin{itemize}
    \item max\_features: [$1.0$, $0.333$, sqrt]
    \item min\_samples\_leaf: [$2$, $5$, $10$, $15$]
    \item n\_estimators: [$32$, $64$, $100$]
\end{itemize}

\subsubsection{Extended Discussion of Results}

Table~\ref{table:lba_results_appendix} shows benchmarking results on the split of ATOM3D with 30\% sequence similarity, extending Table~\ref{table:lba_results} in the main text with more baselines. Notably, we show the prediction results for the models that only use the SO(3)-invariant ($\ell = 0$) component of each neighborhood's Zernike transform (our Zernike inv. baseline), instead of the complete embeddings learned by H-AE. Using H-AE consistently outperforms this baseline, implying that invariant information encoded in higher spherical degrees, and extracted by H-AE, can lead to more expressive models for downstream regression tasks.

\clearpage

\begin{table}[h]
\begin{center}
\resizebox{0.5\textwidth}{!}{
\begin{tabular}{l l c c | c c}
    \bf Type & \bf Method & \bf z & \bf bw & \bf LC Acc. & \bf KNN Acc.\\
    \hline
    \multirow{8}{*}{Unsupervised}
    & H-AE NR/R      & 120 & 30 & 0.877 & 0.875 \\
    & H-AE R/R       & 120 & 30 & 0.881 & 0.886 \\
    & H-AE NR/R      &  16 & 30 & 0.820 & 0.862 \\
    & H-AE R/R       &  16 & 30 & 0.833 & 0.876 \\
    & H-VAE NR/R     & 120 & 30 & 0.883 & 0.879 \\
    & H-VAE R/R      & 120 & 30 & 0.884 & \textbf{0.895} \\
    & H-VAE NR/R     &  16 & 30 & 0.812 & 0.848 \\
    & H-VAE R/R      &  16 & 30 & 0.830 & 0.874 \\
\end{tabular}}
\end{center}
\caption{Evaluation of network performances for MNIST-on-the-sphere using a KNN classifier instead of linear classifier in the latent space. Results are significantly better than when using a linear classifier for models with smaller ($z = 16$) latent space, comparable for the other models.\\}
\label{table:mnist_scores_knn}
\end{table}

\begin{table}[h]
\begin{center}
\resizebox{0.7\textwidth}{!}{
\begin{tabular}{l l c c | c | c c c c c}
    \bf Type & \bf Method & \bf z & \bf bw & \newlinecell{\bf Class.\\\bf Acc.} & \bf P@N & \bf R@N & \bf F1@N & \bf mAP & \bf NDCG \\
    \hline
    \multirow{2}{*}{Unsupervised + LC}
    & H-AE        &  40 & 90 & 0.654 & 0.548 & 0.569 & 0.545 & 0.500 & 0.597 \\
    & H-VAE       &  40 & 90 & 0.631 & 0.512 & 0.537 & 0.512 & 0.463 & 0.568 \\
    \hline
    \multirow{2}{*}{Unsupervised + KNN}
    & H-AE        &  40 & 90 & \textbf{0.672} & \textbf{0.560} & \textbf{0.572} & \textbf{0.555} & \textbf{0.501} & \textbf{0.599} \\
    & H-VAE       &  40 & 90 & 0.658 & 0.541 & 0.558 & 0.539 & 0.487 & 0.591 \\
\end{tabular}}
\end{center}
\caption{Evaluation of network performances for Shrec17 using a KNN classifier instead of linear classifier in the latent space. Results are better than when using a linear classifier.\\}
\label{table:shrec17_evaluation_knn}
\end{table}

\begin{table}[h]
    \centering
     \resizebox{\textwidth}{!}{
    \begin{tabular}{l | c c c c | c c c c | c c c c}
         & \multicolumn{4}{c}{\bf H-AE} \vline & \multicolumn{4}{c}{\bf H-VAE ($\beta = 0.025$)} \vline & \multicolumn{4}{c}{\bf H-VAE ($\beta = 0.1$)} \\
        {\bf \# train} & {\bf MSE} & {\bf Cosine loss} & {\bf LC Acc.} & {\bf KNN Acc.} & {\bf MSE} & {\bf Cosine loss} & {\bf LC Acc.} & {\bf KNN Acc.} & {\bf MSE} & {\bf Cosine loss} & {\bf LC Acc.} & {\bf KNN Acc.} \\
        \hline
        0       & $1.3 \times 10^{-2}$ & 1.015 & 0.409 & 0.629  &  $1.5 \times 10^{-2}$ & 0.981 & 0.424 & 0.656  &  $1.5 \times 10^{-2}$ & 0.981 & 0.424 & 0.656 \\
        400     & $9.4 \times 10^{-4}$ & 0.153 & 0.586 & 0.842  &  $9.8 \times 10^{-4}$ & 0.160 & 0.616 & 0.848  &  $1.0 \times 10^{-3}$ & 0.163 & 0.558 & 0.780 \\
        1,000   & $5.9 \times 10^{-4}$ & 0.099 & 0.583 & 0.856  &  $6.3 \times 10^{-4}$ & 0.101 & 0.569 & 0.854  &  $6.9 \times 10^{-4}$ & 0.113 & 0.564 & 0.844 \\
        2,000   & $4.5 \times 10^{-4}$ & 0.073 & 0.560 & 0.900  &  $4.9 \times 10^{-4}$ & 0.081 & 0.554 & 0.905  &  $5.5 \times 10^{-4}$ & 0.092 & 0.593 & 0.890 \\
        5,000   & $3.3 \times 10^{-4}$ & 0.053 & 0.629 & 0.940  &  $3.3 \times 10^{-4}$ & 0.053 & 0.638 & 0.961  &  $4.3 \times 10^{-4}$ & 0.072 & 0.588 & 0.921 \\
        20,000  & $2.2 \times 10^{-4}$ & 0.034 & 0.578 & 0.972  &  $2.4 \times 10^{-4}$ & 0.037 & 0.667 & 0.971  &  $2.9 \times 10^{-4}$ & 0.047 & 0.662 & 0.966 \\
    \end{tabular}
    }
    \caption{Quantitative data ablation results on the Toy amino acids dataset. A random-guessing classifier has an expected accuracy of 0.050.}
    \label{table:toy_aas_data_ablation_z_eq_2}
\end{table}

\begin{table}[h]
    \centering
     \begin{tabular}{l | c c c c}
        {\bf $\beta$} & {\bf MSE} & {\bf Cosine} & {\bf LC Acc.} & {\bf KNN Acc.}\\
        \hline
        0 (AE) & $2.2 \times 10^{-4}$ & 0.034 & 0.580 & 0.972 \\
        0.025  & $2.4 \times 10^{-4}$ & 0.037 & 0.666 & 0.971 \\
        0.05   & $2.5 \times 10^{-4}$ & 0.039 & 0.669 & 0.968 \\
        0.1    & $2.9 \times 10^{-4}$ & 0.047 & 0.661 & 0.966 \\
        0.25   & $6.8 \times 10^{-4}$ & 0.132 & 0.597 & 0.854 \\
        0.5    & $1.1 \times 10^{-3}$ & 0.203 & 0.467 & 0.722 \\
    \end{tabular}   \caption{Quantitative data ablation results on the Toy amino acids dataset. Models were trained on the full dataset (\# train = 20,000).}
    \label{table:toy_aas_kld_ablation_z_eq_2}
\end{table}

\begin{table}[ht]
\begin{center}
\begin{tabular}{l l c c c c | c c}
    \bf Dataset & \bf Method & \bf $L$ & \bf $z$ & \bf bw & \bf \# Params & \bf MSE & \bf Cosine \\
    \hline
    \hline
    \multirow{8}{*}{MNIST}
    & H-AE NR/R     & 10 & 120 & 30 & 453k & $6.2 \times 10^{-4}$ & 0.017 \\
    & H-AE R/R      & 10 & 120 & 30 & 453k & $6.8 \times 10^{-4}$ & 0.018 \\
    & H-AE NR/R     & 10 &  16 & 30 & 227k & $9.3 \times 10^{-4}$ & 0.025 \\
    & H-AE R/R      & 10 &  16 & 30 & 227k & $8.9 \times 10^{-4}$ & 0.024 \\
    & H-VAE NR/R    & 10 & 120 & 30 & 453k & $1.4 \times 10^{-3}$ & 0.037 \\
    & H-VAE R/R     & 10 & 120 & 30 & 453k & $1.4 \times 10^{-3}$ & 0.037 \\
    & H-VAE NR/R    & 10 &  16 & 30 & 227k & $2.2 \times 10^{-3}$ & 0.057 \\
    & H-VAE R/R     & 10 &  16 & 30 & 227k & $2.1 \times 10^{-3}$ & 0.055 \\
    \hline
    \multirow{2}{*}{Shrec17}
    & H-AE          & 14 &  40 & 90 & 518k & $1.8 \times 10^{-4}$ & 0.130 \\
    & H-VAE         & 14 &  40 & 90 & 518k & $2.2 \times 10^{-4}$ & 0.151 \\
    \hline
    \multirow{12}{*}{Protein NBs}
    & H-AE          &  3 &  64 & - & 631k & $3.3 \times 10^{-4}$ & 0.084 \\
    & H-AE          &  4 &  64 & - & 950k & $3.9 \times 10^{-4}$ & 0.125 \\
    & H-AE          &  6 &  64 & - & 1.6M & $5.4 \times 10^{-4}$ & 0.219 \\
    & H-AE          &  8 &  64 & - & 2.4M & $6.3 \times 10^{-4}$ & 0.287 \\
    \cline{2-8}
    & H-AE          &  3 & 128 & - & 1.2M & $2.1 \times 10^{-4}$ & 0.053 \\
    & H-AE          &  4 & 128 & - & 1.7M & $2.8 \times 10^{-4}$ & 0.087 \\
    & H-AE          &  6 & 128 & - & 2.6M & $4.0 \times 10^{-4}$ & 0.156 \\
    & H-AE          &  8 & 128 & - & 3.8M & $4.9 \times 10^{-4}$ & 0.213 \\
    \cline{2-8}
    & H-AE          &  3 & 256 & - & 2.8M & $1.0 \times 10^{-4}$ & 0.025 \\
    & H-AE          &  4 & 256 & - & 3.8M & $1.9 \times 10^{-4}$ & 0.060 \\
    & H-AE          &  6 & 256 & - & 4.4M & $2.7 \times 10^{-4}$ & 0.102 \\
    & H-AE          &  8 & 256 & - & 7.9M & $3.5 \times 10^{-4}$ & 0.148
\end{tabular}
\end{center}
\caption{\textbf{Test MSE and Cosine loss for H-(V)AE models trained on MNIST, Shrec17 and Protein Neighborhoods.} MSE and Cosine values are strongly correlated within datasets but not across datasets.}
\label{table:mse_and_cosine}
\end{table}

\begin{table}[ht]
\begin{center}
\resizebox{\textwidth}{!}{
\begin{tabular}{l l c c c c c c c | c | c c}
    \bf Dataset & \bf Method & \bf $L$ & \bf $z$ & \bf TP-type & $\mathbf{C_{init}}$ & \bf ChannelsList & \bf DegreesList & \bf \# Params & \bf Speed & \bf MSE & \bf Cosine \\
    \hline
    \hline
    \multirow{3}{*}{MNIST}
    & H-AE  & 10 & 16 &     ETP & None &      [16,16,16,16,16,16] & [10,10,8,4,2,1] & 227k & \textbf{1.0} & $\mathbf{9.3 \times 10^{-4}}$ & \textbf{0.025}  \\
    & H-AE  & 10 & 16 & Full-TP & None &             [7,6,6,6,16] &    [10,8,4,2,1] & 229k & 1.1 & $1.6 \times 10^{-3}$ & 0.044  \\
    & H-AE  & 10 & 16 & Full-TP & None &           [5,5,5,5,7,16] & [10,10,8,4,2,1] & 227k & 1.7 & $1.5 \times 10^{-3}$ & 0.041  \\
    \hline
    \multirow{2}{*}{Shrec17}
    & H-AE  & 14 & 40 &     ETP &   12 &  [12,12,12,20,24,32,40] &   [14,14,14,8,4,2,1] & 518k & 1.0 & $\mathbf{1.8 \times 10^{-4}}$ & \textbf{0.130} \\
    & H-AE  & 14 & 40 & Full-TP & None &            [5,5,5,8,40] &         [14,8,4,2,1] & 518k & \textbf{0.9} & $1.9 \times 10^{-4}$ & 0.137    \\
    & H-AE  & 14 & 40 & Full-TP & None &        [4,3,3,6,6,6,40] &   [14,14,14,8,4,2,1] & 513k & 1.9 & $2.0 \times 10^{-4}$ & 0.142    \\
    \hline
    \multirow{4}{*}{Protein NBs}
    & H-AE  & 6 & 64 &     ETP &   44 &  [50,50,50,64,64,64] &   [6,6,6,4,2,1] & 1.6M & \textbf{1.0} & $\mathbf{5.4 \times 10^{-4}}$ & \textbf{0.219} \\
    & H-AE  & 6 & 64 & Full-TP & None &          [8,8,12,36] &       [6,4,2,1] & 1.6M & 2.1 & $6.7 \times 10^{-4}$ & 0.286 \\
    \cline{2-12}
    & H-AE  & 8 & 64 &     ETP &   44 &  [50,50,50,64,64,64] &   [6,6,6,4,2,1] & 2.4M & \textbf{1.0} & $\mathbf{6.3 \times 10^{-4}}$ & \textbf{0.287} \\
    & H-AE  & 8 & 64 & Full-TP & None &          [8,8,12,36] &       [6,4,2,1] & 2.6M & 3.3 & $7.6 \times 10^{-4}$ & 0.371 \\
\end{tabular}
}
\end{center}
\caption{\textbf{Training speed and reconstruction ablations of H-(V)AE models with different Tensor Product rules.} To make comparison fair, models were trained using the same training hyperparameters as described in~\ref{sec:exp_details}, and all models were constructed to have comparable number of parameters. Speed was computed as training time and divided by the time of the model using ETP within each dataset. Models with the ETP consistently generates better reconstructions and are usually the fastest. The speed and performance gains of the ETP are most apparent on the Protein Neighborhoods task, where we also note that, as the angular resolution of the data ($L$) is increased from 6 to 8, the relative speed gain of the ETP over the Full-TP is significantly accentuated (from 2.1x to 3.3x).}
\label{table:tp_ablation}
\end{table}

\begin{table}[ht]
\begin{center}
\resizebox{0.8\textwidth}{!}{
\begin{tabular}{l l c c c | c c}
    \bf Dataset & \bf Method & \bf $L$ & \bf z & \bf \# train & \bf Equiv. Error & \bf Abs. Value \\
    \hline
    \hline
    \multirow{2}{*}{MNIST}
    & H-AE NR/R     & 10 &  16 & - & (2.3 $\pm$ 2.7)$\times 10^{-4}$& (1.0 $\pm$ 0.9)$\times 10^{-1}$ \\
    & H-VAE NR/R    & 10 &  16 & - & (2.5 $\pm$ 1.9)$\times 10^{-4}$& (1.3 $\pm$ 1.8)$\times 10^{-1}$ \\
    \hline
    \multirow{2}{*}{Shrec17}
    & H-AE          & 14 &  40 & - & (1.4 $\pm$ 2.4)$\times 10^{-5}$& (0.4 $\pm$ 1.1)$\times 10^{-2}$ \\
    & H-VAE         & 14 &  40 & - & (1.4 $\pm$ 3.5)$\times 10^{-5}$& (0.4 $\pm$ 2.2)$\times 10^{-2}$ \\
    \hline
    \multirow{4}{*}{Toy Aminoacids}
    & H-AE                   &  4 &  2 &  1,000 & (4.7 $\pm$ 3.7)$\times 10^{-5}$& (2.6 $\pm$ 4.1)$\times 10^{-2}$ \\
    & H-VAE $\beta = 0.025$  &  4 &  2 &  1,000 & (5.4 $\pm$ 3.9)$\times 10^{-5}$& (2.8 $\pm$ 4.0)$\times 10^{-2}$ \\
    & H-AE                   &  4 &  2 & 20,000 & (9.3 $\pm$ 6.4)$\times 10^{-5}$& (3.5 $\pm$ 4.7)$\times 10^{-2}$ \\
    & H-VAE $\beta = 0.025$  &  4 &  2 & 20,000 & (9.0 $\pm$ 4.7)$\times 10^{-5}$& (3.0 $\pm$ 4.3)$\times 10^{-2}$ \\
    \hline
    \multirow{12}{*}{Protein NBs}
    & H-AE          &  3 &  64 & - & (3.3 $\pm$ 4.2)$\times 10^{-5}$& (1.6 $\pm$ 2.1)$\times 10^{-2}$ \\
    & H-AE          &  4 &  64 & - & (0.6 $\pm$ 0.6)$\times 10^{-5}$& (0.6 $\pm$ 2.1)$\times 10^{-2}$ \\
    & H-AE          &  6 &  64 & - & (0.9 $\pm$ 1.0)$\times 10^{-5}$& (0.4 $\pm$ 1.3)$\times 10^{-2}$ \\
    & H-AE          &  8 &  64 & - & (0.7 $\pm$ 1.0)$\times 10^{-5}$& (0.2 $\pm$ 0.9)$\times 10^{-2}$ \\
    \cline{2-7}
    & H-AE          &  3 & 128 & - & (1.1 $\pm$ 1.3)$\times 10^{-5}$& (1.0 $\pm$ 3.0)$\times 10^{-2}$ \\
    & H-AE          &  4 & 128 & - & (0.6 $\pm$ 0.7)$\times 10^{-5}$& (0.6 $\pm$ 2.5)$\times 10^{-2}$ \\
    & H-AE          &  6 & 128 & - & (0.1 $\pm$ 0.1)$\times 10^{-5}$& (0.2 $\pm$ 1.7)$\times 10^{-2}$ \\
    & H-AE          &  8 & 128 & - & (0.1 $\pm$ 0.1)$\times 10^{-5}$& (0.1 $\pm$ 1.0)$\times 10^{-2}$ \\
    \cline{2-7}
    & H-AE          &  3 & 256 & - & (0.6 $\pm$ 0.4)$\times 10^{-5}$& (0.6 $\pm$ 2.0)$\times 10^{-2}$ \\
    & H-AE          &  4 & 256 & - & (0.4 $\pm$ 0.6)$\times 10^{-5}$& (0.6 $\pm$ 2.2)$\times 10^{-2}$ \\
    & H-AE          &  6 & 256 & - & (0.1 $\pm$ 0.1)$\times 10^{-5}$& (0.2 $\pm$ 1.6)$\times 10^{-2}$ \\
    & H-AE          &  8 & 256 & - & (0.09 $\pm$ 0.07)$\times 10^{-5}$& (0.1 $\pm$ 1.3)$\times 10^{-2}$ \\
\end{tabular}
}
\end{center}
\caption{\textbf{Mean equivariance error for some of our trained H-(V)AE models.} Errors were computed over 2,000 randomly sampled spherical tensors, each with a randomly sampled rotation. Standard deviation is shown alongside the mean. We also show the mean and standard deviation of the absolute value of the output coefficients, to enable contextualization of the measured equivariance error. The equivariance error due to numerical error (absolute difference in coefficients by rotating input vs. output tensor) is consistently three orders of magnitude lower than the typical absolute value of the coefficients, indicating that equivariance is preserved. The same trend occurs for \textit{untrained} models (not shown here for simplicity).}
\label{table:equivariance_error}
\end{table}

\begin{table*}[t]
\begin{center}
\resizebox{0.8\textwidth}{!}{
\begin{tabular}{l | c c c c}
     & \multicolumn{3}{c}{\newlinecell{\textbf{Ligand Binding Affinity}\\\textbf{30\% Similarity}}} \\
    {\bf Method} & {RMSD $\downarrow$} & {Pearson's r $\uparrow$} & {Spearman's r $\uparrow$} & {Kendall's $\tau$ $\uparrow$} \\
    \hline
    \hline
    DeepDTA~\cite{ozturk_deepdta_2018} &                  $1.565$           & $0.573$           & $0.574$           & - \\ 
    DeepAffinity~\cite{karimi_deepaffinity_2019} &             $1.893 \pm 0.650$ & $0.415$           & $0.426$           & - \\
    Cormorant~\cite{anderson_cormorant_2019} &                $1.568 \pm 0.012$ & $0.389$           & $0.408$           & - \\
    ProtTrans~\cite{elnaggar_prottrans_2022} &                $1.544 \pm 0.015$ & $0.438 \pm 0.053$ & $0.434 \pm 0.058$ & - \\
    3DCNN~\cite{townshend_atom3d_2021} &                    $1.414 \pm 0.021$ & $0.550$           & $0.553$           & - \\
    GNN~\cite{townshend_atom3d_2021} &                      $1.570 \pm 0.025$ & $0.545$           & $0.533$           & - \\
    MaSIF~\cite{gainza_deciphering_2020} &                    $1.484 \pm 0.018$ & $0.467 \pm 0.020$ & $0.455 \pm 0.014$ & - \\
    DGAT~\cite{nguyen_graphdta_2021} &                     $1.719 \pm 0.047$ & $0.464$           & $0.472$           & - \\
    DGIN~\cite{nguyen_graphdta_2021} &                     $1.765 \pm 0.076$ & $0.426$           & $0.432$           & - \\
    DGAT-GCN~\cite{nguyen_graphdta_2021} &                 $1.550 \pm 0.017$ & $0.498$           & $0.496$           & - \\
    GVP-GNN~\cite{jing_learning_2021} &                  $1.648 \pm 0.014$ & $0.213 \pm 0.013$ & $0.164 \pm 0.009$ & $0.110 \pm 0.012$ \\
    EGNN~\cite{satorras_en_2022} &                     $1.492 \pm 0.012$ & $0.489 \pm 0.017$ & $0.472 \pm 0.008$ & $0.329 \pm 0.014$ \\
    HoloProt~\cite{somnath_multi-scale_2022} &                 $1.464 \pm 0.006$ & $0.509 \pm 0.002$ & $0.500 \pm 0.005$ & - \\
    GBPNet~\cite{aykent_gbpnet_2022} &                   $1.405 \pm 0.009$ & $0.561$           & $0.557$           & - \\
    EGNN + PLM~\cite{wu_when_2022} &               $1.403 \pm 0.013$ & $0.565 \pm 0.016$ & $0.544 \pm 0.005$ & $0.379 \pm 0.007$ \\
    ProtMD~\cite{wu_pre-training_2022}   &     $1.367 \pm 0.014$ & \underline{$0.601 \pm 0.036$} & \underline{$0.587 \pm 0.042$} & - \\
    \hline
    Zernike Inv. + Linear Regression & $1.455 \pm 0.005$ & $0.513 \pm 0.005$ & $0.516 \pm 0.006$ & $0.357 \pm 0.005$ \\
    Zernike Inv. + Random Forest & \underline{$1.361 \pm 0.011$} & $0.587 \pm 0.009$ & $0.584 \pm 0.010$ & \underline{$0.408 \pm 0.008$} \\
    H-AE + Linear Regression & $1.397 \pm 0.019$ & $0.560 \pm 0.017$ & $0.568 \pm 0.018$ & $0.397 \pm 0.016$ \\
    H-AE + Random Forest &     $\mathbf{1.332 \pm 0.012}$ & $\mathbf{0.612 \pm 0.009}$ & $\mathbf{0.619 \pm 0.009}$ & $\mathbf{0.436 \pm 0.006}$
\end{tabular}
}
\end{center}
\caption{{\bf Comprehensive benchmarking results on the Ligand Binding Affinity task with Atom3D's 30\% similarity split.} Models are sorted by date of release. In addition to the H-AE infomed models, we also report the performance of baseline models that only use the SO(3)-invariant ($\ell = 0$) component of each neighborhood's Zernike transform (Zernike Inv.). H-AE consistently outperforms this baseline, indicating that the SO(3)-invariant information from the higher-degree features extracted by H-AE are informative for this regression task. Best scores are in \textbf{bold} and second-best scores are \underline{underlined}. Errors for our models are computed as the standard deviation in prediction by 10 Machine Learning models trained with bootstrapped data.}
\label{table:lba_results_appendix}
\end{table*}

\clearpage{}

\newpage{}

\begin{figure}
    \centering
    \begin{tabular}{c}
        \includegraphics[width=0.8\textwidth]{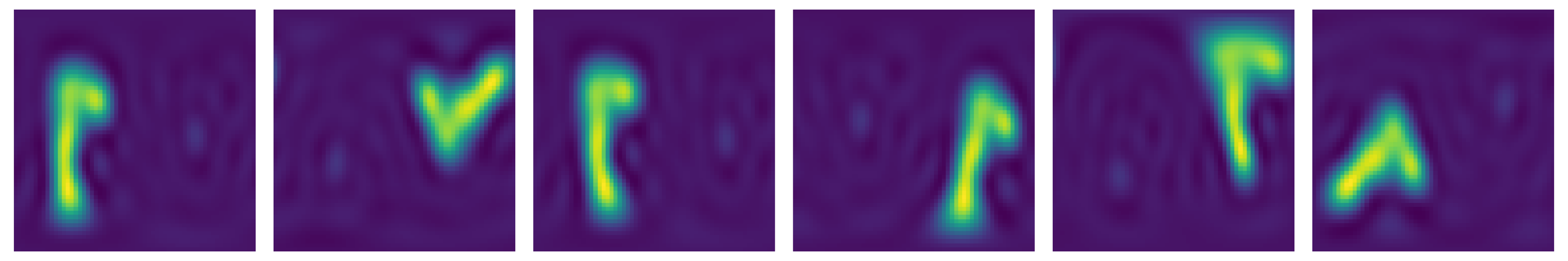} \\
        \includegraphics[width=0.8\textwidth]{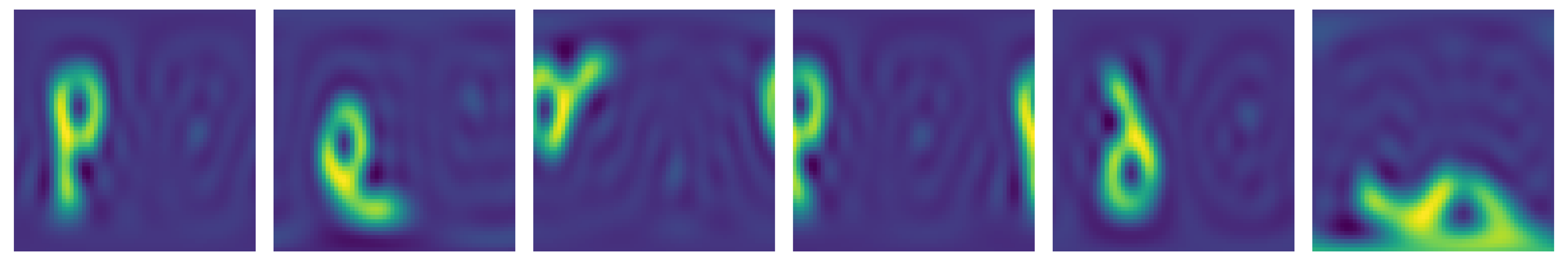} \\
        \includegraphics[width=0.8\textwidth]{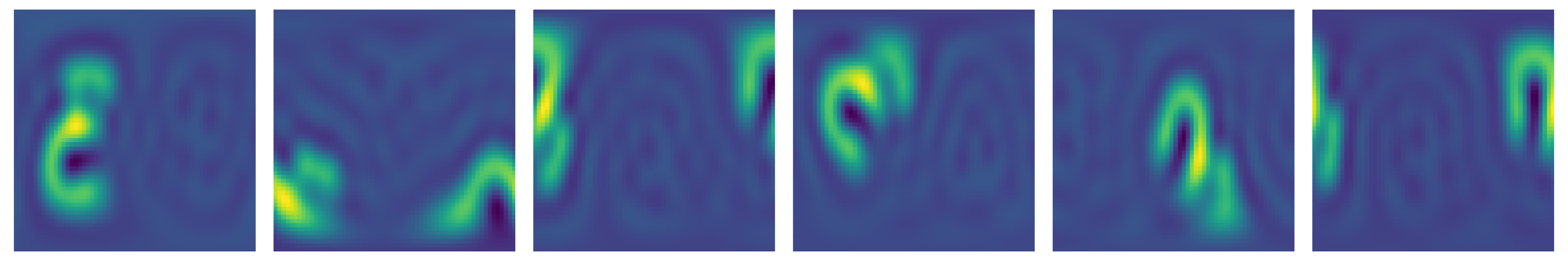}
    \end{tabular}
    \caption{\textbf{Visual proof of the disentanglement in the latent space of MNIST-on-the-sphere.} For each row, the invariant embedding $\mathbf{z}$ is held fixed, and a different frame (i.e., the rotation matrix) is used. Frames are sampled randomly and differ across rows, with the exception of the first column, which is always the identity frame. Then, $\mathbf{z}$ and the frame are fed to the decoder and the Inverse Fourier Transform is used to generate the reconstructed spherical image, which is projected onto a plane for the ease of visualization. Modulo the distortions given by projecting the image onto a plane, it is clear that the invariant embedding contains all semantic information, and the frame solely determines the orientation of the image.\\\\\\}
    \label{fig:disentanglement_mnist}
\end{figure}

\begin{figure}[h!]
    \centering
    \includegraphics[width=0.8\textwidth]{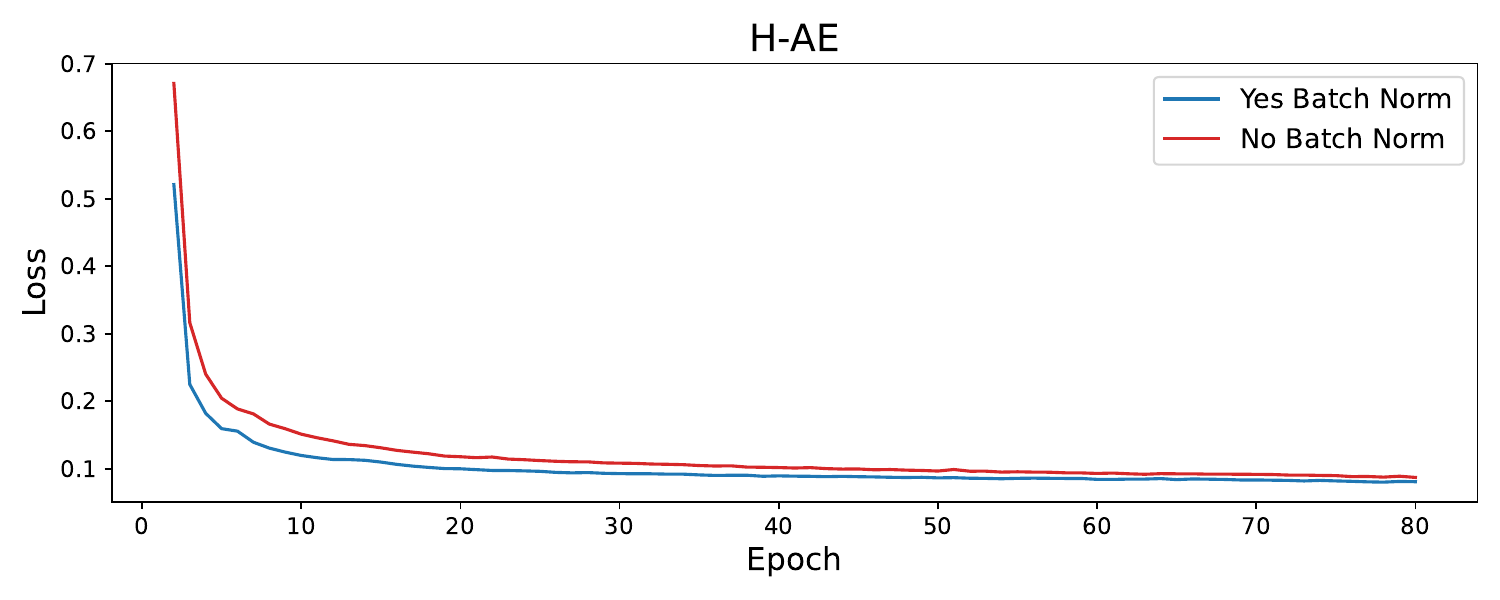}
    \caption{\textbf{Training loss trace of H-AE, with and without Batch Norm, on MNIST-on-the-sphere.} Models were trained with the (NR/R; z = 16; AE) specification. The loss on validation data follows the same trend, but it is not shown for simplicity.}
    \label{fig:training_loss_trace_H-AE}
\end{figure}


\begin{figure}[h]
    \centering
    \includegraphics[width=0.6\textwidth]{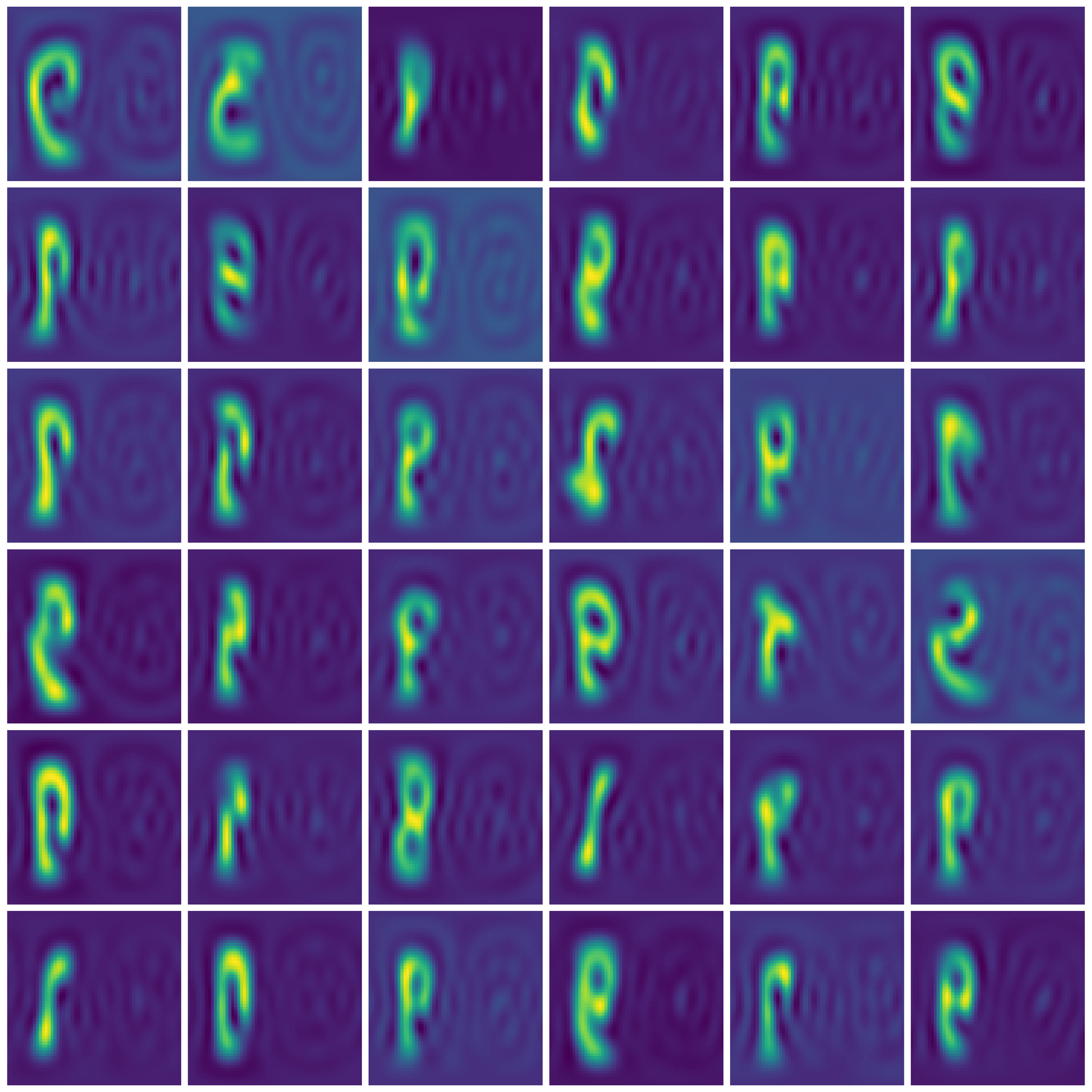}
    \caption{\textbf{Random samples generated by the (NR/R; z = 16; VAE) MNIST-on-the-sphere model.} We sample invariant latent embeddings from the prior distribution (isotropic normal) and feed them to the decoder alongside the canonical frame to generate tensors. We then compute the inverse SFT to map the generated tensor to images in real space. The samples show a wide range of diversity in digit and style.
    }
    \label{fig:mnist_samples}
\end{figure}

\begin{figure}[h]
    \centering
    \resizebox{\textwidth}{!}{
    \begin{tabular}{c | c | c}
         \includegraphics[width=0.1\textwidth]{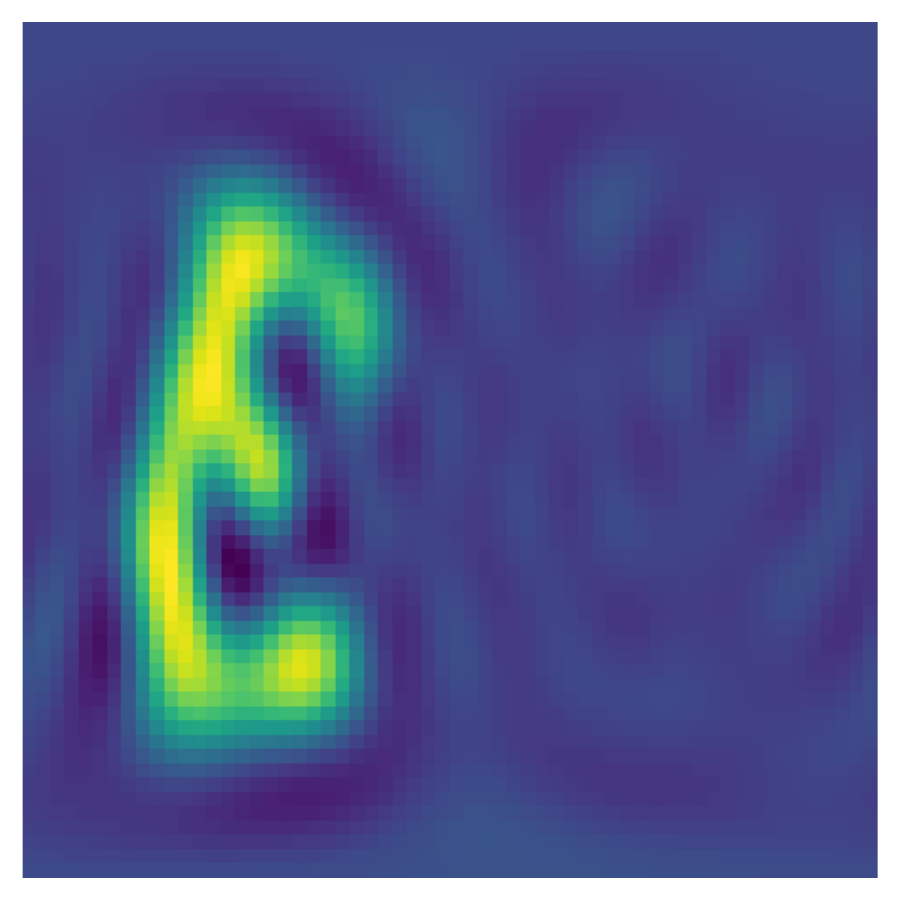} &
         \includegraphics[width=1.0\textwidth]{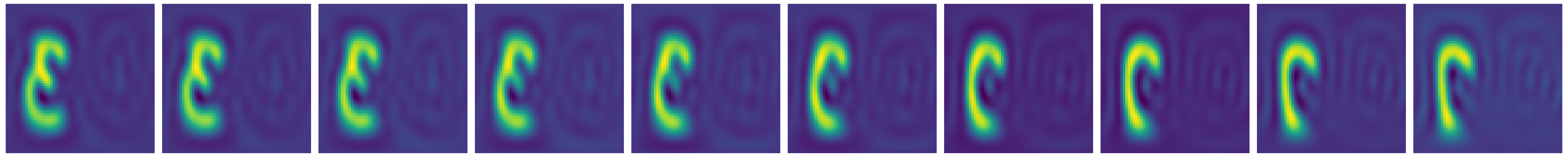} &
         \includegraphics[width=0.1\textwidth]{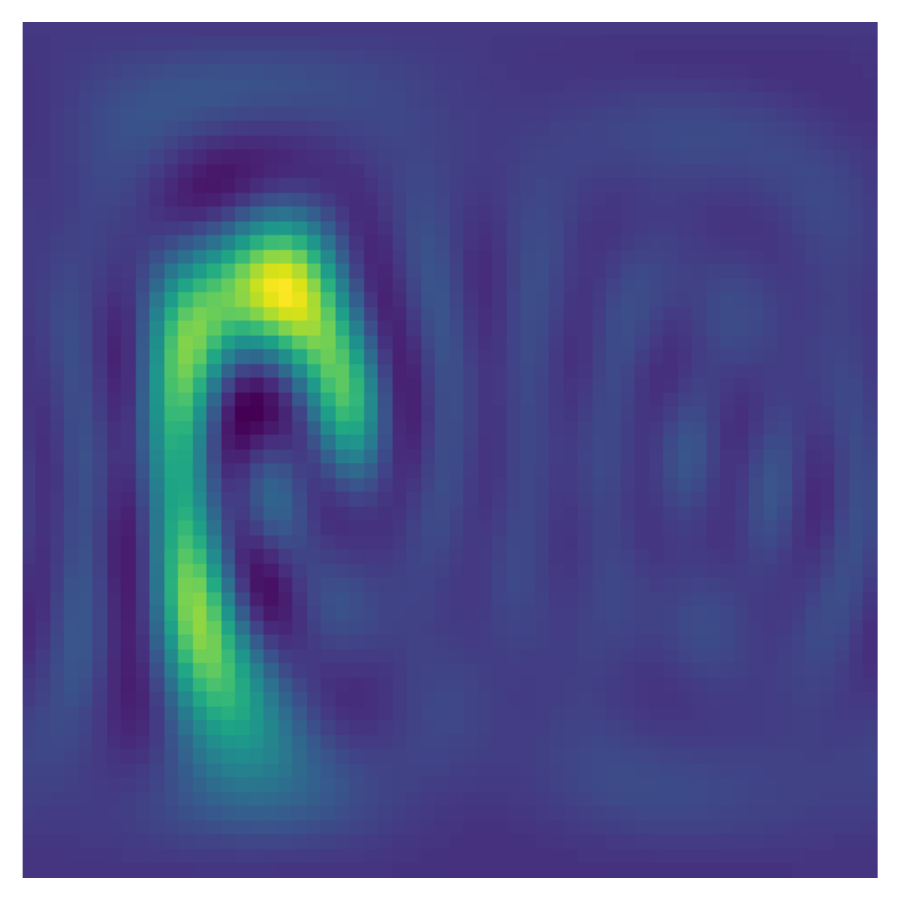}
         \\
         \includegraphics[width=0.1\textwidth]{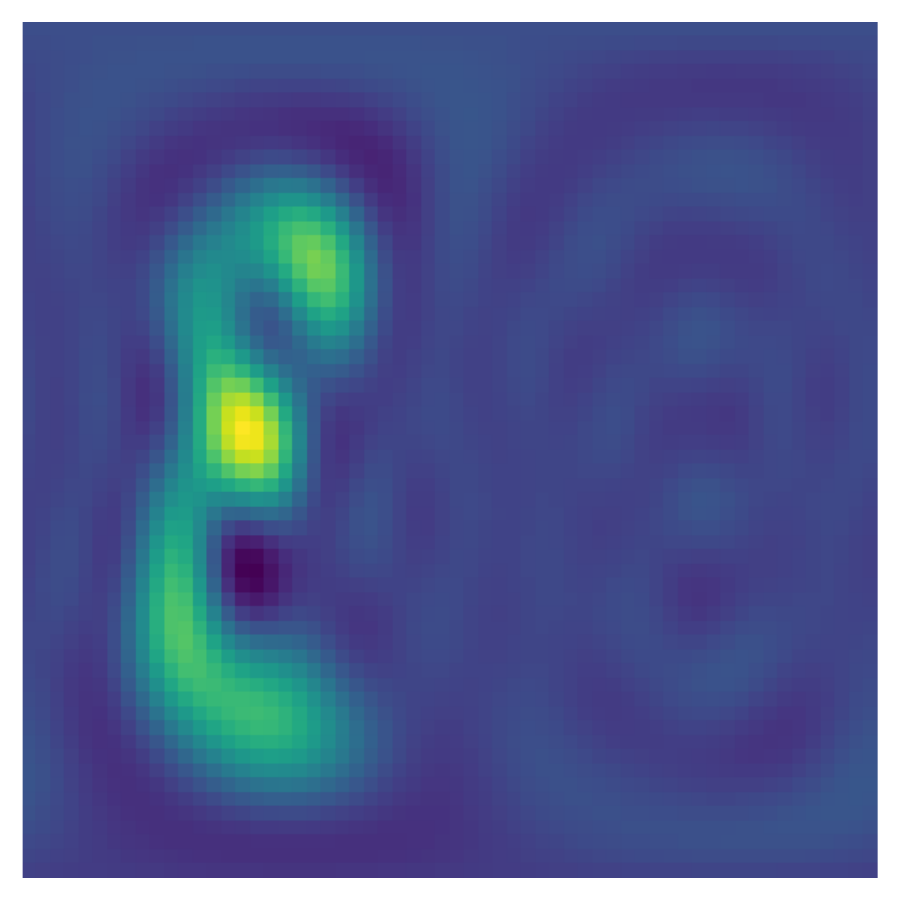} &
         \includegraphics[width=1.0\textwidth]{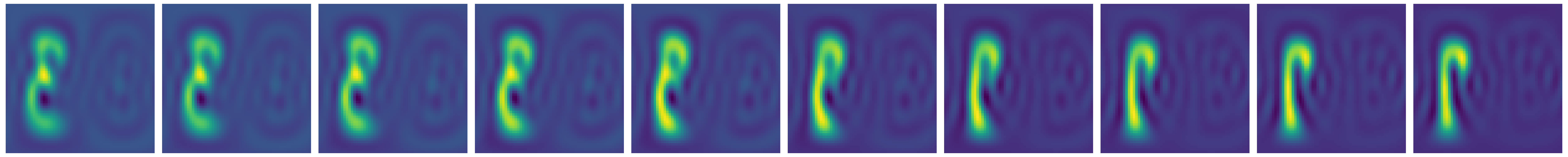} &
         \includegraphics[width=0.1\textwidth]{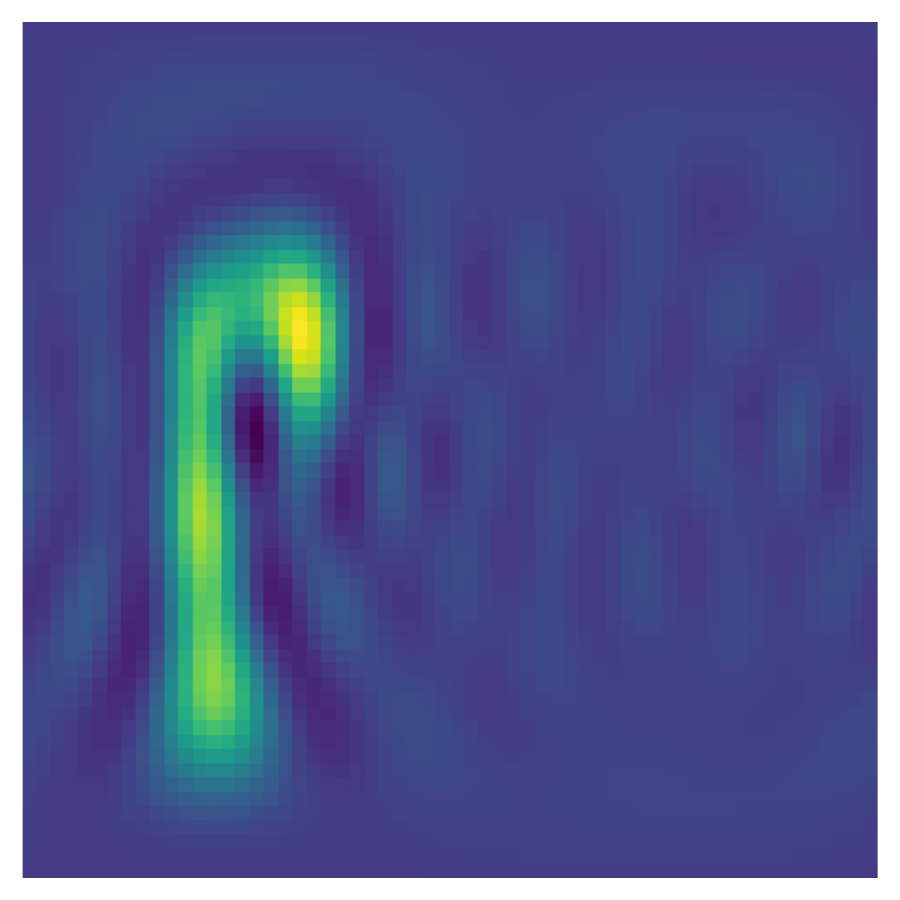}
         \\
         \includegraphics[width=0.1\textwidth]{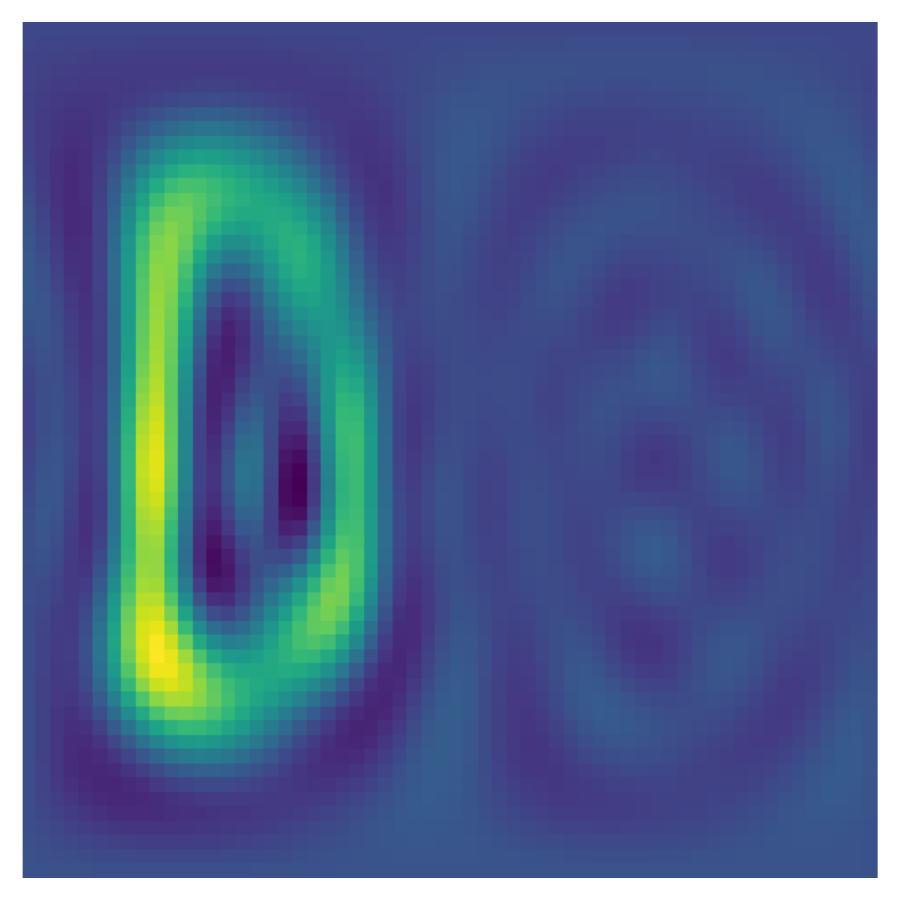} &
         \includegraphics[width=1.0\textwidth]{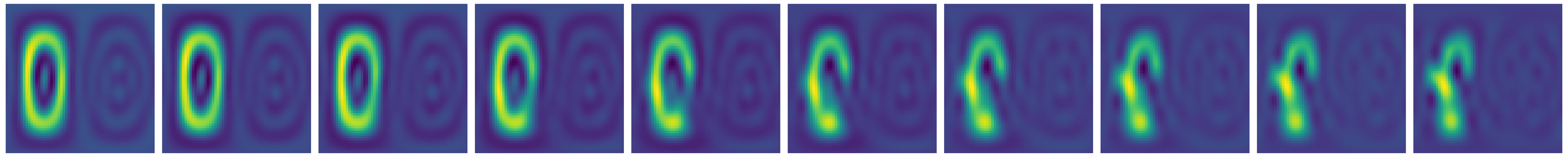} &
         \includegraphics[width=0.1\textwidth]{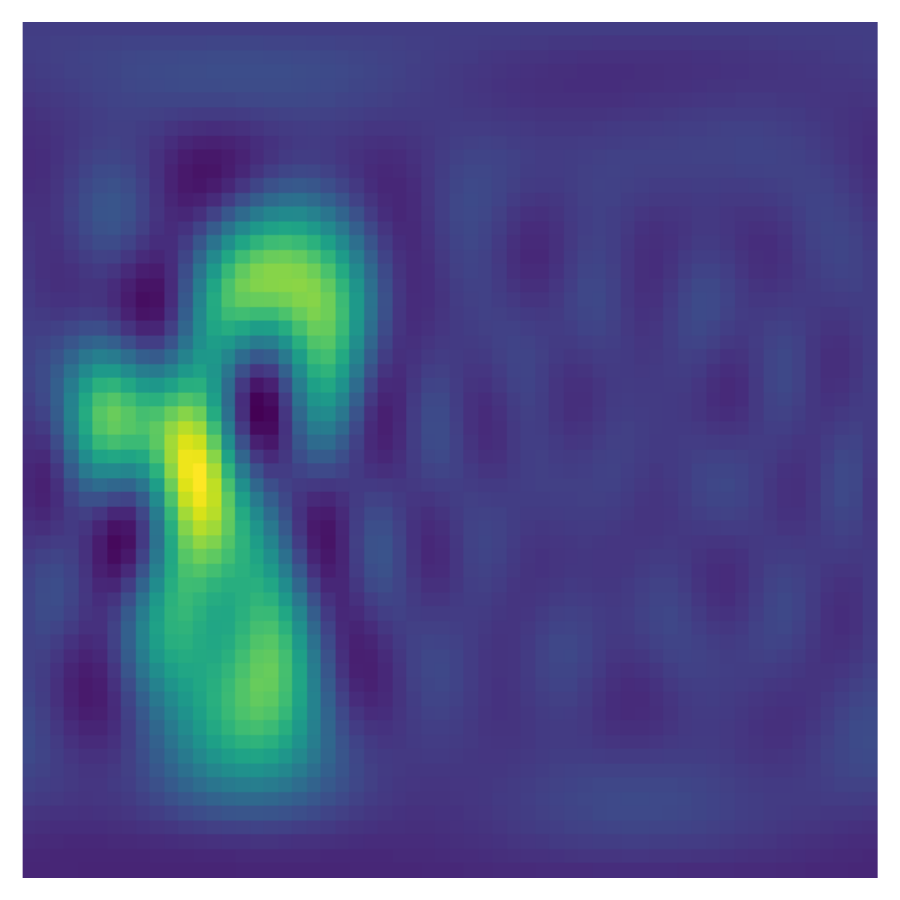}
         \\
         \includegraphics[width=0.1\textwidth]{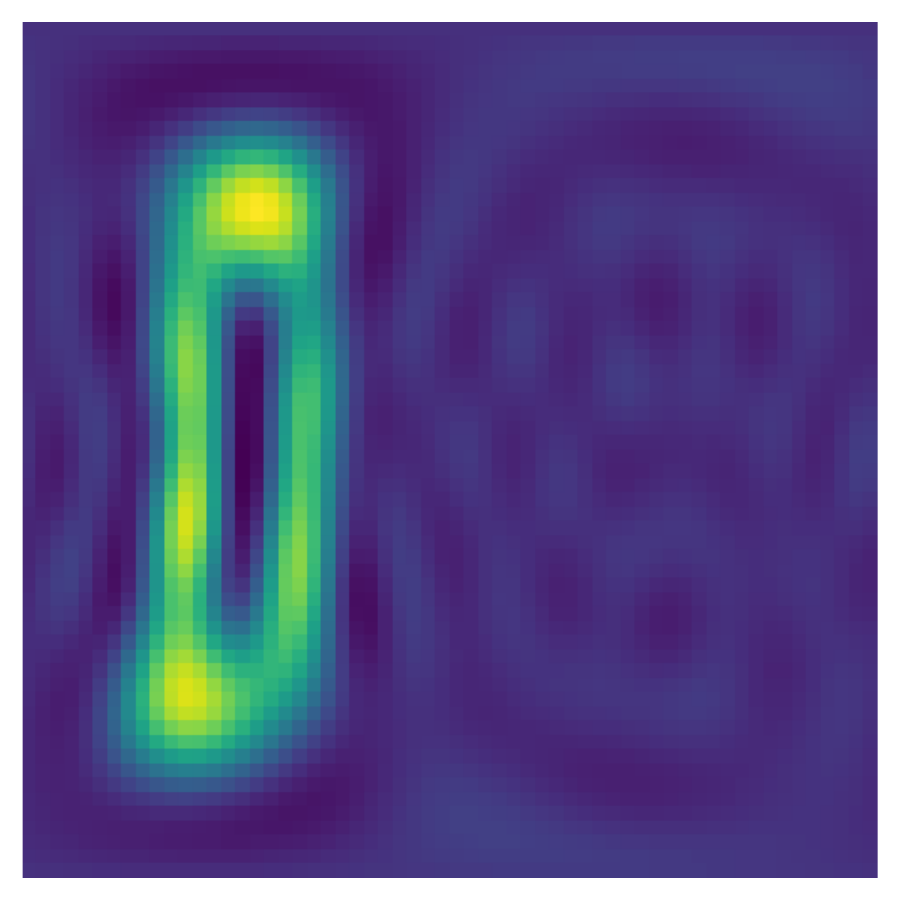} &
         \includegraphics[width=1.0\textwidth]{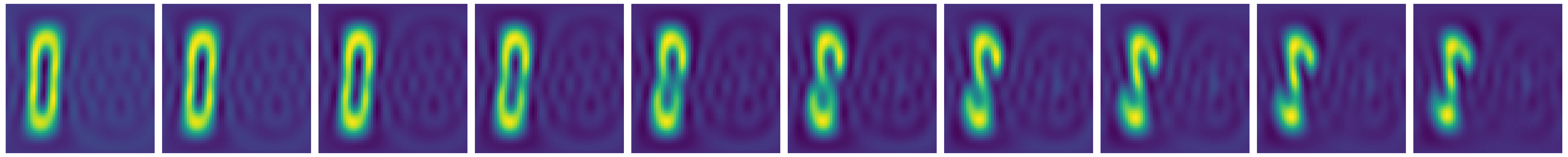} &
         \includegraphics[width=0.1\textwidth]{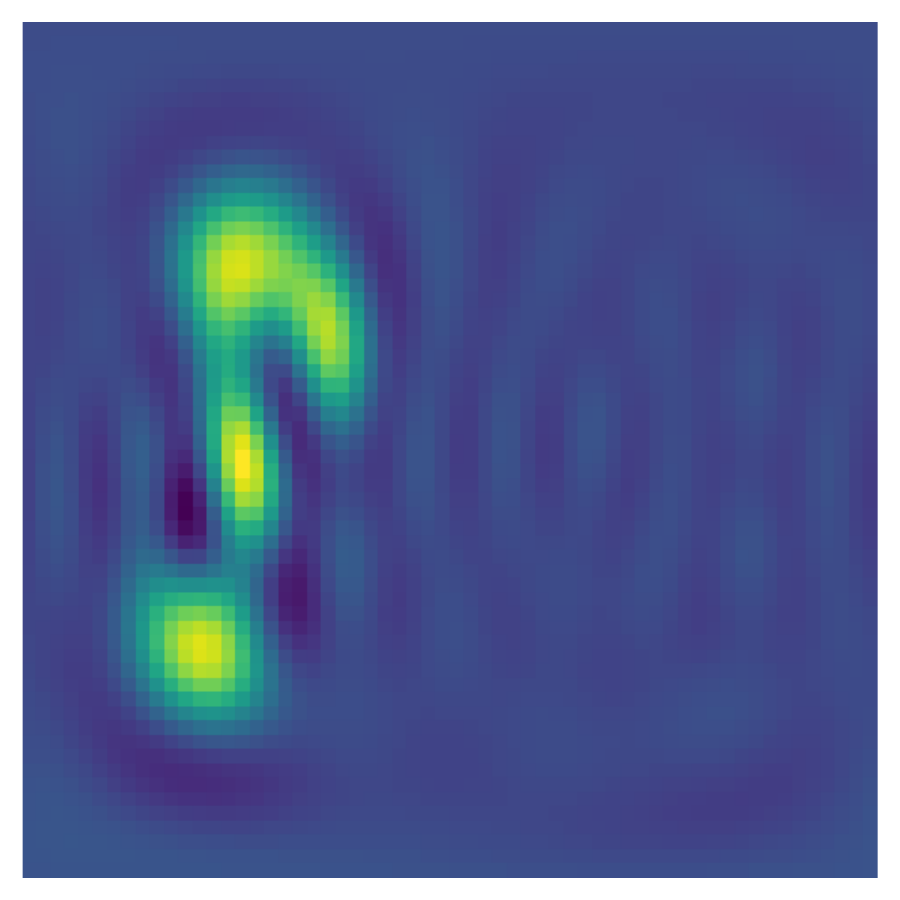}
         \\
         \includegraphics[width=0.1\textwidth]{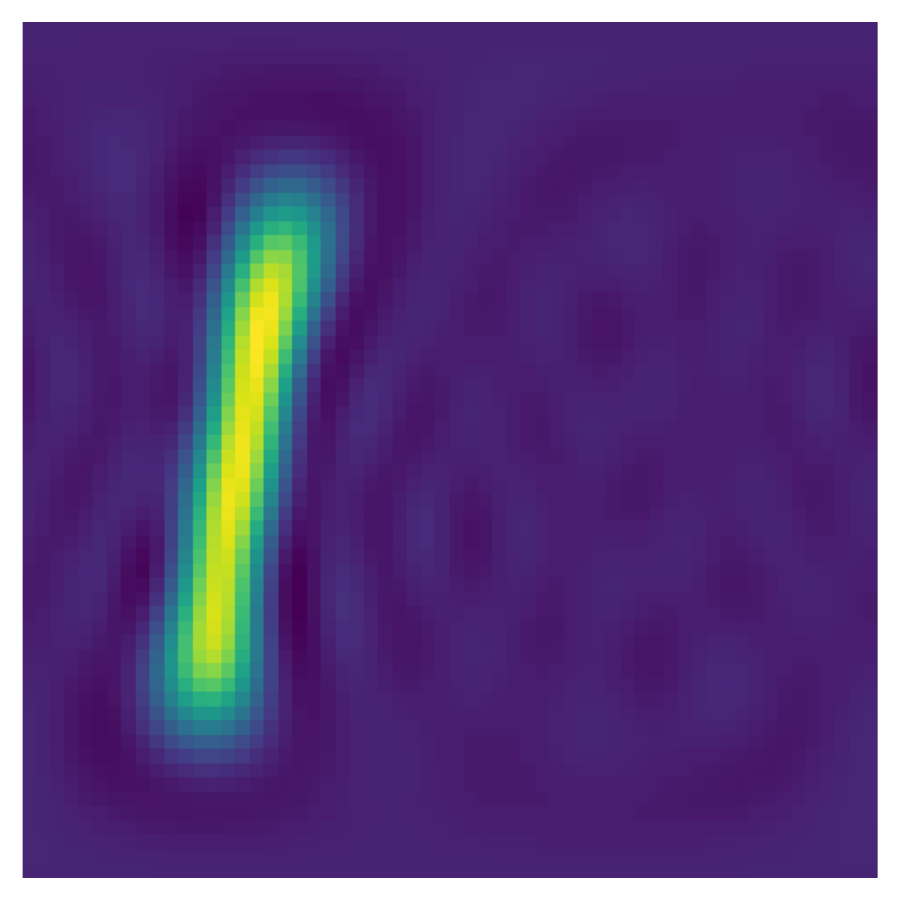} &
         \includegraphics[width=1.0\textwidth]{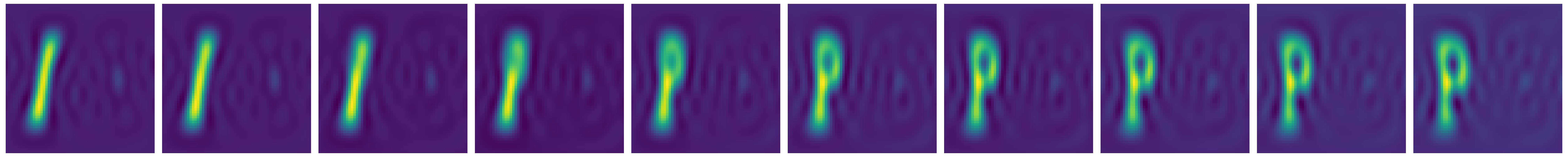} &
         \includegraphics[width=0.1\textwidth]{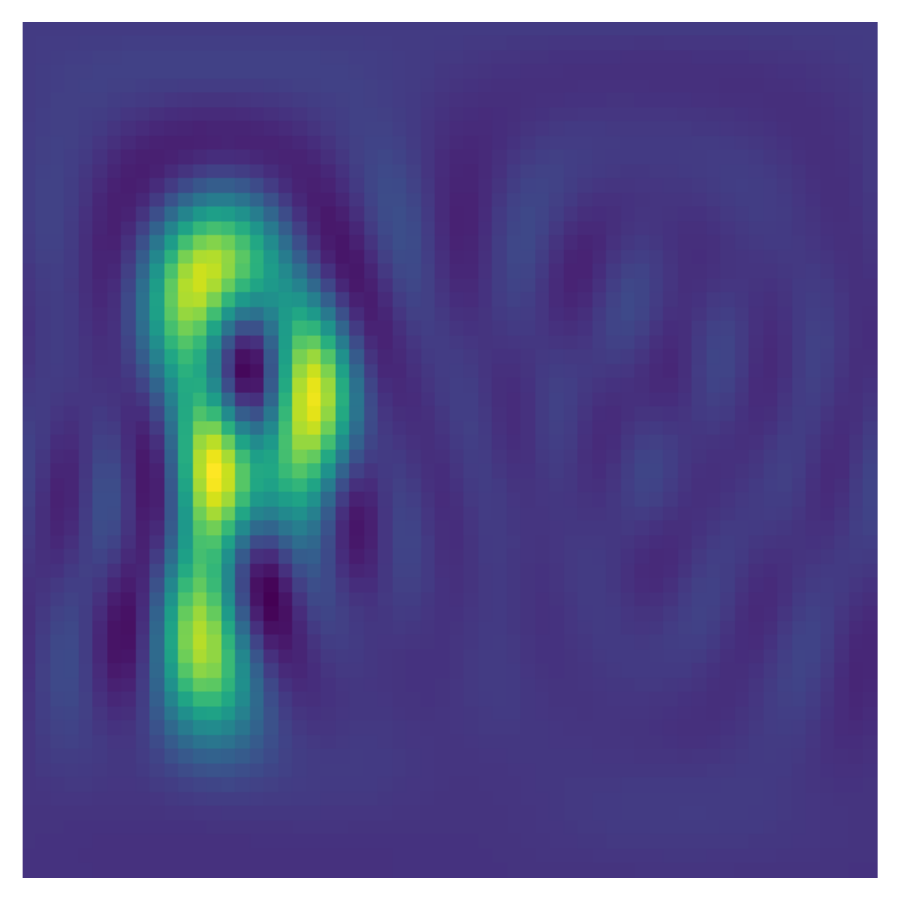}
         \\
         \includegraphics[width=0.1\textwidth]{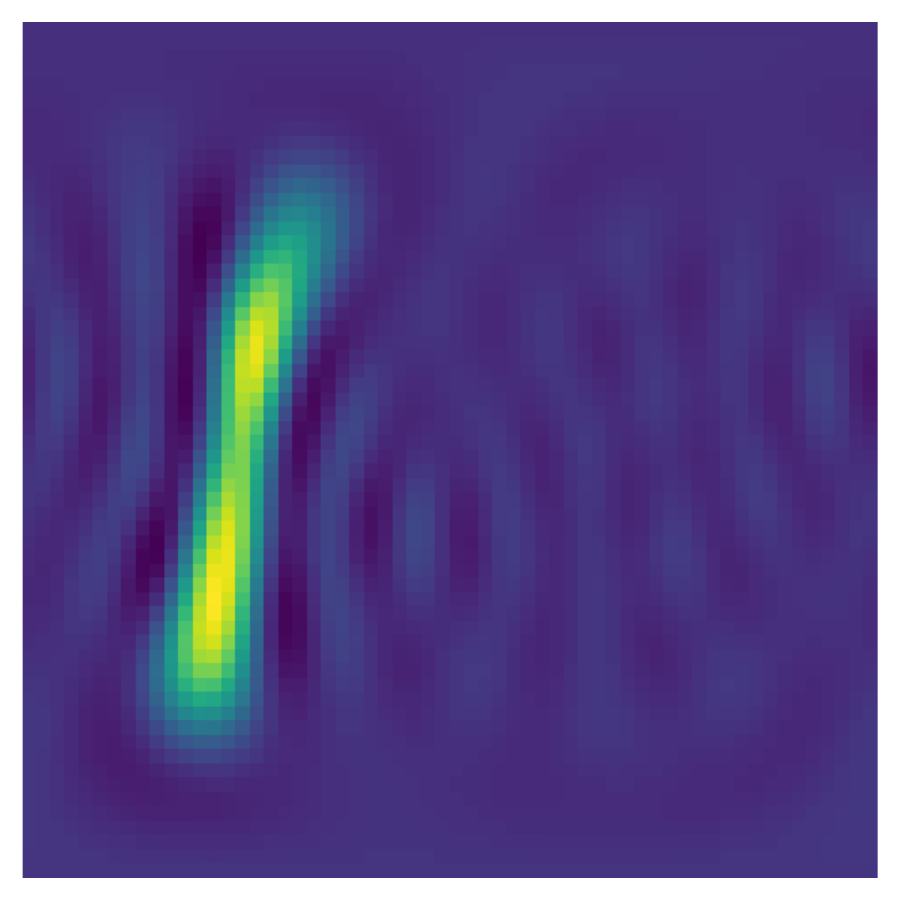} &
         \includegraphics[width=1.0\textwidth]{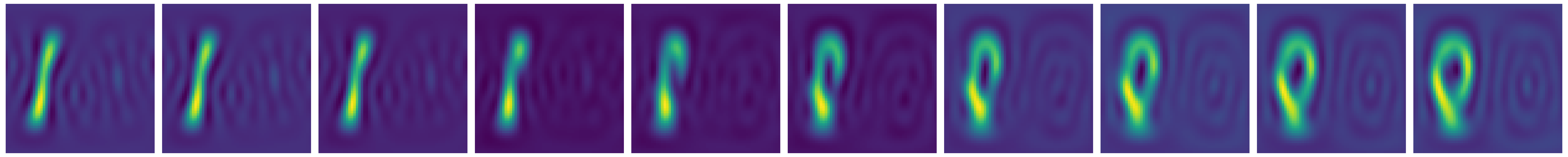} &
         \includegraphics[width=0.1\textwidth]{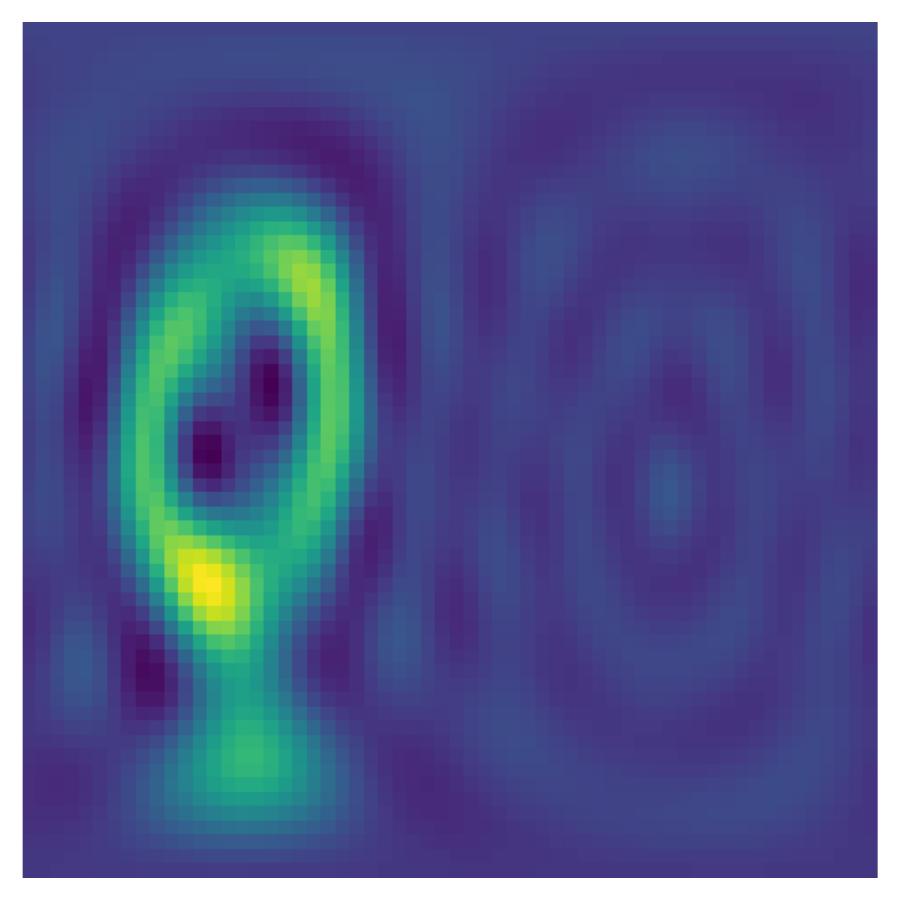}
         \\
         \includegraphics[width=0.1\textwidth]{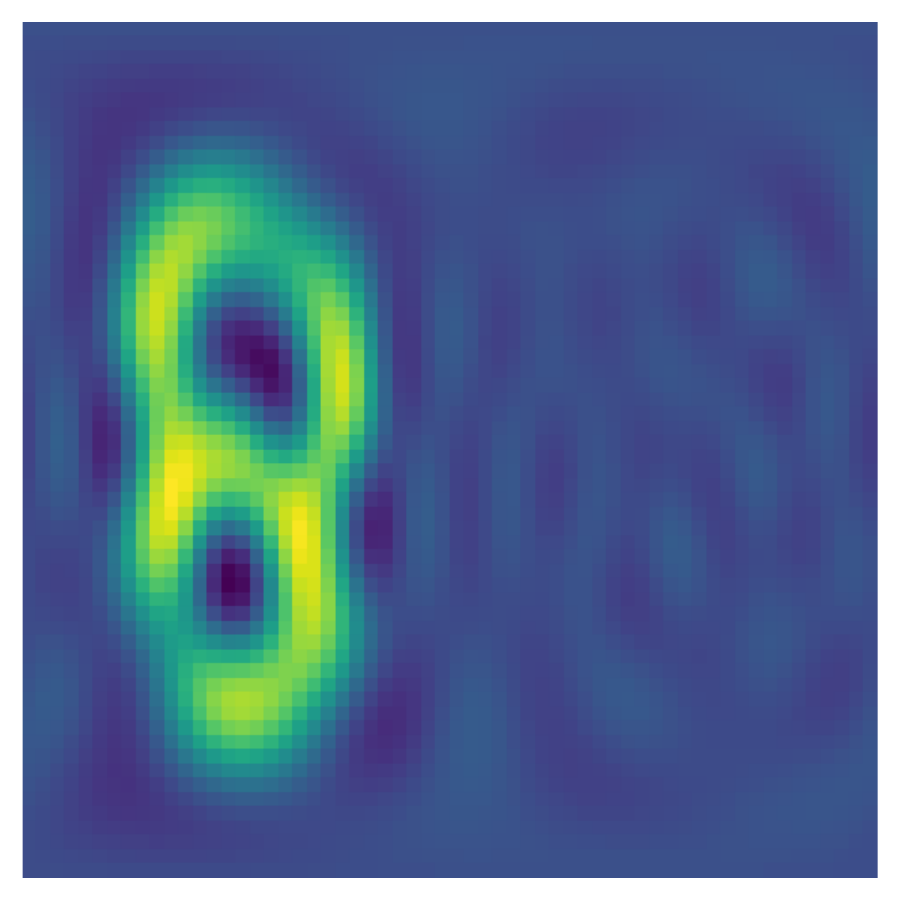} &
         \includegraphics[width=1.0\textwidth]{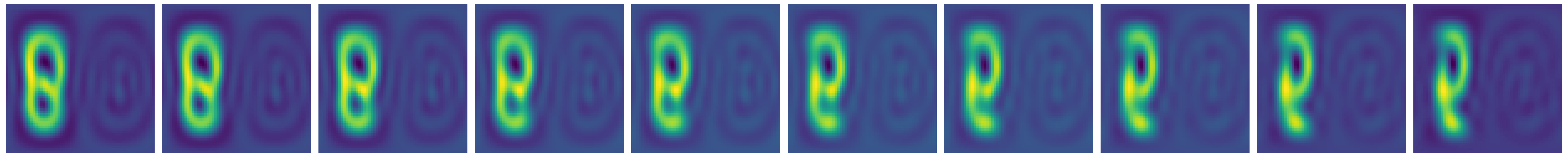} &
         \includegraphics[width=0.1\textwidth]{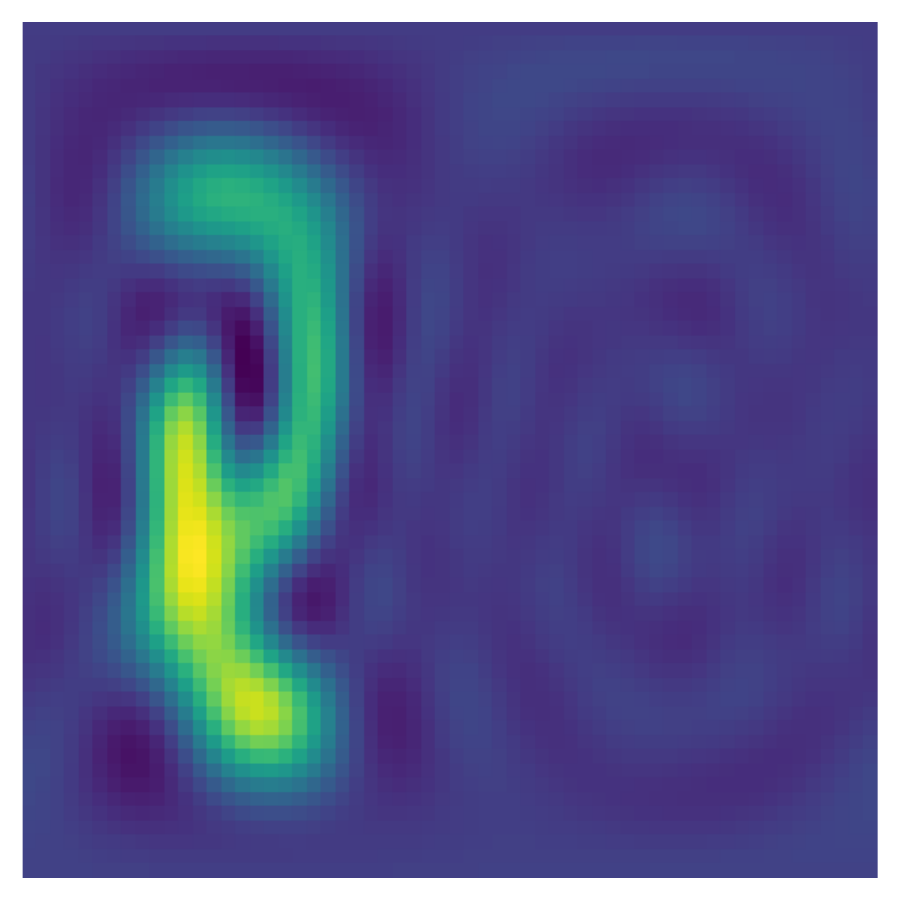}
         \\
         \includegraphics[width=0.1\textwidth]{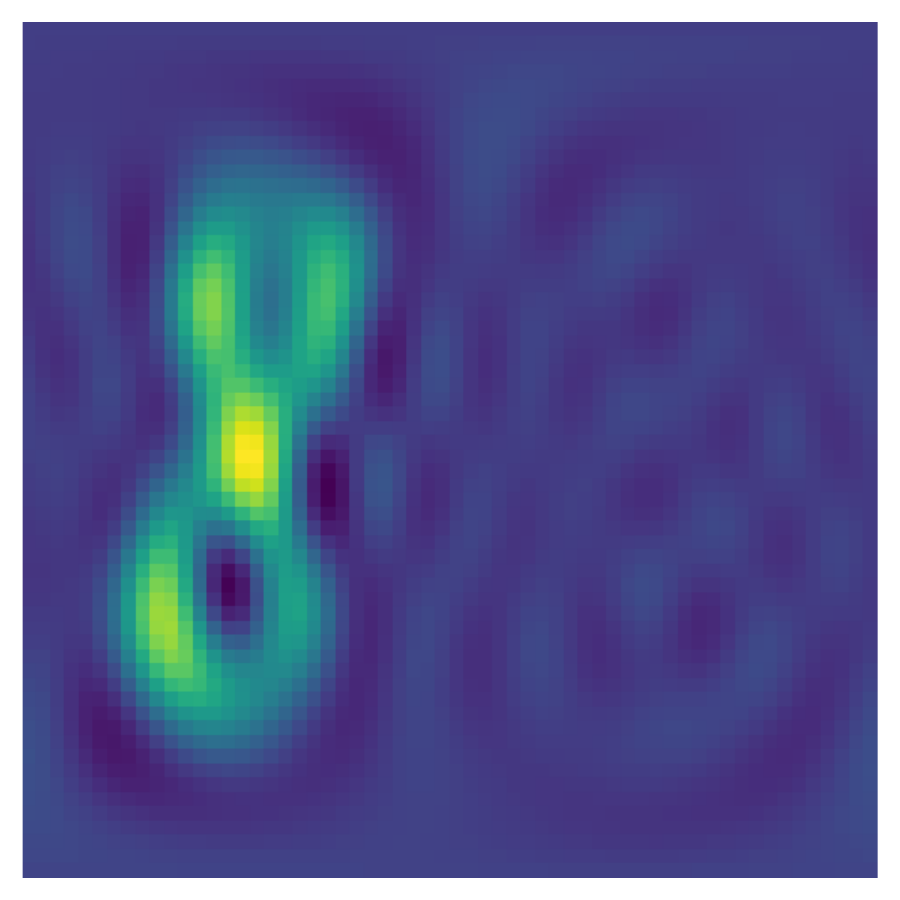} &
         \includegraphics[width=1.0\textwidth]{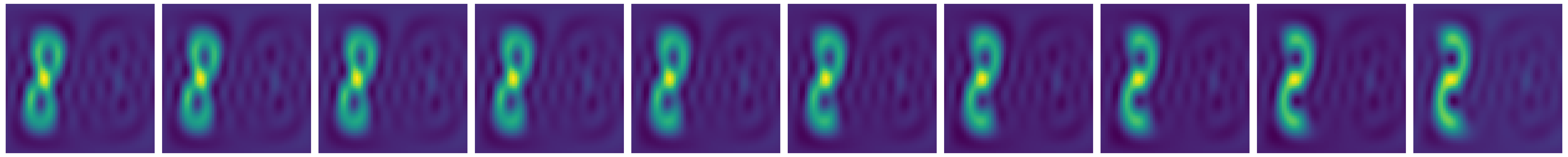} &
         \includegraphics[width=0.1\textwidth]{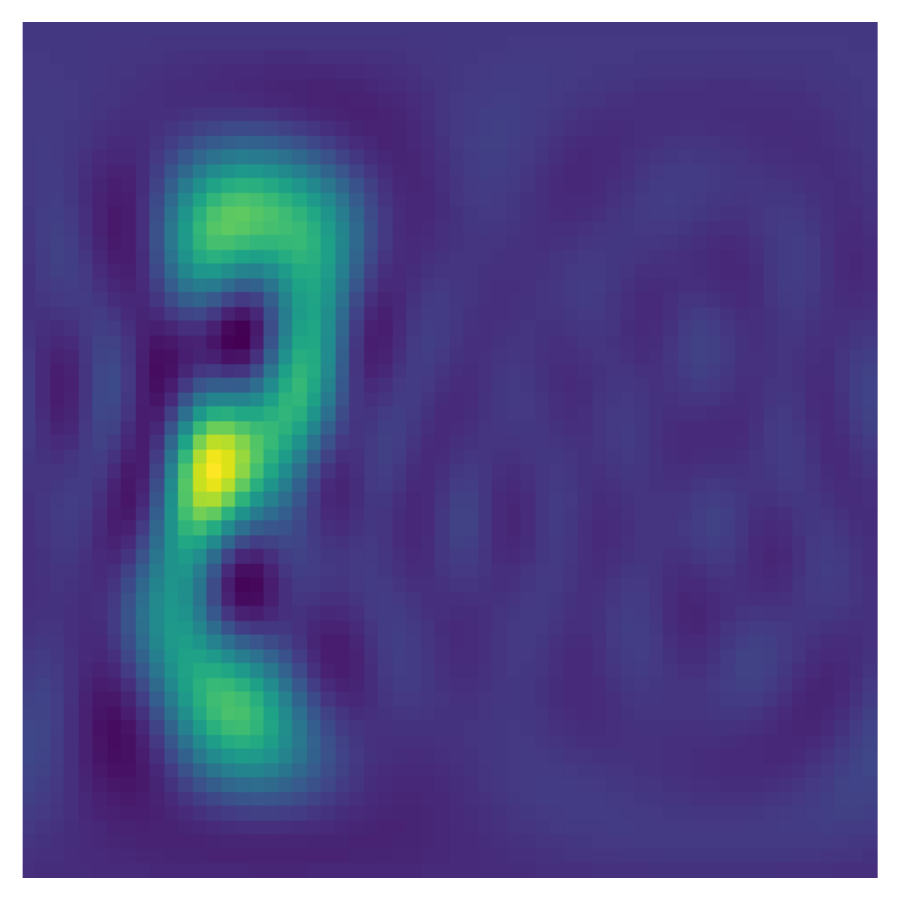}
    \end{tabular}}
    \caption{\textbf{Trajectories across the latent space for the (NR/R; z = 16; VAE) MNIST-on-the-sphere model.} We compute pairs of invariant latent embeddings using the model's encoder, and linearly interpolate between them through the latent space. We then feed the interpolated embeddings into the decoder, together with the canonical frame, and compute the inverse SFT to get the image in real space. The left and right columns show the original images (after forward and inverse SFT) rotated to be placed in the learned canonical frame, whereas the center rows show the interpolated images. We can see that all trajectories are smooth, respecting spatial consistency, sign of a well structured latent space.}
    \label{fig:mnist_trajectories}
\end{figure}

\begin{figure}[h]
    \centering
    \resizebox{\textwidth}{!}{
    \begin{tabular}{c c}
    \includegraphics[width=0.5\textwidth]{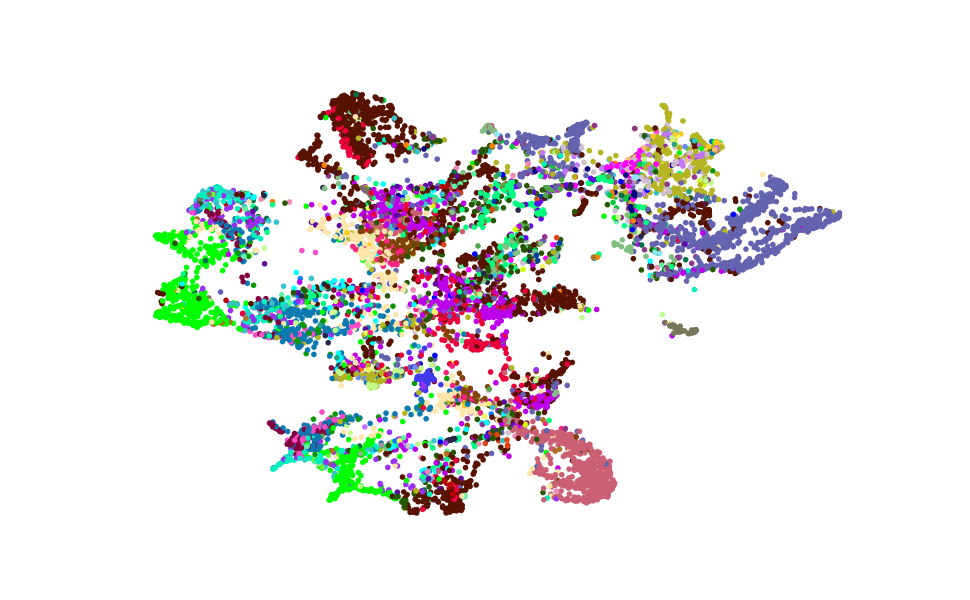}
    &
    \includegraphics[width=0.5\textwidth]{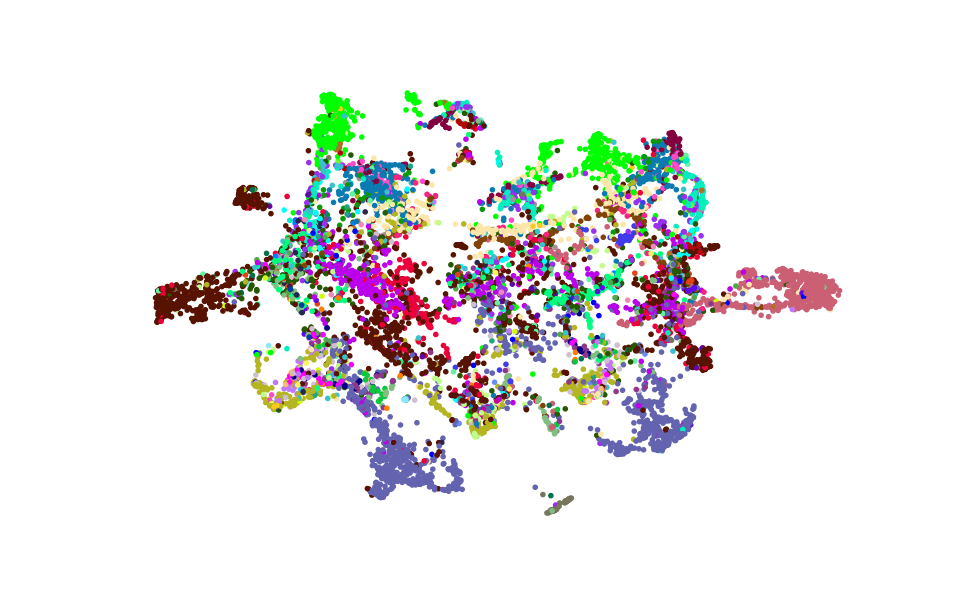}
    \end{tabular}}
    \caption{\textbf{2D visualization via UMAP of the invariant latent embeddings of Shrec17 test data learned by H-(V)AE.} Left: H-AE, Right: H-VAE. Points are colored by class (55 classes).}
    \label{fig:shrec17_umaps}
\end{figure}

\begin{figure}[h]
    \includegraphics[width=1.0\textwidth]{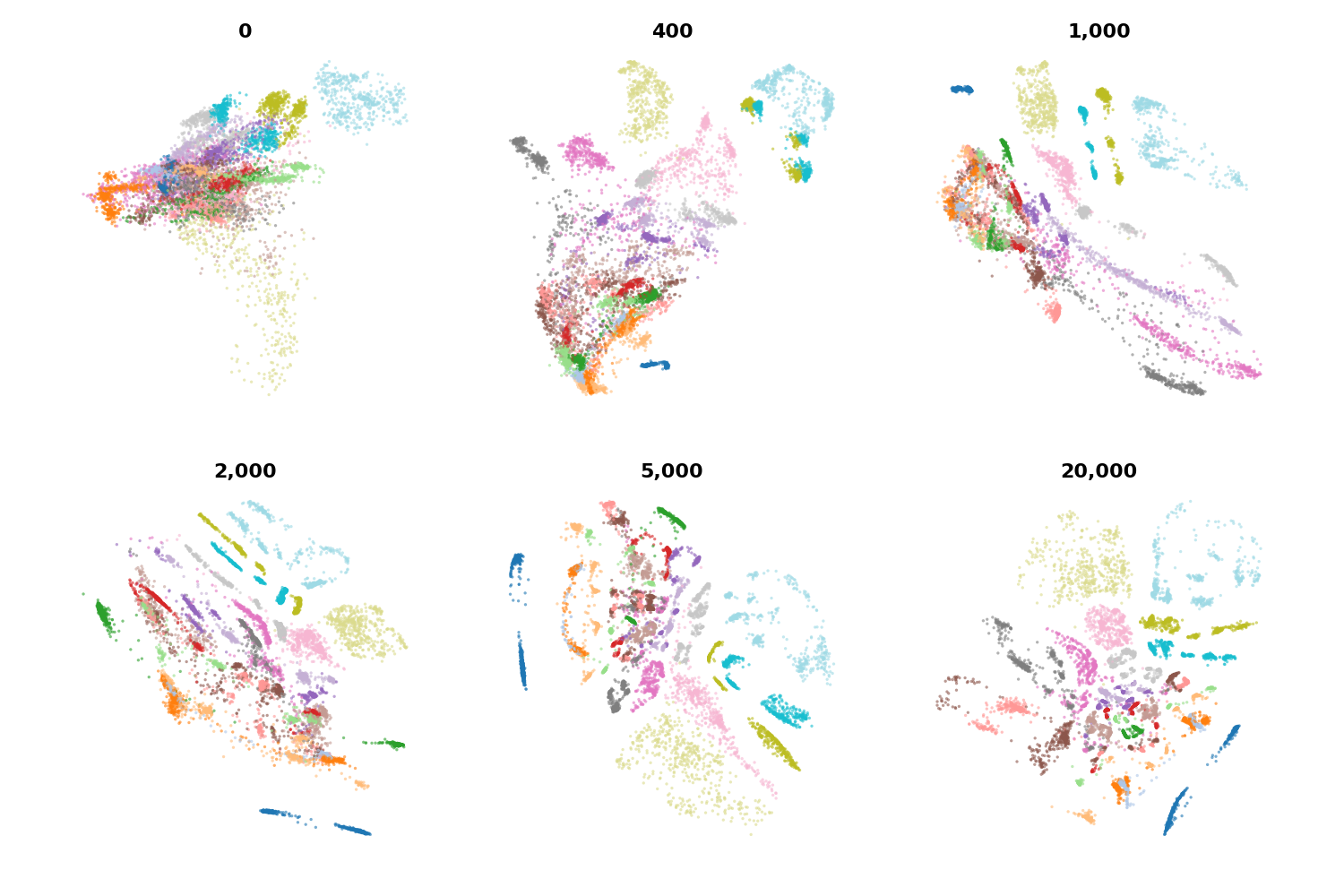}
    \caption{\textbf{Amino acid latent space learned by H-AE.} Visualization of the test data's invariant latent space learned by H-AE trained with varying amounts of the training data. As more training data is added, the separation of clusters containing residues with most similar conformations becomes more distinct. Notably, even with no training data, conformation clusters can be identified.}
    \label{fig:toy_aminoacids_data_ablation_umap_AE}
\end{figure}

\begin{figure}[h]
    \includegraphics[width=1.0\textwidth]{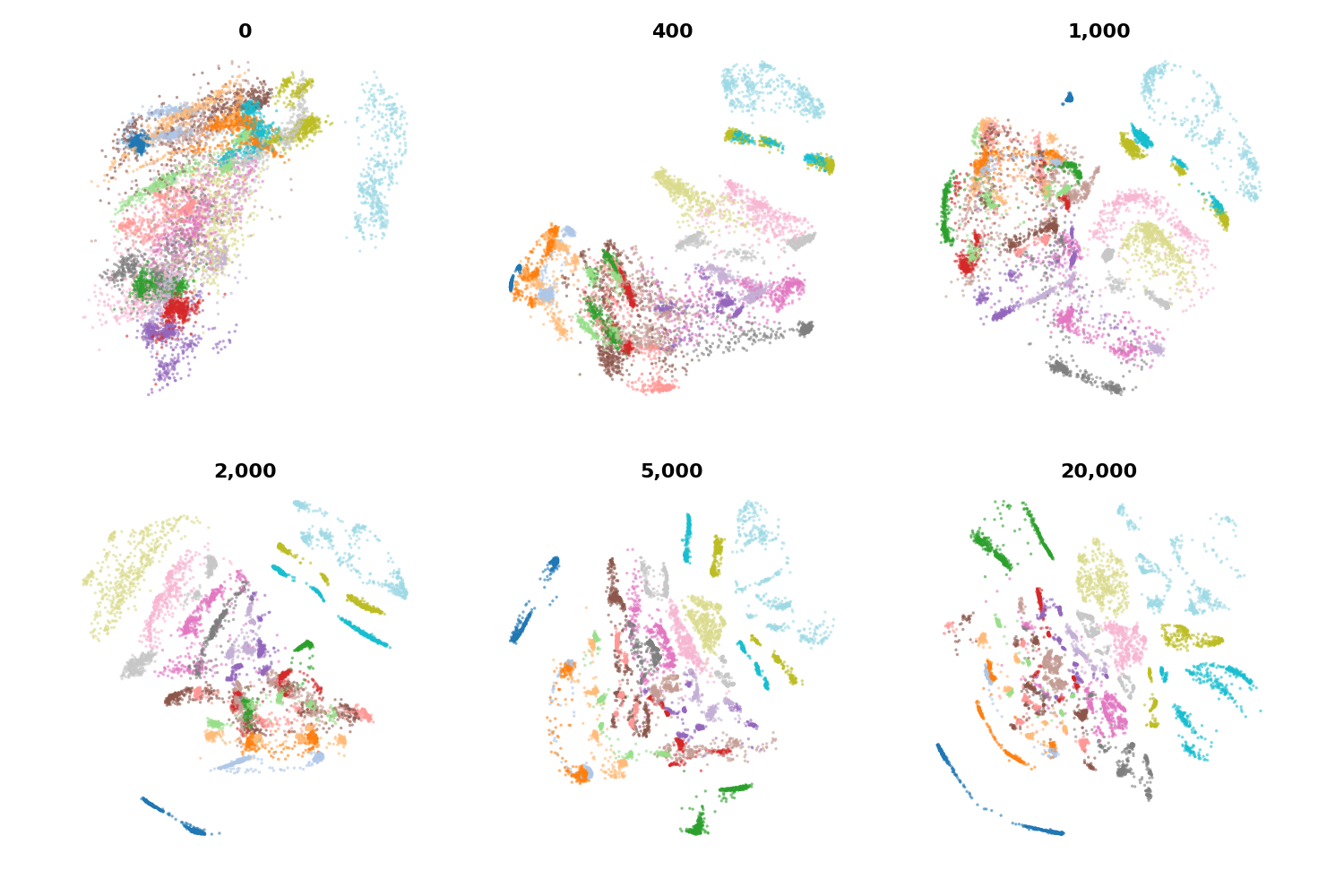}
    \caption{\textbf{Amino acid latent space learned by H-VAE.} Visualization of the test data's invariant latent space learned by H-VAE ($\beta = 0.025$) trained with varying amounts of training data.}
    \label{fig:toy_aminoacids_data_ablation_umap_VAE}
\end{figure}

\begin{figure}[h]
    \centering
    \includegraphics[width=0.8\textwidth]{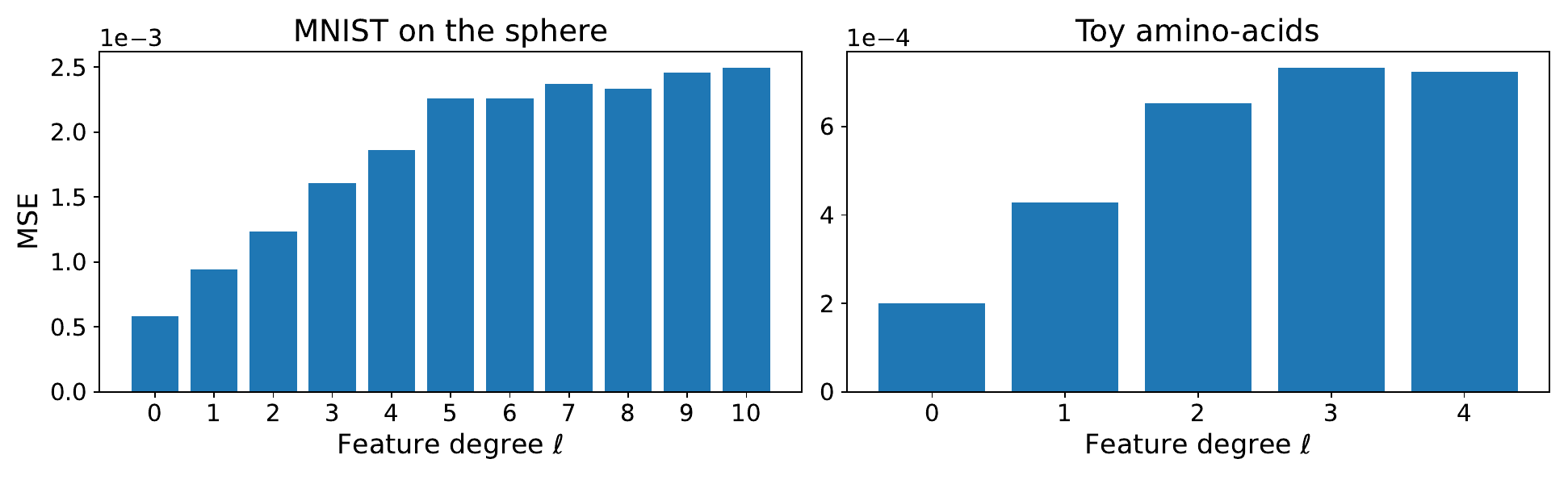}
    \caption{\textbf{Reconstruction loss as a function of feature degree $\ell$.} Test reconstruction loss (MSE) of H-VAE split by feature degree $\ell$, for the MNIST-on-the-sphere (left) and toy amino acids dataset (right). In both cases, features of larger degrees are harder to reconstruct accurately. The increase in loss is more steep for smaller degrees.}
    \label{fig:loss_per_feature}
\end{figure}

\begin{figure}[h]
    \centering
    \includegraphics[width=0.8\textwidth]{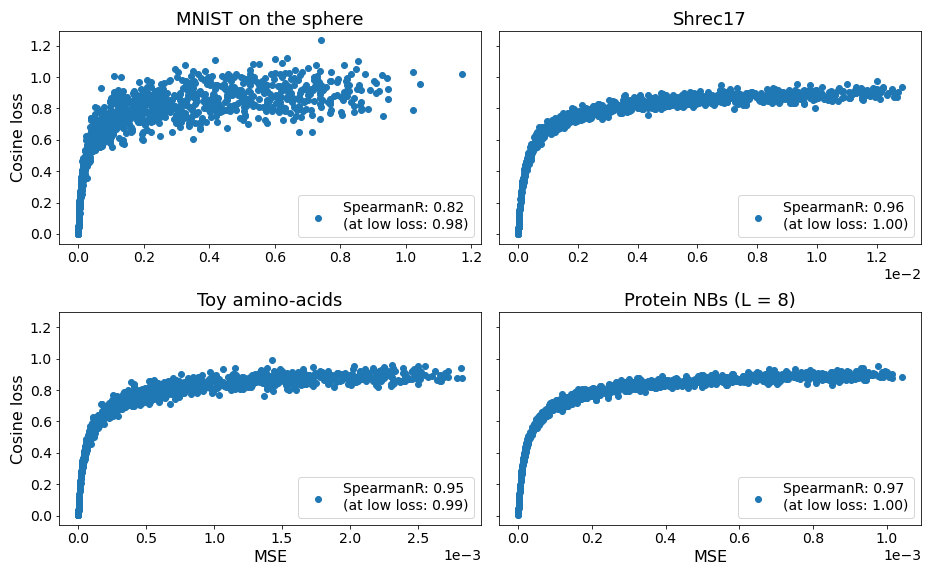}
    \caption{\textbf{Correlation between Cosine loss and MSE values between pairs of random tensors}. For each dataset, we sample a batch of $N = 1000$ tensors with dataset-specific feature degrees and channel sizes, where each coefficient is sampled from a normal distribution. We mimic the normalization step performed in the real experiment and normalize each tensor by the average total norm of the batch. We then generate a ``noisy" version of each tensor by adding (normalized) Gaussian noise to each coefficient with standard deviation sampled from a uniform distribution between $0$ and some maximum noise level ($10$ in these plots). This procedure results in $N$ pairs of tensors with varying degrees of similarity between them. We compute the MSE and Cosine loss for all $N$ pairs of tensors and visualize them. The two loss values are well correlated in rank as measured by Spearman Correlation. The correlation is significantly stronger in the regime of reconstruction loss below a Cosine loss of 0.5 (SpearmanR $\sim 0.99$), a value well above the maximum Cosine loss achieved by H-(V)AE in all our experiments. All the p-values for the Spearman Correlations shown in the plot are significant ($< 0.05$).}
    \label{fig:cosine_vs_mse}
\end{figure}

\end{document}